\documentclass[10pt,reqno]{amsart}
\usepackage{amsaddr}
\usepackage[a4paper, total={6.5in, 8.4in}]{geometry}

\usepackage[symbol]{footmisc}
\usepackage{caption}
\usepackage{comment}
\usepackage{amsmath}
\usepackage{amssymb}
\usepackage{amsthm}
\usepackage[latin1]{inputenc}
\usepackage{eurosym}
\usepackage{graphicx}
\usepackage{epsfig}
\usepackage{hyperref}
\usepackage{dsfont}
\usepackage[space]{cite}
\usepackage{subcaption}
\usepackage{caption}
\usepackage{placeins}
\usepackage{mathtools}
\usepackage{xcolor}
\usepackage{mathrsfs} 
\usepackage[ruled,vlined]{algorithm2e}
\usepackage[title]{appendix}

\usepackage[displaymath,mathlines]{lineno}
\modulolinenumbers[1]

\allowdisplaybreaks

\usepackage{hyperref}
\usepackage{ifthen}
\usepackage{comment}

\makeindex
\usepackage{color}
\usepackage{float}
\usepackage{commath}

\newcommand{\argmin}{\operatorname{argmin}}

\renewcommand{\div}{\operatorname{div}}

\usepackage{subcaption}  % 加载 subcaption 包

\def\leq{\leqslant}
\def\geq{\geqslant}

\numberwithin{equation}{section}

\newtheoremstyle{thmlemcorr}{10pt}{10pt}{\itshape}{}{\bfseries}{.}{10pt}{{\thmname{#1}\thmnumber{
			#2}\thmnote{ (#3)}}}
\newtheoremstyle{thmlemcorr*}{10pt}{10pt}{\itshape}{}{\bfseries}{.}\newline{{\thmname{#1}\thmnumber{
			#2}\thmnote{ (#3)}}}
\newtheoremstyle{defi}{10pt}{10pt}{\itshape}{}{\bfseries}{.}{10pt}{{\thmname{#1}\thmnumber{
			#2}\thmnote{ (#3)}}}
\newtheoremstyle{remexample}{10pt}{10pt}{}{}{\bfseries}{.}{10pt}{{\thmname{#1}\thmnumber{
			#2}\thmnote{ (#3)}}}
\newtheoremstyle{ass}{10pt}{10pt}{}{}{\bfseries}{.}{10pt}{{\thmname{#1}\thmnumber{
			A#2}\thmnote{ (#3)}}}

\theoremstyle{thmlemcorr}
\newtheorem{theorem}{Theorem}
\numberwithin{theorem}{section}
\newtheorem{lemma}[theorem]{Lemma}
\newtheorem{corollary}[theorem]{Corollary}
\newtheorem{proposition}[theorem]{Proposition}

\theoremstyle{thmlemcorr*}
\newtheorem{theorem*}{Theorem}
\newtheorem{lemma*}[theorem]{Lemma}
\newtheorem{corollary*}[theorem]{Corollary}
\newtheorem{proposition*}[theorem]{Proposition}
\newtheorem{problem*}[theorem]{Problem}
\newtheorem{conjecture*}[theorem]{Conjecture}

\theoremstyle{defi}

\newtheorem{problem}{Problem}

\theoremstyle{remexample}

\usepackage{bm}
\usepackage{xcolor}

\newenvironment{notations}{
    \noindent\textbf{Notations:} \it % Bold title and italicized text
}{
    \normalfont % Return to normal font
}
\theoremstyle{ass}

\newtheorem*{notations*}{Notations}

\allowdisplaybreaks

\title[Learning Surrogate Potential Mean Field Games via Gaussian Processes]{Learning Surrogate Potential Mean Field Games via Gaussian Processes: A Data-Driven Approach to Ill-Posed Inverse Problems}
\author{Jingguo Zhang$^1$, Xianjin Yang$^{2,*}$, Chenchen Mou$^3$, Chao Zhou$^1$}

\address[J. Zhang]{
	$^1$Department of Mathematics and Risk Management Institute, National University of Singapore, Singapore.}
\email{e0983423@u.nus.edu}

\thanks{$^*$Corresponding author.}
\address{$^2$Department of Computing and Mathematical Sciences, California Institute of Technology, Pasadena, CA 91125, USA.}
\email{yxjmath@caltech.edu}

\address[C. Mou]{
	$^3$Department of Mathematics, City University of Hong Kong, Hong Kong SAR, China.}
\email{chencmou@cityu.edu.hk}

\email{matzc.nus.edu.sg@gmail.com}

\begin{document}
\begin{abstract}
Mean field games (MFGs) describe the collective behavior of large populations of interacting agents. In this work, we tackle ill-posed inverse problems in potential MFGs, aiming to recover the agents' population, momentum, and environmental setup from limited, noisy measurements and partial observations. These problems are ill-posed because multiple MFG configurations can explain the same data, or different parameters can yield nearly identical observations. Nonetheless, they remain crucial in practice for real-world scenarios where data are inherently sparse or noisy, or where the MFG structure is not fully determined. Our focus is on finding surrogate MFGs that accurately reproduce the observed data despite these challenges.

We propose two Gaussian process (GP)-based frameworks: an inf-sup formulation and a bilevel approach. The choice between them depends on whether the unknown parameters introduce concavity in the objective.  In the inf-sup framework, we use the linearity of GPs and their parameterization structure to maintain convex-concave properties, allowing us to apply standard convex optimization algorithms. In the bilevel framework, we employ a gradient-descent-based algorithm and introduce two methods for computing the outer gradient. The first method leverages an existing solver for the inner potential MFG and applies automatic differentiation, while the second adopts an adjoint-based strategy that computes the outer gradient independently of the inner solver.

Our numerical experiments show that when sufficient prior information is available, the unknown parameters can be accurately recovered. Otherwise, if prior information is limited, the inverse problem is ill-posed, but our frameworks can still produce surrogate MFG models that closely match observed data. These data-consistent models offer insights into underlying dynamics and enable applications such as forecasting and scenario analysis. Finally, because typical potential MFGs are formulated as linear PDE-constrained convex minimization problems, our methods naturally extend to other inverse problems in linear PDE-constrained convex settings.

\end{abstract}
% \keywords{Mean Field Game, Green Economy, Green Insurance, Deep Learning Method}
\maketitle
\section{Introduction}
\label{Introduction}
In this paper, we address ill-posed inverse problems in potential mean field games (MFGs), aiming to recover strategies and environmental parameters from limited, noisy observations of agent populations and environmental configurations. While our primary focus is on numerical methods, the approach naturally extends to broader domains where agents follow MFG dynamics. For example, in epidemiology, it can help infer disease transmission patterns from limited infection data, and in finance, it can uncover traders' strategies from discrete stock holdings, shedding light on market behavior and systemic risk. In these real-world applications, the inverse problems are naturally ill-posed: data are often sparse and noisy, multiple MFG configurations can explain the same observations, and different parameters may still produce closely matching data. Moreover, the MFG structure itself may be partially unknown. In this context, our goal is to find surrogate MFGs that suitably interpret the underlying observations despite these inherent challenges.

MFGs, originally introduced in \cite{lasry2006jeux1, lasry2006jeux, lasry2007mean, huang2006large, huang2007large, huang2007nash, huang2007invariance}, investigate the aggregate behavior of numerous indistinguishable rational agents. As the agent population grows, the resulting Nash equilibrium of a classical MFG is encapsulated by two coupled partial differential equations (PDEs): the Hamilton--Jacobi--Bellman (HJB) equation, which determines the representative agent's value function and the Fokker--Planck (FP) equation, which governs the distribution of agents. For applications of MFGs in various fields, we direct the reader to \cite{gueant2011mean, gomes2015economic, lee2020mean, gao2021belief, gomes2021mean, evangelista2018first, lee2021mean}.

In recent years, the consistency and comprehensiveness of MFGs have been thoroughly investigated in diverse settings. Much of this work originates from the pioneering contributions of Lasry and Lions, further expounded in Lions' lectures at Coll\`ege de France \cite{lionslec}. The limited number of MFG formulations that admit closed-form solutions underscores the importance of numerical methods in MFG research.

A standard time-dependent MFG model typically takes the form
\begin{align}
\label{intro_MFG}
\begin{cases}
-\partial_t u(t, x) - \nu \Delta u + H(x, Du) = f(t, x,  m),
&\quad \forall (t, x) \in (0, T)\times \mathbb{R}^d,\\
\partial_t m(t, x) - \nu \Delta m - \div\bigl(m \,D_pH(x, Du)\bigr) = 0,
&\quad \forall (t, x) \in (0, T)\times \mathbb{R}^d,\\
m(0, x) = \mu(x), \quad u(T, x) = g(x),
&\quad \forall x \in \mathbb{R}^d,
\end{cases}
\end{align}
where $T$ denotes the terminal time, $u$ is the value function for a representative agent, $m$ represents the probability density function of agent locations, $\nu$ describes the volatility of agent movements, $H$ is the Hamiltonian, $f$ encapsulates the mean field interactions, $\mu$ is the initial agent distribution, and $g$ specifies the terminal cost. Under the MFG framework in \eqref{intro_MFG}, the agents' optimal control is given by $-D_pH(x, Du)$ \cite{lasry2006jeux1, lasry2006jeux, lasry2007mean}.

A valuable technique for proving the existence of solutions to \eqref{intro_MFG} is the variational approach introduced in \cite{lasry2007mean}. The central insight, at least at a formal level, is that \eqref{intro_MFG} can be viewed as the first-order condition for the minimizers of the following optimization problem:

\begin{equation}
    \begin{aligned}
\inf _{(m, w)} & \int_0^T \int_{\mathbb{R}^d}[b(x, m(x, t), w(x, t))+F(x,t,m(x, t))] \mathrm{d} x+\int_{\mathbb{R}^d} \mathcal{G}(x, m(x, T)) \mathrm{d} x, \\ &
\text { subject to } \partial_t m-\nu \Delta m+\operatorname{div}(w)=0 \text { in } \mathbb{R}^d \times (0, T), \\
&\quad\quad\quad\quad\quad m(x, 0)=m_0(x) \text { in } \mathbb{R}^d.
\end{aligned}
\label{Potentialt1}
\end{equation}
Here, \(w\) represents the momentum of the agents and is related to \(m\) and \(u\) via \(w = -m\,D_pH\bigl(x,Du\bigr)\). 
In \eqref{Potentialt1}, the functions $b: \mathbb{R}^d \times \mathbb{R} \times \mathbb{R}^d \rightarrow \mathbb{R} \cup\{+\infty\}$, $F: \mathbb{R}^d \times \mathbb{R}^+\times \mathbb{R} \rightarrow$ $\mathbb{R} \cup\{+\infty\}$, and $\mathcal{G}: \mathbb{R}^d \times \mathbb{R} \rightarrow$ $\mathbb{R} \cup\{+\infty\}$ are defined as follows
\begin{align}
\label{eq:defB}
b(x,m,w)
\;=\;
\begin{cases}
m\,H^*\!\Bigl(x,\,-\frac{w}{m}\Bigr), & \text{if } m>0,\\[4pt]
0, & \text{if } (m,w)=(0,0),\\[4pt]
+\infty, & \text{otherwise},
\end{cases}
\end{align}
where \(H^*\) is the Legendre--Fenchel conjugate of \(p\mapsto H(x,p)\). We also let
\begin{align}
\label{eq:coupling_terminal} 
F(x, t, m)
\;=\;
\begin{cases}
\displaystyle \int_{0}^{m} f\bigl(x,  t, m'\bigr)\,dm', & \text{if } m \ge 0,\\[4pt]
+\infty, & \text{otherwise},
\end{cases}
\quad
\mathcal{G}(x,m)
\;=\;
\begin{cases}
\displaystyle \int_{0}^{m} g\bigl(x,m'\bigr)\,dm', & \text{if } m \ge 0,\\[4pt]
+\infty, & \text{otherwise}.
\end{cases}
\end{align}
Under the assumption that \(f\) and \(g\) are non-decreasing, problem~\eqref{Potentialt1} is convex, allowing convex duality techniques to establish existence and uniqueness results for MFG \cite{cardaliaguet2015weak, cardaliaguet2015mean, cardaliaguet2015second, meszaros2015variational, meszaros2018variational}. Moreover, since \eqref{Potentialt1} is a linearly constrained convex minimization problem, it can be solved efficiently using standard convex optimization methods \cite{benamou2015augmented, andreev2017preconditioning, briceno2018proximal, briceno2019implementation, briceno2023primal, briceno2024forward}. Beyond these theoretical and computational advantages, the potential MFG formulation in \eqref{Potentialt1} also motivates the development of new machine learning algorithms \cite{lin2020apac, zhang2023mean}.

In this work, we study ill-posed inverse problems in potential MFGs aimed at deducing agents' strategies and underlying environmental factors. Our approach leverages observations of the agents' distribution together with limited environmental data. Specifically, we focus on the inverse problem for the potential MFG system in \eqref{Potentialt1} under a non-Euclidean metric cost, given by
\begin{align}
\label{eq:non_Eucli_norm}
b(m,w)
\;=\;
\begin{cases}
\displaystyle \frac{|\Lambda w|^q}{q\,m^{q-1}}, & \text{if } m>0,\\[4pt]
0, & \text{if } (m,w)=(0,0),\\[4pt]
+\infty, & \text{otherwise},
\end{cases}
\end{align}
with coupling and terminal terms as in \eqref{eq:coupling_terminal}. Here, the convex conjugate $H^*$ of the Hamiltonian in \eqref{eq:defB} is replaced by \(\tfrac{|\Lambda p|^q}{q}\) for \(q>1\), and \(\Lambda\in \mathbb{R}^{d \times d}\) is introduced to encode a non-Euclidean metric.  Employing a non-Euclidean metric allows for the capture of intricate geometric features of the domain, such as curvature, anisotropy, and spatial variations. A space-dependent, non-Euclidean metric can also be considered. However, for simplicity, we treat \(\Lambda\) here as a constant real-valued matrix. A natural extension would be to model each component of \(\Lambda\) as a function and parametrize it via a GP if spatial variation of the metric is required.

We are concerned with the following problem.
\begin{problem}
\label{main_prob}
Suppose that agents are engaged in an MFG within a competitive environment (For instance, suppose that agents play a potential MFG specified in \eqref{Potentialt1} with the kinetic energy defined in \eqref{eq:non_Eucli_norm}). Based on limited and noisy observations of the agents' population ($m$ in \eqref{intro_MFG}) and partial observations on the environment ($\nu, q, \Lambda$ $F$ and $V$ in \eqref{Potentialt1}), we infer the complete population (values of $m$), the momentum  (values of $w$), and the setup of the environment ($\nu$, $q, \Lambda$, $F$ and $V$). 
\end{problem}

To motivate a potential application of Problem~\ref{main_prob}, consider a large number of traders interacting in a single market, each aiming to optimize personal objectives (e.g., returns or risk). Such a setting naturally fits a potential MFG, where each trader's strategy depends on global market conditions (the ``environment"). In practice, one might only observe the aggregate positions of certain traders at discrete times (akin to partial, noisy measurements of \(m\)) and a few global indicators like volatility or interest rates (analogous to partial observations of \(\nu,q, \Lambda,F\) and \(V\)). From this limited data, the objective is to recover the complete market state: the full distribution of trader holdings (the ``population" $m$), their strategies (the momentum $w$), and key market parameters (\(\nu\), \(q\), $\Lambda$, $F$ and \(V\)).

We further explore Problem \ref{main_prob} by delving into the realms of both general stationary and time-dependent potential MFGs, as detailed in Sections \ref{sec:inf_sup_framework}, \ref{sec:bilevel_framework_stationary} and \ref{sec:time_dependent}. Stationary MFGs' inverse problems are of interest due to their representation as asymptotic limits of time-dependent MFGs \cite{
cannarsa2020long, cirant2021long, cardaliaguet2012long}. Agents typically reach a steady state rapidly under appropriate conditions \cite{
cannarsa2020long, cirant2021long, cardaliaguet2012long}. Therefore, in practical scenarios, observed MFG data is likely in this steady state, emphasizing the need to recover strategies and environmental information at equilibrium.

Problem \ref{main_prob} can be ill-posed, as multiple parameter profiles may explain the same observed data. For instance, as discussed in \cite{guo2024decoding}, if \(\nu = 1\) and only the population density \(m\) is observed, then \(-Du\) can be uniquely identified on the support of \(m\). However, when \(\nu\) is unknown, the Fokker--Planck equation alone does not suffice to determine both \(\nu\) and \(-Du\). Furthermore, different combinations of \(F\) and \(V\) may yield identical MFGs that generate the observed data, and small variations in these parameters can produce nearly indistinguishable outcomes. In this paper, we incorporate prior information and regularization techniques to learn a surrogate MFG that adequately explains the observed data.

We tackle Problem \ref{main_prob} through GPs, a powerful non-parametric statistical method renowned for its ability to model and predict complex phenomena with a degree of uncertainty. The beauty of GPs lies in their inherent flexibility, allowing them to capture a wide range of behaviors and patterns without being tied down by a strict parametric form. We provide a general GP framework that determines the unknown variables using maximum a posteriori estimators. These estimators are conditioned on the minimization of the PDE-constrained potential MFGs. Additionally, the process incorporates regularization through the inclusion of observations that are subject to noises.

More precisely, we propose two frameworks that depend on the underlying mathematical structure of the problem. If the objective function is concave w.r.t the unknowns, we adopt an inf-sup formulation, which permits a primal-dual approach to solving the inverse problem. For a general potential MFG (where the objective function is not concave in unknowns), we introduce a bilevel formulation in which the outer minimization seeks to optimize a regularized objective plus a data fidelity term, subject to an inner minimization that enforces the potential MFG constraint. A related bilevel approach for inverse problems in potential MFGs is presented in~\cite{yu2024bilevel}; however, there are notable differences. First, our approach differs in scope. We aim to recover a larger set of unknowns, specifically $\nu,  q, \Lambda, F,$ and $V$, whereas \cite{yu2024bilevel} considers a more restricted recovery problem. Second, our methods differ in the training of the bilevel minimization problems.  In~\cite{yu2024bilevel}, the unknowns are discretized via finite-difference, and an alternating minimization is used: the inner problem is partially solved by gradient descent (unrolled for a fixed number of steps), then the outer gradient is computed. By contrast, we parametrize the unknowns with GPs and propose two methods for solving our bilevel problem. The first method solves the inner minimization by a primal-dual algorithm~\cite{briceno2018proximal, briceno2019implementation} and relies on automatic differentiation for the outer gradient. The second method employs an adjoint approach, which derives an analytic expression for the outer gradient and, therefore, does not rely on the details of how the inner problem is solved. While the automatic differentiation strategy can be simpler to implement, its accuracy depends on both the inner solver and the method used therein. The adjoint method avoids this limitation by directly computing the exact outer gradient.

We note that in \cite{guo2024decoding}, the authors consider a similar inverse problem to Problem \ref{main_prob}, but in the more general context of MFGs formulated via PDEs. There, the unknowns are parameterized by GPs and identified through the minimization of a loss function comprising a data-fidelity term and a regularization term, subject to PDE constraints enforced at collocation points. The authors employ a Gauss--Newton method to solve this inverse problem; however, it is well known that Gauss--Newton converges only when the initial guess is sufficiently close to the true solution. By contrast, our focus on \emph{potential} MFGs enables us to leverage a convex structure. In the inf--sup formulation of Section \ref{sec:inf_sup_framework}, the inverse problem reduces to a convex--concave minimization that we solve via the primal--dual method of \cite{briceno2018proximal}, thereby ensuring global convergence. Meanwhile, in the bilevel formulation of Sections \ref{sec:bilevel_framework_stationary} and \ref{sec:time_dependent}, the inner minimization is itself the solution of a potential MFG. As a result, each parameter set produced by the solver automatically satisfies some potential MFG system. This enables us to identify a surrogate MFG that reproduces data closely matching the underlying dataset, even if the parameters we obtain do not correspond to those of the true model generating the data.

In both the inf-sup and bilevel frameworks for solving the inverse problem, we solve \eqref{Potentialt1} using the finite-difference and primal-dual methods from \cite{briceno2018proximal, briceno2019implementation}. While a purely GP-based method for solving potential MFGs is a promising direction for future research, our primary focus is on inverse problems for recovering unknown parameters; hence, we employ existing, well-established methods for computing potential MFGs. Notably, our approach is compatible with any solver, but the methods in \cite{briceno2018proximal, briceno2019implementation} are particularly advantageous due to their accuracy and their well-understood convergence properties.

Another key contribution of our work, compared to the existing literature, is the way we handle the coupling function \(F\). For uniqueness, it is necessary that \(F\) be convex, satisfying the Lasry--Lions monotonicity condition \cite{lasry2007mean}. The standard approach in prior research is to assume a convex ansatz for \(F\); for example, taking \(F\) to be an entropy of the density, a polynomial of the density, or a convolution of the density with some kernel \cite{ding2022mean, guo2024decoding}. In contrast, we propose a unified framework for handling \(F\) under three distinct scenarios, each corresponding to a different level of prior knowledge about \(F\) and leading to three separate estimation strategies. First, as in previous studies \cite{guo2024decoding}, if we assume a known parameterized convex structure for \(F\), we estimate its unknown coefficients. For instance, we may assume \(F(m) = \frac{1}{\alpha} m^\alpha\) for some \(\alpha > 0\) and estimate \(\alpha\). Next, we may know that \(F\) belongs to a predefined library of convex functions generated by weighted combinations of multiple convex functions.  Finally, in the most general case, we assume no prior knowledge about \(F\) other than its convexity. In this setting, we approximate \(F\) using a suitable basis expansion or a GP framework, enforcing convexity through a regularization term on the parameterized surrogate.  This flexible approach accommodates a broad range of forms for \(F\) while ensuring the required convexity condition.

In Section \ref{sec:Experiment}, we demonstrate the effectiveness of our methodology through various examples, focusing on reconstructing population profiles, agent strategies, and environmental settings from partial and noisy observations. The numerical results indicate that Problem \ref{main_prob} is ill-posed without sufficient prior information, as multiple parameter sets can fit the data without necessarily matching the true parameters. However, when adequate prior knowledge is available, the unknowns can be accurately recovered. Even in cases with limited priors, our bilevel framework reliably produces surrogate MFG models that closely match observed data, offering insights into underlying dynamics and supporting applications such as forecasting and scenario analysis. Furthermore, since potential MFGs can be formulated as linear PDE-constrained convex minimization problems, our methods extend naturally to broader inverse problems in such settings.

\subsection{Related Works in MFGs}
Although numerous numerical algorithms exist for solving MFGs (see, for example, \cite{achdou2010mean, achdou2012iterative, nurbekyan2019fourier, liu2021computational, liu2020splitting, briceno2018proximal, briceno2019implementation, carmona2021convergence, carmona2019convergence, ruthotto2020machine, lin2020apac, lauriere2021numerical, achdou2020mean, gomes2020hessian, mou2022numerical, meng2023sparse}), relatively few studies address inverse problems in MFGs \cite{liu2023inverse, ding2022mean, chow2022numerical, klibanov2023convexification, klibanov2023h, imanuvilov2023lipschitz, liu2023simultaneously, ding2023determining, ren2024policy}. Among the works focusing on numerical methods for inverse MFGs, the study \cite{ding2022mean} introduces models to reconstruct ground metrics and interaction kernels in the running costs, demonstrating efficiency and robustness through numerical experiments. The paper \cite{chow2022numerical} presents a numerical algorithm to solve an inverse MFG problem based on partial boundary measurements. The work \cite{klibanov2023convexification} proposes a globally convergent convexification method for recovering the global interaction term from a single measurement. A more recent study \cite{yang2023context} tackles mean-field control inverse problems via operator learning, training models on pairs of input and output data computed under identical parameters. By contrast, this paper estimates unknowns using only a single observation.

In \cite{yu2024bilevel}, a bilevel optimization framework is proposed for inverse MFGs with unknown obstacles and metrics. The work \cite{guo2024decoding} introduces a GP method for solving inverse problems in general MFGs, aiming to recover all unknown components. Unlike \cite{guo2024decoding}, which focuses on general MFGs and employs a parametric convex ansatz for the coupling function, our approach targets potential MFGs and places particular emphasis on recovering the unknown coupling function under the assumption that \(F\) is convex, thereby offering more flexibility in capturing complex coupling effects.

\subsection{Related Works in GPs}
GP regression, a Bayesian non-parametric technique for supervised learning, is notable for its ability to quantify uncertainty. GPs have been successfully applied to solving and learning ordinary differential equations (ODEs) \cite{hamzi2023learning, yang2023learning} and PDEs \cite{chen2021solving, chen2023sparse, yang2023mini, raissi2017machine, raissi2018numerical}. In the context of MFGs, GPs have also been employed to solve MFG systems \cite{mou2022numerical, meng2023sparse} and tackle inverse problems \cite{guo2024decoding}.

This paper extends the earlier work on learning unknowns in PDEs using GPs to address inverse problems in potential MFGs, which can be formulated as a PDE-constrained convex minimization problem. The key insight behind using GPs for approximating unknowns is that the GP framework, being a linear model, preserves the convexity structure of the problem. This property allows us to formulate an inf-sup framework for inverse problems, as detailed in Section \ref{sec:inf_sup_framework}, and to employ convex optimization methods to solve the resulting inf-sup minimization problem. This advantage distinguishes the GP-based approach from other parameterization methods, such as neural networks, which do not inherently preserve the convexity structure.

\medskip
\noindent
\begin{notations}
\noindent
In this paper, we write a real-valued vector in boldface (e.g., $\boldsymbol{v}$) unless it specifically denotes a point in the physical domain. The Euclidean norm of $\boldsymbol{v}$ is $|\boldsymbol{v}|$, and its transpose is $\boldsymbol{v}^T$. For a vector $\boldsymbol{v}$, $v_i$ denotes its $i^\text{th}$ component. For a matrix $M$, $|M|$ stands for its $L^2$ norm. Given a function $u$ and a vector $\boldsymbol{v}$ of length $N$, $u(\boldsymbol{v})$ is the vector $(u(v_1), \dots, u(v_N))$. We use $\delta_x$ to denote the Dirac delta function concentrated at $x$. Let $\Omega \subset \mathbb{R}^d$. We denote the interior and boundary of $\Omega$ by $\operatorname{int}\Omega$ and $\partial\Omega$, respectively. The notation $\mathcal{N}(0,\gamma^2 I)$ refers to the standard multivariate normal distribution with covariance $\gamma^2 I$ (where $\gamma>0$). In a normed vector space $V$, $\|\cdot\|_V$ signifies its norm.

Consider a Banach space $\mathcal{U}$ endowed with a quadratic norm $\|\cdot\|_{\mathcal{U}}$. Its dual space, $\mathcal{U}^*$, is equipped with the duality pairing $[\cdot,\cdot]$. Suppose there exists a covariance operator $\mathcal{K}_{\mathcal{U}} : \mathcal{U}^* \to \mathcal{U}$ that is linear, bijective, symmetric (i.e.\ $[\mathcal{K}_{\mathcal{U}}\phi, \psi] = [\mathcal{K}_{\mathcal{U}}\psi, \phi]$), and positive ($[\mathcal{K}_{\mathcal{U}}\phi, \phi] > 0$ whenever $\phi \neq 0$). The norm in $\mathcal{U}$ is then given by
\[
\|u\|_{\mathcal{U}}^2 = \bigl[\mathcal{K}_{\mathcal{U}}^{-1}u,\,u\bigr], 
\quad
\forall\,u \in \mathcal{U}.
\]
Let $\boldsymbol{\phi} = (\phi_1, \dots, \phi_P)$ with $P \in \mathbb{N}$ be an element of ${(\mathcal{U}^*)}^{\otimes P}$. For $u \in \mathcal{U}$, we define
\[
[\boldsymbol{\phi},\,u] := \bigl([\phi_1,\,u], \dots, [\phi_P,\,u]\bigr).
\]
Let \(K\) be a kernel, and let \(\boldsymbol{\phi} = \{\phi_i\}\) be a collection of linear operators in the dual space of the reproducing kernel Hilbert space (RKHS) associated with \(K\). We define $K\bigl(x,\boldsymbol{\phi}\bigr)$ as a vector whose \(i^\text{th}\) component is
\(\displaystyle \int_{\Omega} K(x,x')\,\phi_i(x')\,\mathrm{d}x'\).
We also define \(K(\boldsymbol{\phi}, \boldsymbol{\phi})\) as the Gram matrix whose \((i,j)^\text{th}\) entry is
\[
\int_{\Omega}\!\!\int_{\Omega} K(x,x')\,\phi_i(x)\,\phi_j(x')
\,\mathrm{d}x\,\mathrm{d}x'.
\]

\end{notations}

\section{Overviews for Potential MFGs and GP Regressions}
\label{sec:Prerequisites}
\subsection{Stationary Potential MFGs}
\label{subsec:stationary}
In this subsection, we focus on the following MFG problem on the $d$-dimensional torus $\mathbb{T}^d$ with non-Euclidean metric
\begin{align}
\label{eq:gsmfg}
\begin{cases}
\displaystyle \inf_{m,w} \;\int_{\mathbb{T}^d}\!\Bigl[b(m,w) + V(x)\,m(x) + F\bigl(m(x)\bigr)\Bigr]\!\,dx,\\[6pt]
\text{subject to}\quad -\nu\Delta m(x) + \div(w)(x) = 0,\quad x \in \mathbb{T}^d,\\
\quad\quad\quad\quad\quad\int_{\mathbb{T}^d} m(x)\,dx = 1,
\end{cases}
\end{align}
where
\[
b(m, w) = 
\begin{cases}
\tfrac{|\Lambda w|^q}{q\,m^{q-1}}, & m > 0,\\
0, & (m, w) = (0, 0),\\
+\infty, & \text{otherwise},
\end{cases}
\quad
F(m) = 
\begin{cases}
\displaystyle \int_0^{m} f(s)\,\mathrm{d}s, & m \ge 0,\\
0, & \text{otherwise}.
\end{cases}
\]
Here, $\Lambda\in  \mathbb{R}^{d\times d}$ represents the matrix of a non-Euclidean metric, $w$ denotes the flux, $m$ is the density distribution of agents, $V$ is a spatial cost, $f$ introduces density-dependent coupling, and $\nu>0$ is the diffusion coefficient. Given $(m, w)$, this formulation captures the transport cost $b(m, w)$ and the density-dependent term $F(m) + Vm$. 

\medskip

\noindent
\textbf{Discretization.} Below, we summarize the discretization of \eqref{eq:gsmfg} on $\mathbb{T}^2$. 
Following \cite{briceno2018proximal}, we discretize $\mathbb{T}^2$ by a toroidal grid $\mathbb{T}_h^2$ with spacing $h$ and $N_h = 1/h$ points per dimension. For a function $y : \mathbb{T}_h^2 \to \mathbb{R}$, the notation $y_{i,j}$ refers to its value at the grid point $(i,j)$. We define the discrete gradient and Laplacian by
\[
(D_1 y)_{i,j} := \frac{y_{i+1,j} - y_{i,j}}{h}, \quad
(D_2 y)_{i,j} := \frac{y_{i,j+1} - y_{i,j}}{h}, \quad
(\Delta_h y)_{i,j} := -\frac{4y_{i,j} - y_{i+1,j} - y_{i-1,j} - y_{i,j+1} - y_{i,j-1}}{h^2}.
\]
Let $\mathcal{M}_h = \mathbb{R}^{N_h \times N_h}$ represent the space of discrete densities and $\mathcal{W}_h = (\mathbb{R}^4)^{N_h \times N_h}$ the space of discrete flows. Let $K:=\mathbb{R}_{+}\times\mathbb{R}_{-}\times\mathbb{R}_{+}\times\mathbb{R}_-$. We introduce a discrete version non-Euclidean metric matrix $\Lambda\in \mathbb{R}^{4\times 4}$ and define the cost functional
\begin{align}
\label{eq:BBhat}
\mathcal{B}(m,w) = \sum_{i,j} \hat{b}( m_{i,j}, w_{i,j}), \quad
\hat{b}(m,w) =
\begin{cases}
\displaystyle \frac{|\Lambda w|^q}{q\,m^{q-1}}, & m>0, w\in K, \\
0, & (m,w)=(0,0), \\
+\infty, & \text{otherwise},
\end{cases}
\end{align}
along with
\[
\mathcal{F}(m) = \sum_{i,j}  F\bigl(m_{i,j}\bigr).
\]
To discretize the constraint $-\nu \Delta m +  \div w = 0$ and mass normalization, we set
\[
(A m)_{i,j} = -\nu (\Delta_h m)_{i,j}, \quad
(B w)_{i,j} = (D_1 w^1)_{i-1,j} + (D_1 w^2)_{i,j} + (D_2 w^3)_{i,j-1} + (D_2 w^4)_{i,j},
\]
and combine them into
\[
G(m, w) = \bigl( A m + B w,\;\, h^2 \sum_{i,j} m_{i,j} \bigr).
\]
Thus, the fully discrete problem reads
\begin{align}
\label{eq:disc:stationary}
\inf_{(m,w)\in \mathcal{M}_h\times\mathcal{W}_h}
\bigl[\mathcal{B}(m,w) + \mathcal{F}(m) +  \sum_{i,j} V(x_{i,j})m_{i,j}\bigr],
\quad
\text{subject to } G(m,w) = (0,1).
\end{align}
The authors of \ \cite{briceno2018proximal} evaluate several algorithms for solving \eqref{eq:disc:stationary}, finding that the primal-dual method \cite{chambolle2011first} excels in both speed and accuracy.

\subsection{Time-Dependent Potential MFGs}
\label{subsec:time-dependent}In this subsection, 
we consider the following time-dependent MFG on the $d$-dimensional torus 
\begin{align}
\label{eq:time-dependent}
\begin{cases}
\displaystyle \inf_{m, w}
\int_{0}^{T}\!\int_{\mathbb{T}^d}
\Bigl[
b\bigl(m(x,t), w(x,t)\bigr) + V(x,t)\,m(x,t) + F\bigl(t, m(x,t)\bigr)
\Bigr]
\,dx\,dt 
\;+\;
\int_{\mathbb{T}^d}
\mathcal{G}\bigl(x,m(x,T)\bigr)\,dx,\\[6pt]
\text{s.t. } \quad
\partial_t m - \nu\,\Delta m + \mathrm{div}(w)=0,
\quad
m(\cdot,0)=m_0,
\end{cases}
\end{align}
where $T$ is the terminal time, 
\[
b(m,w)
=
\begin{cases}
\frac{|\Lambda w|^q}{qm^{q-1}}, & \text{if } m > 0,\\[4pt]
0, & \text{if } (m,w)=(0,0),\\[4pt]
+\infty, & \text{otherwise},
\end{cases}
\]
and \[
F(t, m)
=
\begin{cases}
\displaystyle \int_{0}^{m} f(t, m')\,dm', & \text{if } m\ge 0,\\[4pt]
+\infty, & \text{otherwise},
\end{cases}
\quad
\mathcal{G}(x,m)
=
\begin{cases}
\displaystyle \int_{0}^{m} g(x,m')\,dm', & \text{if } m\ge 0,\\[4pt]
+\infty, & \text{otherwise}.
\end{cases}
\]
Here, \(w\) denotes the flux, \(m\) is the density of agents, \(V\) is a space-time cost function, \(F\) is a continuous coupling function, and \(\mathcal{G}\) represents the terminal cost. The parameter \(q>1\) is an exponent in the cost functional, and \(\Lambda\in  \mathbb{R}^{d\times d}\) denotes the (possibly non-Euclidean) spatial metric. Lastly, \(m_0\) is a nonnegative initial density with a total mass of 1.

\medskip
\noindent
\textbf{Discretization.}
Let \(0=t_0 < t_1 < \dots < t_{N_T}=T\) be a partition of the interval \([0,T]\) with \(\Delta t = t_{k+1}-t_k\), and let \(\{x_{i,j}\}\) be a spatial mesh on \(\mathbb{T}^2\). We denote the discrete values \(m^k_{i,j}\approx m(x_{i,j},t_k)\) and \(w^k_{i,j}\approx w(x_{i,j},t_k)\), collecting them into the arrays
\[
m = \bigl\{m^k_{i,j}\bigr\},
\quad
w = \bigl\{w^k_{i,j}\bigr\}.
\]
Define the spaces \(\mathcal{M}\) for these discrete density arrays and \(\mathcal{W}\) for the flux arrays. We approximate \(\partial_t m - \nu\,\Delta m\) by a linear operator \(\widetilde{A}\) and \(\mathrm{div}(w)\) by a linear operator \(\widetilde{B}\). Concretely, \(\widetilde{A}\) uses finite differences in time and a discrete Laplacian, so \((\widetilde{A}m)^k_{i,j}\) approximates
\(\bigl(m^{k+1}_{i,j}-m^k_{i,j}\bigr)/\Delta t - \nu\,\Delta_h\bigl(m^{k+1}\bigr)_{i,j}\).
Similarly, \((\widetilde{B}w)^k_{i,j}\) approximates the divergence of \(w^k\) at \((i,j)\). The discrete continuity equation then reads
\[
\widetilde{A}\,m + \widetilde{B}\,w = (0, m^0),
\]
where \(m^0\) encodes the initial condition at \(t=0\). Next, we define the discrete cost functional by local sums of the form \(\widehat{b}(m^k_{i,j},w^k_{i,j})\), where \(\widehat{b}\) is as in \eqref{eq:BBhat}. Let \(V^k_{i,j}\approx V(x_{i,j},t_k)\). We then set
\[
\mathcal{B}(m,w)
=
\sum_{k=0}^{N_T-1}
\sum_{(i,j)}
\widehat{b}\bigl(m^k_{i,j},\,w^k_{i,j}\bigr),
\quad
\mathcal{F}(m)
=
\sum_{k=0}^{N_T-1}
\sum_{(i,j)}
F\bigl(m^k_{i,j}\bigr).
\]
The fully discrete MFG problem becomes
\begin{align}
\label{eq:time-dependent:discr}
\inf_{(m,w)\in \mathcal{M}\times\mathcal{W}}
\Bigl[
\mathcal{B}(m,w)
+
\mathcal{F}(m) + \sum_{k=0}^{N_T-1}\sum_{(i,j)}V^k_{i,j}\,m^k_{i,j} + \sum_{(i,j)}
\mathcal{G}\bigl(x_{i,j},\,m^{N_T}_{i,j}\bigr)
\Bigr]
\text{ subject to }
\widetilde{A}\,m + \widetilde{B}\,w = (0,m^0).
\end{align}
Solving this finite-dimensional convex program yields discrete approximations \((m,w)\) of the continuous MFG solution. By refining the spatial mesh and taking \(\Delta t \to 0\), these discrete solutions converge (in appropriate senses) to solutions of the original PDE system \cite{briceno2019implementation}. In practice, one can again apply the primal-dual method to solve \eqref{eq:time-dependent:discr} \cite{briceno2019implementation}.

\subsection{Gaussian Process Regression for Function Approximation}
\label{Guassian}
\noindent
GPs provide a flexible, nonparametric framework for approximating unknown functions in statistical learning, accommodating a wide range of complex phenomena without imposing a fixed functional form. Let $\Omega \subseteq \mathbb{R}^d$ be open. A real-valued GP ${f}: \Omega \to \mathbb{R}$ is defined so that, for any finite set of points $\boldsymbol{x}$, the random variable ${f}(\boldsymbol{x}) \in \mathbb{R}^{N}$ follows a joint Gaussian distribution. Such ${f}$ is characterized by a mean function ${\mu}: \Omega \to \mathbb{R}$ and a covariance function $K: \Omega \times \Omega \to \mathbb{R}$, i.e.,
\[
\mathbb{E}[{f}(x)] \;=\; {\mu}(x)
\quad \text{and} \quad
\mathrm{Cov}\bigl({f}(x), {f}(x')\bigr)
\;=\;
K(x, x'),
\]
for all $x, x' \in \Omega$, which we denote by ${f} \sim \mathcal{G}\mathcal{P}\bigl({\mu}, K\bigr)$. In the context of learning real-valued functions, the objective is to construct a GP estimator ${f}^\dagger$ from training data $\bigl(x_i, y_i\bigr)_{i=1}^N$. Under the prior ${f} \sim \mathcal{G}\mathcal{P}({0},K)$, we define
\[
{f}^{\dagger}
\;=\;
\mathbb{E}\bigl[{f} \,\mid\, {f}(x_i) = y_i,\; i=1,\dots,N\bigr].
\]
Let $\boldsymbol{y}$ be the vector whose $i^\text{th}$ element is $y_i$. The posterior mean ${f}^{\dagger}$ admits the closed-form expression
\[
{f}^{\dagger}(x)
\;=\;
K\bigl(x, \boldsymbol{x}\bigr)\,
K\bigl(\boldsymbol{x}, \boldsymbol{x}\bigr)^{-1}\,\boldsymbol{y},
\]
where $K\bigl(x, \boldsymbol{x}\bigr)$ is an $N$-dimensional vector obtained by concatenating $K(x,x_i)$ for $i=1,\dots,N$, and
\[
K\bigl(\boldsymbol{x}, \boldsymbol{x}\bigr)
\;=\;
\begin{bmatrix}
K\bigl(x_1,x_1\bigr) & \cdots & K\bigl(x_1,x_N\bigr)\\
\vdots & \ddots & \vdots\\
K\bigl(x_N,x_1\bigr) & \cdots & K\bigl(x_N,x_N\bigr)
\end{bmatrix}.
\]
Moreover, ${f}^{\dagger}$ can be interpreted as the minimizer of the following optimal recovery problem in the associated vector-valued RKHS $\mathcal{U}$ associated with the kernel $K$:
\[
\min_{{f} \in \mathcal{U}}
\|{f}\|_{\mathcal{U}}^2
\quad
\text{s.t.}
\quad
{f}(x_i) = {y}_i,\;\; i=1,\dots,N.
\]

\section{An Inf-Sup Framework for Inverse Problems of Potential MFGs}
\label{sec:inf_sup_framework}
In this section, we present an inf-sup minimization framework for solving inverse problems in stationary MFGs, assuming the unknown parameters appear as concave terms in potential MFG objective functions. Leveraging the linearity of GP models, this approach preserves the intrinsic convexity-concavity structure of the objective. For clarity, we illustrate the method by solving inverse problems for \eqref{eq:gsmfg} with unknown \(V\), though it naturally extends to time-dependent MFGs and other concave unknowns.

Here, we consider the following inverse problem.
\begin{problem}
\label{inf_sup_st_prob}
Suppose agents are playing in the MFG \eqref{eq:gsmfg}. Given \( V^*, F^*, \Lambda^*, q^* \), and \( \nu^* \), assume that \eqref{eq:gsmfg} admits a unique classical solution \( (m^*, w^*) \). Furthermore, suppose we have access to partial and noisy observations of \( m^* \) and \( V^* \), while \( F^*, \Lambda^*, q^* \), and \( \nu^* \) are known. The goal is to estimate the functions \( w^* \), \( m^* \), and \( V^* \). To elaborate, we have 
\begin{enumerate}
    \item \textbf{Partial noisy observations on $m^*$}.  We have a set of linear operators $\{\phi_{m,l}^o\}_{l=1}^{N_m}$, $N_m\in \mathbb{N}$ where some noisy observations on $m^*$ are available. These observations are represented as $\boldsymbol{m}^o$, i.e. $\boldsymbol{m}^o = ([\phi_{m,1}^o, m^*], \dots, [\phi_{m, N_m}^o, m^*]) + \boldsymbol{\epsilon}_m$, $\boldsymbol{\epsilon}_m\sim \mathcal{N}(0, \gamma^2_m I)$, $\gamma_m>0$. We call $\boldsymbol{\phi}_m^o=(\phi_{m,1}^o, \dots, \phi_{m, N_m}^o)$ the vector of observation operators. For instance, if we only have observations of $m^*$ at a finite set of collocation points,  $\boldsymbol{\phi}^o_m$ contains Diracs centered at these points. 
    \item \textbf{Partial noisy observations of ${V}^*$}. We possess noisy observations at collections of  linear operators $\{\phi^{o}_{V,e}\}_{e=1}^{N_{v}}$. Denote $\boldsymbol{\phi}_{V}^o = (\phi^o_{V,1}, \dots, \phi^o_{V, N_v})$. These observations can be collected into a vector $\boldsymbol{V}^o$, 
\begin{align*}
\boldsymbol{V}^o = ([\phi^{o}_{V,1}, V^*], \dots, [\phi^{o}_{V, N_v}, V^*]) + \boldsymbol{\epsilon}_V, \boldsymbol{\epsilon}_V \sim \mathcal{N}(0, \gamma^2_VI), \gamma_V>0. 
\end{align*}
\end{enumerate}
\textbf{Inverse Problem Statement}\\
Given \( {F}^* \), \( \Lambda^* \), \( q^* \), and \( \nu^* \), assume that agents participate in the stationary MFG described in \eqref{eq:gsmfg}. The task is to infer \( w^* \), \( m^* \), and \( V^* \) using the observed data \( \boldsymbol{m}^o \) and \( \boldsymbol{V}^o \).
\end{problem}

\begin{comment}
In this section, we consider the general form of stationary  MFG 
\begin{equation}
   \begin{cases}-\nu \Delta u+H(x, \nabla u)-\lambda =f(x,m(x)), &  \text { in } \mathbb{T}^n, \\-\nu \Delta m-\operatorname{div}(m \nabla u) =0, & \text { in } \mathbb{T}^n, \\m \geq 0, \quad \int_{\mathbb{T}^n} m \mathrm{~d} x=1, \quad \int_{\mathbb{T}^n} u \mathrm{~d} x=0. & \end{cases} \label{Origins}
\end{equation}
Its corresponding potential form is as follows 
\begin{equation}
\begin{aligned}
        \inf _{(m, w)} \quad &\int_{\mathbb{T}^n}[b(m(x), w(x))+F(x, m(x))] \mathrm{d} x,\\
         \text{s.t.} \quad &-\nu \Delta m+\operatorname{div}(w)=0 \text { in } \mathbb{T}^n,\\
         & \int_{\mathbb{T}^n} m(x) \mathrm{d} x=1 .
\end{aligned}
\label{Potentialstationary}
\end{equation}
\end{comment}

\subsection{An Inf-sup Framework}
To solve Problem~\ref{inf_sup_st_prob}, we assume that \(V\) lies in a RKHS \(\mathcal{V}\) associated with a kernel \(K_{\mathcal{V}}\). This assumption is not restrictive, as RKHSs with Mat\'ern kernels are equivalent to Sobolev spaces \cite{kanagawa2018gaussian}. Since the objective function in \eqref{eq:gsmfg} depends linearly on \(V\), we can formulate the following inf-sup problem:
\begin{equation}
\label{eq:inf-sup-framework}
\begin{aligned}
        \inf _{(m, w)}\sup _{V\in \mathcal{V}} \quad & \int_{\mathbb{T}^d}b(m, w)+V(x)m(x)+F(m)(x)\dif x+\frac{\alpha_{m^o}}{2}\left|\left[\boldsymbol{\phi}^o_m, m\right]-\boldsymbol{m}^o\right|^2\\&-\frac{\alpha_{v^o}}{2}\left|[\boldsymbol{\phi}_V^o, V]-\boldsymbol{V}^o\right|^2-\frac{\alpha_v}{2}||V||_{\mathcal{V}}^2,\\
         \text{s.t.} \quad &-\nu \Delta m+\operatorname{div}(w)=0 \text { in } \mathbb{T}^d,\\
         & \int_{\mathbb{T}^n} m(x) \mathrm{d} x=1,
\end{aligned}
\end{equation}
where $\alpha_{m^o}$, $\alpha_{v^o}$, and $\alpha_v$ are positive regularization parameters to be chosen. 
The inf problem enforces the forward MFG dynamics and data fidelity for \(m\), while the sup problem enforces data fidelity for \(V\) and ensures an appropriately regularized solution in \(\mathcal{V}\).

\subsection{A Discretized Inf-Sup Framework in 2D}  
This subsection presents a space-discretized version of \eqref{eq:inf-sup-framework} on the torus \(\mathbb{T}^2\).
Because our main interest lies in the inverse problem of recovering unknown parameters, we employ the existing
finite difference-based primal-dual method from \cite{briceno2018proximal} to solve \eqref{eq:inf-sup-framework},
leaving the development of a fully GP-based approach for future work.

Specifically, we follow the discretization scheme from \cite{briceno2018proximal}, as summarized in
Subsection~\ref{subsec:stationary}. We partition the domain \(\mathbb{T}^2\) into a toroidal grid
\(\mathbb{T}_h^2\) with uniform spacing \(h > 0\), where \(N_h = 1/h\) denotes the number of grid points
in each dimension. We denote by \(\mathcal{M}_h = \mathbb{R}^{N_h \times N_h}\) the space of discrete densities
and by \(\mathcal{W}_h = (\mathbb{R}^4)^{N_h \times N_h}\) the space of discrete flows.

The terms \(\mathcal{B}(m, w)\), \(\mathcal{F}(m)\), and the operator \(G(m, w)\) remain as defined
in Subsection~\ref{subsec:stationary}: \(\mathcal{B}(m, w)\) models the transport cost of the flow
\(w\) under the density \(m\) while enforcing admissibility of \((m, w)\),
\(\mathcal{F}(m)\) encodes density-dependent coupling effects, and
\(G(m, w)\) imposes the transportation PDE constraint and ensures mass conservation in \(m\).
Therefore, by discretizing \eqref{eq:inf-sup-framework}, we obtain
\begin{equation}
  \begin{aligned}
\inf_{m, w \in \mathcal{M}_h \times \mathcal{W}_h}\sup_{V\in \mathcal{V}}&\; \mathcal{B}(m, w) + \mathcal{F}(m) +\sum_{i,j}V(x_{i,j})m_{i,j}+\frac{\alpha_{m^o}}{2}\left|\left[\boldsymbol{\phi}^o_m, m\right]-\boldsymbol{m}^o\right|^2\\
&-\frac{\alpha_{v^o}}{2}\left|[\boldsymbol{\phi}_V^o, {V}]-\boldsymbol{V}^o\right|^2-\frac{\alpha_v}{2}||V||_{\mathcal{V}}^2\\
\text{s.t.}&\; \quad G(m, w) = (0, 1).
\label{eq:inf-sup:discre}
\end{aligned}  
\end{equation}

\subsection{A Finite-Dimensional Minimization Problem}\label{Inf-Sup framework}
The minimization problem \eqref{eq:inf-sup:discre} involves \( V \) defined in the RKHS \(\mathcal{V}\), leading to an infinite-dimensional optimization problem. To address this, following \cite{chen2021solving}, we introduce a latent vector \(\boldsymbol{v}\) and reformulate the problem as a two-level minimization. We define \(\boldsymbol{v} := (\boldsymbol{v}^{(1)}, \boldsymbol{v}^{(2)})\), where \(\boldsymbol{v}^{(1)}\) represents the values of \( V \) at the mesh grid points and \(\boldsymbol{v}^{(2)}\) corresponds to its values at the observation points. Let \(\boldsymbol{x}\) denote the vector of grid points and \(\boldsymbol{\delta}_{\boldsymbol{x}}\) the vector of Dirac measures centered at these points. Then, \eqref{eq:inf-sup:discre} is equivalent to the following problem:
\begin{equation}
  \begin{aligned}
  \inf_{m, w \in \mathcal{M}_h \times \mathcal{W}_h}\; &\mathcal{B}(m, w) + \mathcal{F}(m) + \sum_{i,j}V(x_{i,j})m_{i,j}+\frac{\alpha_{m^o}}{2}\left|\left[\boldsymbol{\phi}^o_m, m\right]-\boldsymbol{m}^o\right|^2\\
  &+ \sup_{\boldsymbol{v}}-\frac{\alpha_{v^o}}{2}|\boldsymbol{v}^{(2)}-\boldsymbol{V}^o|^2 + \begin{cases} \sup_{V\in \mathcal{V}}&
 -\frac{\alpha_v}{2}||V||_{\mathcal{V}}^2\\
 \text{s.t}&[\boldsymbol{\delta}_{\boldsymbol{x}}, V] = \boldsymbol{v}^{(1)}, [\boldsymbol{\phi}_V^o, V] = \boldsymbol{v}^{(2)}, 
  \end{cases}\\
\text{s.t.}\; &G(m, w) = (0, 1).
\label{eq:inf-sup:2l}
\end{aligned}  
\end{equation}
Let  $\boldsymbol{\psi}^V:=\left(\boldsymbol{\delta}_{\boldsymbol{x}}, \boldsymbol{\phi}_V^o\right)$. By the representer theorem \cite{owhadi2019operator}, the first-level minimization problem admits an explicit solution and yields 
\begin{align*}
V(x) = K_{\mathcal{V}}(x, {\boldsymbol{\psi}^V})K_{\mathcal{V}}(\boldsymbol{\psi}^V, {\boldsymbol{\psi}^V})^{-1}\boldsymbol{v}, 
\end{align*}
Hence, \eqref{eq:inf-sup:2l} is reduced to 
\begin{equation}
  \begin{aligned}
  \inf_{m, w \in \mathcal{M}_h \times \mathcal{W}_h}\sup_{\boldsymbol{v}}\; &\mathcal{B}(m, w) + \mathcal{F}(m) +\frac{\alpha_{m^o}}{2}\left|\left[\boldsymbol{\phi}^o_m, m\right]-\boldsymbol{m}^o\right|^2 + \sum_{i,j}\boldsymbol{v}^{(1)}_{i,j}m_{i,j}-\frac{\alpha_{v^o}}{2}|\boldsymbol{v}^{(2)}-\boldsymbol{V}^o|^2 \\
  &- \frac{\alpha_v}{2} \boldsymbol{v}^TK_{\mathcal{V}}(\boldsymbol{\psi}^V, \boldsymbol{\psi}^V)^{-1}\boldsymbol{v} \\
\text{s.t.}\; &G(m, w) = (0, 1).
\label{eq:inf-sup:finite}
\end{aligned}  
\end{equation}
Let \(\iota_{G(m,w)=(0,1)}\) be the indicator function of the set $
\bigl\{(m,w)\in \mathcal{M}_h \times \mathcal{W}_h\mid  G(m,w)=(0,1)\bigr\}.$ Denote by \(\iota_{G(m,w)=(0,1)}^*\) its Legendre transform. Then, for any dual variable \(\boldsymbol{\sigma}\), we have
$
\iota_{G(m,w)=(0,1)}(m,w)
\;=\;
\sup_{\boldsymbol{\sigma}}\langle \boldsymbol{\sigma},\,(m,w)\rangle
\;-\;
\iota_{G(m,w)=(0,1)}^*(\boldsymbol{\sigma}).
$ Then, \eqref{eq:inf-sup:finite} is equivalent to 
\begin{equation}
  \begin{aligned}
  \inf_{m, w \in \mathcal{M}_h \times \mathcal{W}_h}\sup_{\boldsymbol{\sigma}, \boldsymbol{v}}\; &\mathcal{B}(m, w) + \mathcal{F}(m) +\frac{\alpha_{m^o}}{2}\left|\left[\boldsymbol{\phi}^o_m, m\right]-\boldsymbol{m}^o\right|^2 + \sum_{i,j}\boldsymbol{v}^{(1)}_{i,j}m_{i,j}-\frac{\alpha_{v^o}}{2}|\boldsymbol{v}^{(2)}-\boldsymbol{V}^o|^2 \\
  &- \frac{\alpha_v}{2} \boldsymbol{v}^TK_{\mathcal{V}}(\boldsymbol{\psi}^V, \boldsymbol{\psi}^V)^{-1}\boldsymbol{v} + \langle \boldsymbol{\sigma}, (m, w)\rangle - \iota_{G(m, w)=(0, 1)}^*(\boldsymbol{\sigma}).
\label{eq:inf-sup:pd}
\end{aligned}  
\end{equation}
We notice that the objective function is convex in $(m, w)$ and concave in $(\boldsymbol{\sigma}, \boldsymbol{v})$. Hence, we can use the primal-dual method \cite{briceno2018proximal, chambolle2011first} to solve \eqref{eq:inf-sup:pd}.

\section{A Bilvel Framework for General Inverse Problems of Stationary Potential MFGs}
\label{sec:bilevel_framework_stationary}

When the unknown parameters are not concave in the objective function, the inf-sup minimization framework becomes inapplicable. To address this challenge, we propose a bilevel formulation in which the inner problem solves the potential MFG for given unknowns, and the outer problem measures data fidelity, regularizes the unknowns, and enforces structural constraints. The inner problem is solved via the existing primal-dual algorithm from \cite{briceno2018proximal}, which guarantees both existence and convergence. Although a GP-based potential MFG solver could be developed, our present work focuses on recovering unknown parameters, and we defer the design of such a solver to future research. To solve the bilevel problem, we propose two methods: an unrolled differentiation approach and an adjoint method. In the unrolled differentiation approach, we approximate the inner solution using the primal-dual algorithm \cite{briceno2018proximal}, then apply automatic differentiation to compute gradients for the outer objective, which are incorporated into a stochastic optimizer. The adjoint method derives the outer gradient analytically, making it independent of the inner solver's iterative accuracy and robust to the choice of the solver.

Specifically, we consider the stationary  MFG \eqref{eq:gsmfg} and the following inverse problem.
\begin{problem}
\label{st_prob}
Suppose that agents are playing in the MFG \eqref{eq:gsmfg}. Assume that given  $V^*$, ${F}^*$, $\Lambda^*$, $q^*$, and $\nu^*$, \eqref{eq:gsmfg} admits a unique classical solution $(m^*, w^*)$. Suppose that we only have some partial noisy observations on $m^*$ and on $V^*$. The objective is to infer the values of  $w^*$, $m^*$, ${V}^*$, ${F}^*$,  $\Lambda^*$, $q^*$ and $\nu^*$. To elaborate, we have 
\begin{enumerate}
    \item \textbf{Partial noisy observations on $m^*$}.  We have a set of linear operators $\{\phi_{m,l}^o\}_{l=1}^{N_m}$, $N_m\in \mathbb{N}$ where some noisy observations on $m^*$ are available. These observations are represented as $\boldsymbol{m}^o$, i.e. $\boldsymbol{m}^o = ([\phi_{m,1}^o, m^*], \dots, [\phi_{m, N_m}^o, m^*]) + \boldsymbol{\epsilon}_m$, $\boldsymbol{\epsilon}_m\sim \mathcal{N}(0, \gamma^2_m I), \gamma_m>0$. We call $\boldsymbol{\phi}^o_m=(\phi_{m,1}^o, \dots, \phi_{m, N_m}^o)$ the vector of observation operators. 
    \item \textbf{Partial noisy observations of ${V}^*$}.  We possess noisy observations at collections of  linear operators $\{\phi^{o}_{V,e}\}_{e=1}^{N_{v}}$. Denote $\boldsymbol{\phi}_V^o = (\phi_{V, 1}^o, \dots, \phi_{V, N_v}^o)$. These observations can be collected into a vector $\boldsymbol{V}^o$, 
\begin{align*}
\boldsymbol{V}^o = ([\phi^{o}_{V,1}, V^*], \dots, [\phi^{o}_{V, N_v}, V^*]) + \boldsymbol{\epsilon}_V, \boldsymbol{\epsilon}_V \sim \mathcal{N}(0, \gamma^2_VI), \gamma_V>0.  
\end{align*}
\end{enumerate}
\textbf{Inverse Problem Statement}\\
Suppose that agents are involved in the stationary MFG in \eqref{eq:gsmfg}, we infer $w^*$, $m^*$, $V^*$, ${F}^*$, $\Lambda^*$, $q^*$, and $\nu^*$, based on $\boldsymbol{m}^o$ and $\boldsymbol{V}^o$.
\end{problem}

\subsection{General Bilevel Frameworks} 
In this subsection, we propose general GP-based bilevel frameworks for solving Problem \ref{st_prob}. We approximate the spatial cost \(V\) and each entry of  \(\Lambda\) 
using GPs.
 The indices \( q \) and \( \nu \) are parameterized as Gaussian variables. To approximate the unknown coupling function \( F \) while maintaining simplicity, we assume \( F \) is local and represents a general convex function mapping \(\mathbb{R}\) to \(\mathbb{R}\). To enforce convexity, we introduce the penalization term
\begin{align*}
-\mathbb{E}_{\xi \sim \mu} \mathbb{E}_{\zeta \sim \mu} \left[\langle F'(\xi) - F'(\zeta), \xi - \zeta \rangle \right],
\end{align*}
where \( \mu \) is the probability density function of \(\xi\) and \(\zeta\). This term ensures \( F \) satisfies the convexity condition, which requires \( F'(\xi) \leq F'(\zeta) \) whenever \(\xi \leq \zeta\). The expectations are taken over \(\xi, \zeta \sim \mu\), where \( \mu \) can be Gaussian, uniform, or empirical, depending on the problem setup. For numerical implementation, \(\xi\) and \(\zeta\) can be sampled from \(\mu\), and the penalization term can be approximated using Monte Carlo integration.

Let $\mathcal{V}$ and $\mathscr{F}$ be RKHSs s.t. $V\in \mathcal{V}$ and $F\in \mathscr{F}$. 
To solve Problem \ref{st_prob}, we propose the following bilevel optimal recovery problem
\begin{equation}
  \begin{aligned}
\inf_{V, {F}, \Lambda, q, \nu} \quad & \alpha_{m^o}\left|\left[\boldsymbol{\phi}^o_m, m\right]-\boldsymbol{m}^o\right|^2+\alpha_v \|V\|_{\mathcal{V}}^2 + \alpha_{v^o}|[\boldsymbol{\phi}_V^o, V]-\boldsymbol{V}^o|^2\\&+\alpha_{f} \|{F}\|_{\mathscr{F}}^2-\alpha_{fp}E_{\xi\sim\mu}E_{\zeta\sim\mu}\left[\langle {F}'(\xi) - {F}'(\zeta), \xi - \zeta\rangle\right]+\alpha_{\lambda}|\Lambda|^2+\alpha_{q}|q|^2 + \alpha_{\nu} |\nu|^2,
\\
\text{s.t.} \quad & (m, w) \text{ solves \eqref{eq:gsmfg}},
\label{biorp}
\end{aligned}  
\end{equation}
where $\alpha_{m^o}$, $\alpha_v$, $\alpha_{v^o}$, $\alpha_f$, $\alpha_{fp}$, $\alpha_\lambda$, $\alpha_q$, $\alpha_{\nu}$ are regularization coefficients. This bilevel formulation provides a structured framework to solve inverse problems in the context of MFGs. It simultaneously estimates the unknown parameters \(V, F, \Lambda, q, \nu\) in the upper-level problem while solving the MFG system at the lower level to ensure consistency with the observed data and the mathematical model.

The bilevel optimization problem consists of several terms designed to balance fidelity to observed data, regularization of unknowns, and the mathematical structure of the MFG. The coefficient \(\alpha_m\) scales the term \(| [\boldsymbol{\phi}^o_m, m] - \boldsymbol{m}^o |^2\), which ensures that the observed density \(\boldsymbol{m}^o\) is accurately reconstructed. The term weighted by \(\alpha_V\), \(\| V \|_{\mathcal{V}}^2\), penalizes the magnitude of the potential \(V\) to promote stability and avoid overfitting. To further match observations, \(\alpha_{v^o} | [\boldsymbol{\phi}_V^o, V] - \boldsymbol{V}^o |^2\) enforces consistency between $V$  and observed data.

The function \(F\) is regularized via \(\alpha_f \| F \|_{\mathscr{F}}^2\), and its convexity is imposed by \(-\alpha_{fp} E_{\xi,\zeta \sim \mu}[\langle F'(\xi) - F'(\zeta), \xi - \zeta\rangle]\). Similarly, \(\alpha_\lambda |\Lambda|^2\) controls the magnitude of the matrix \(\Lambda\), \(\alpha_q | q |^2\) constrains the parameter \(q\), and \(\alpha_\nu |\nu|^2\) regularizes the diffusion coefficient \(\nu\). By carefully selecting these coefficients, the objective function strikes a balance between accurate data fitting, regularity of the solutions, and adherence to the underlying MFG structure.

Next, we consider three variants of \eqref{biorp} based on the level of prior knowledge about the function \(F\), each corresponding to a different estimation strategy. First, we assume \(F\) follows a power function form, \(F(m) = \frac{1}{\alpha}m^\alpha\), where \(\alpha > 0\) is an unknown exponent to be estimated. This approach extends to other parameterized convex functions where the structure of \(F\) is known, and only the parameters require estimation. Second, we generalize this idea to cases where \(F\) is believed to belong to a convex function library, represented as a combination of known convex basis functions, allowing for the estimation of multiple parameters to capture more complex coupling effects. Finally, we address the most general scenario, where \(F\) is entirely unknown apart from its convexity constraint and is approximated using either a basis expansion or a GP framework. The following subsections detail the methodologies for each of these cases.

\subsubsection{Estimation of \(F\) as a Power Function}
\label{Powerfunctionmethod}
In this case, we assume \(F\) takes the form of a power function, \(F(m) = \frac{1}{\alpha}m^\alpha\), where \(\alpha \geq 1\) is an unknown exponent to be estimated. We require \(\alpha \geq 1\) for \(F\) to remain convex. Convexity ensures the well-posedness of the optimization problem and reflects diminishing returns or stability in systems where increasing density \(m\) leads to nonlinear coupling effects. Power functions of this form are particularly useful in modeling phenomena where the coupling effect scales predictably with the density. Examples include systems with self-reinforcing growth or saturating growth.

To ensure that \(\alpha \geq 1\), we parameterize it as \(\alpha = \ln(e^{\beta} + 1) + 1\), where \(\beta\) is estimated as a Gaussian random variable. This formulation guarantees the convexity of \(F\) and allows for the estimation of \(\beta\) within a probabilistic framework. In this case, we consider the following variant of \eqref{biorp}
\begin{equation*}
  \begin{aligned}
\inf_{V, {F}, \Lambda, q, \nu} \quad & \alpha_{m^o}\left|\left[\boldsymbol{\phi}^o_m, m\right]-\boldsymbol{m}^o\right|^2+\alpha_v \|V\|_{\mathcal{V}}^2 + \alpha_{v^o}|[\boldsymbol{\phi}_V^o, V]-\boldsymbol{V}^o|^2\\&+\alpha_{\beta} |\beta|^2+\alpha_{\lambda}|\Lambda|^2+\alpha_{q}|q|^2 + \alpha_{\nu} |\nu|^2,
\\
\text{s.t.} \quad & (m, w) \text{ solves \eqref{eq:gsmfg}},
\end{aligned}  
\end{equation*}
with  
\begin{align*}
F(m(x)) + V(x)m(x) =  \frac{1}{\ln (e^{\beta}+1)+1}m(x)^{\ln (e^{\beta} + 1) + 1}+V(x)m(x),
\end{align*}
where $\alpha_\beta$ is a penalization coefficient.

\subsubsection {Estimation of \(F\) within a Convex Function Library}
\label{Function group}
When the explicit mathematical structure of \(F\) is unknown, but we assume it belongs to a convex function library formed by a combination of convex functions, we parametrize \(F\) as a weighted sum:
\[
F(m) = \sum_{k} \gamma_k f_k(m),
\]
where \( f_k(m) \) are known convex functions (e.g., \( m^\alpha, m \log m, e^m \)) from the library, and \(\gamma_k > 0\) are weights to be estimated.
This formulation captures a wide variety of coupling effects by leveraging the flexibility of combining multiple components. For example, \(m^\alpha\) can model scaling effects, \(m \log m\) can represent entropy-like behavior, and other functions can introduce additional nonlinearities. This flexibility makes it well-suited for scenarios where the coupling effect cannot be fully captured by a single functional form. The convexity of \(F(m)\) is preserved by choosing \(\gamma_k > 0\) and ensuring that each component \(f_k(m)\) is convex. 

To enforce the positivity of \(\gamma_k\) for each \(k\), we parameterize it as \(\gamma_k = \log(e^{\widetilde{\gamma}_k} + 1)\), where \(\widetilde{\gamma}_k\) is inferred. Then, 
we reformulate \eqref{biorp} as  
\begin{equation*}
  \begin{aligned}
\inf_{V, {F}, \Lambda, q, \nu} \quad & \alpha_{m^o}\left|\left[\boldsymbol{\phi}^o_m, m\right]-\boldsymbol{m}^o\right|^2+\alpha_v \|V\|_{\mathcal{V}}^2 + \alpha_{v^o}|[\boldsymbol{\phi}_V^o, V]-\boldsymbol{V}^o|^2\\&+\sum_k\alpha_{\gamma_k} |\widetilde{\gamma}_k|^2+\alpha_{\lambda}|\Lambda|^2+\alpha_{q}|q|^2 + \alpha_{\nu} |\nu|^2,
\\
\text{s.t.} \quad & (m, w) \text{ solves \eqref{eq:gsmfg}},
\end{aligned}  
\end{equation*}
with  
\begin{align*}
{F}(m(x)) + V(x)m(x) =  \sum_{k}\gamma_k f_k(m(x))+V(x)m(x). 
\end{align*}

\subsubsection{Estimation of \(F\) as a General Convex Function}\label{General Convex}
In the most general setting, we assume no prior knowledge about \(F\) beyond its convexity, providing maximum flexibility when no structural assumptions can be made. Unlike the previous cases, we cannot guarantee that \(F\) is inherently convex due to its unrestricted nature. Instead, convexity is enforced explicitly through the expectation term included in the objective function:
\[
-\alpha_{fp} E_{\xi\sim\mu}E_{\zeta\sim\mu} [\langle F'(\xi) - F'(\zeta), \xi - \zeta \rangle].
\]
This explicit enforcement of convexity guarantees that the recovered function \(F\) remains mathematically consistent with the requirements of the problem. To approximate \( F \) in this general setting, we consider two complementary strategies: one based on a basis expansion and the other using GP modeling.

In the basis expansion approach, \(F\) is represented as a linear combination of basis functions:
\[
F(m) = \sum_{k} \gamma_k \phi_k(m),
\]
where \(\{\phi_k\}\) are chosen basis functions (e.g., polynomials or random Fourier features \cite{rahimi2007random, mou2022numerical}) and \(\{\gamma_k\}\) are coefficients to be estimated. In each case, the bilevel optimization is reformulated around the finite-dimensional parameter set \(\{\gamma_k\}\), preserving computational feasibility while accommodating a wide range of functional forms for \(F\).

In this case, we estimate the unknowns $\{\gamma_k\}$ by Gaussian random variables and reformulate \eqref{biorp} as 

\begin{equation*}
  \begin{aligned}
\inf_{V, {F}, \Lambda, q, \nu} \quad & \alpha_{m^o}\left|\left[\boldsymbol{\phi}^o_m, m\right]-\boldsymbol{m}^o\right|^2+\alpha_v \|V\|_{\mathcal{V}}^2 + \alpha_{v^o}|[\boldsymbol{\phi}_V^o, V]-\boldsymbol{V}^o|^2+\sum_{k}\alpha_{\gamma_k} |\gamma_k|^2\\&-\alpha_{fp}E_{\xi\sim\mu}E_{\zeta\sim\mu}\left[\left\langle\sum_k\gamma_k\phi'_k(\xi) - \sum_k\gamma_k\phi'_k(\zeta), \xi - \zeta\right\rangle\right]+\alpha_{\lambda}|\Lambda|^2+\alpha_{q}|q|^2 + \alpha_{\nu} |\nu|^2
\\
\text{s.t.} \quad & (m, w) \text{ solves \eqref{eq:gsmfg}},
\end{aligned}  
\end{equation*}
where 
\begin{align*}
F(m(x)) + V(x)m(x) =  \sum_{k}\gamma_k \phi_k(m(x))+V(x)m(x). 
\end{align*}

In contrast, the general bilevel minimization problem in \eqref{biorp} provides a GP-based non-parametric framework where \(F\) is modeled as a random function governed by a prior distribution, removing the need to specify a fixed set of basis functions or their dimensionality.

\subsection{Discretized Bilevel Frameworks in 2D}
In this section, we propose a discretized bilevel optimization framework for solving inverse problems in two-dimensional MFGs. We focus on the space discretization of the general bilevel framework \eqref{biorp}, as other variants follow a similar approach.

Following Subsection \ref{subsec:stationary}, we first discretize the inner MFG forward problem using finite differences and, given the parameters, solve the discretized minimization problem \eqref{eq:disc:stationary}. Next, we approximate the double expectation in the convexity penalization term using the Monte Carlo method. Let \((\widetilde{r}_i)_{i=1}^{N_{\text{MC}}}\) be a set of samples drawn from the domain of \(F\) according to the distribution \(\mu\). We approximate the convexity-imposing term $-\alpha_{fp} E_{\xi\sim\mu}E_{\zeta\sim\mu} [\langle F'(\xi) - F'(\zeta), \xi - \zeta \rangle]$ by $
-\alpha_{fp} \sum_{i,j} [\langle F'(\widetilde{r}_i) - F'(\widetilde{r}_j), \widetilde{r}_i - \widetilde{r}_j \rangle]$, which 
ensures that \(F\) remains a valid convex function.

Then, the bilevel framework is formulated as follows:
\begin{equation}
  \begin{aligned}
\inf_{V, {F}, \Lambda, q, \nu} \quad & \alpha_{m^o}\left|\left[\boldsymbol{\phi}^o_m, m\right]-\boldsymbol{m}^o\right|^2+\alpha_v \|V\|_{\mathcal{V}}^2 + \alpha_{v^o}|[\boldsymbol{\phi}_V^o, V]-\boldsymbol{V}^o|^2\\&+\alpha_{f} \|{F}\|_{\mathscr{F}}^2-\alpha_{fp}\sum_{ij}\left[\langle {F}'(\widetilde{r}_i) - {F}'(\widetilde{r}_j), \widetilde{r}_i - \widetilde{r}_j\rangle\right]+\alpha_{\lambda}|\Lambda|^2+\alpha_{q}|q|^2 + \alpha_{\nu} |\nu|^2
\\
\text{s.t.} \quad & (m, w) \text{ solves \eqref{eq:disc:stationary}.}
\label{eq:disc:space:st}
\end{aligned}  
\end{equation}

In the upper-level problem, the objective is to estimate the unknown parameters \(V, F, \Lambda, q, \nu\) based on observations \(\boldsymbol{m}^o\) and \(\boldsymbol{V}^o\). The term \(\boldsymbol{\phi}^o_m\) is naturally interpreted as a functional acting on the continuous interpolation of \(m : \mathbb{T}_h^2 \to \mathbb{R}\), given that \(m\) is defined on a discrete grid.

\subsection{A Finite-Dimensional Minimization Problem}
\label{subsec:Finite_Min_Prob}

The minimization problem \eqref{eq:disc:space:st} involves variables \( V \) and \( F \) in RKHSs, resulting in an infinite-dimensional optimization. To address this issue, we propose two methods to convert \eqref{eq:disc:space:st} into a finite-dimensional minimization problem: the one-step method and the two-step method.

\subsubsection{One-step Method}
\label{subsubsec:one-step}
We reformulate \eqref{eq:disc:space:st} as a finite-dimensional minimization problem. Let \( V \) belong to a RKHS \(\mathcal{V}\) associated with the kernel $K_{\mathcal{V}}$, and denote by \(\boldsymbol{\phi}_V^o\) the vector of operators in the dual space \(\mathcal{V}^*\), representing observations of \(V\). 
Similarly, let \( F \) be a function in an RKHS \(\mathscr{F}\) associated with the kernel $K_{\mathscr{F}}$, and define \(\boldsymbol{\psi}^F = (\delta_{\widetilde{r}_1}, \dots, \delta_{\widetilde{r}_{N_{\text{MC}}}})\), and \((\widetilde{r}_i)_{i=1}^{N_{\text{MC}}}\) is a set of points used in the Monte Carlo approximation for the convexity penalization term. 
% Similarly, define $F$ as a function within the RKHS $\mathscr{F}$, with $\boldsymbol{\psi}^F=\left(\delta_{r_1}, \ldots, \delta_{r_{N_{\mathrm{F}}}}\right)$. Additionally, the set $\left(\tilde{r}_i\right)_{i=1}^{N_{\mathrm{MC}}}$ is employed in the Monte Carlo approximation for the convexity penalization term. 
We introduce latent variables \(\boldsymbol{v}\) and \(\boldsymbol{z}_F\) and reformulate \eqref{eq:disc:space:st} as the following three-level minimization problem:
\begin{equation}
  \begin{aligned}
\inf_{\boldsymbol{v}, \boldsymbol{z}_F, \Lambda, q, \nu} \quad & \begin{cases}
\inf\limits_{V\in \mathcal{V}, F\in \mathscr{F}}\alpha_{v}\|V\|^2_{\mathcal{V}} + \alpha_f\|F\|^2_{\mathscr{F}}\\
\text{s.t. }  [\boldsymbol{\delta}_{\boldsymbol{x}}, V] = \boldsymbol{v}^{(1)},  [\boldsymbol{\phi}_V^o, V] = \boldsymbol{v}^{(2)}, 
[\boldsymbol{\psi}^F, F] = \boldsymbol{z}_F
\end{cases}\\
&+ \alpha_{m^o}\left|\left[\boldsymbol{\phi}^o_m, m\right]-\boldsymbol{m}^o\right|^2 + \alpha_{v^o}|\boldsymbol{v}^{(2)}-\boldsymbol{V}^o|^2\\& -\alpha_{fp}\sum_{ij}\left[\langle F^{'}(\widetilde{r}_i)-F^{'}(\widetilde{r}_j), \widetilde{r}_i - \widetilde{r}_j\rangle\right]+\alpha_{\lambda}|\Lambda|^2+\alpha_{q}|q|^2 + \alpha_{\nu} |\nu|^2
\\
\text{s.t.} \quad & (m, w) \text{ solves \eqref{eq:disc:stationary}.}
\label{eq:biod:3l}
\end{aligned}  
\end{equation}
Let $\boldsymbol{\psi}^V=(\boldsymbol{\delta}_{\boldsymbol{x}},  \boldsymbol{\phi}_V^o)$. The first-level minimization problem admits an explicit solution and yields 
\begin{align}
\label{eq:repre:VF}
V(x) = K_{\mathcal{V}}(x, {\boldsymbol{\psi}^V})K_{\mathcal{V}}(\boldsymbol{\psi}^V, {\boldsymbol{\psi}^V})^{-1}\boldsymbol{v} \text{ and } F(x) = K_{\mathscr{F}}(x, {\boldsymbol{\psi}^F})K_{\mathscr{F}}(\boldsymbol{\psi}^F, {\boldsymbol{\psi}^F})^{-1}\boldsymbol{z}_F. 
\end{align}
Hence, \eqref{eq:biod:3l} is reduced to 
\begin{equation}
  \begin{aligned}
\inf_{\boldsymbol{v}, \boldsymbol{z}_F, \Lambda, q, \nu} \quad & \alpha_v\boldsymbol{v}^TK_{\mathcal{V}}(\boldsymbol{\psi}^V, \boldsymbol{\psi}^V)^{-1}\boldsymbol{v} + \alpha_{f}\boldsymbol{z}_F^TK_{\mathscr{F}}(\boldsymbol{\psi}^F, \boldsymbol{\psi}^F)^{-1}\boldsymbol{z}_F  + \alpha_{m^o}\left|\left[\boldsymbol{\phi}^o_m, m\right]-\boldsymbol{m}^o\right|^2\\& + \alpha_{v^o}|\boldsymbol{v}^{(2)}-\boldsymbol{V}^o|^2 -\alpha_{fp}\sum_{ij}\left[\langle F^{'}(\widetilde{r}_i)-F^{'}(\widetilde{r}_j), \widetilde{r}_i - r_j\rangle\right]+\alpha_{\lambda}|\Lambda|^2+\alpha_{q}|q|^2 + \alpha_{\nu} |\nu|^2
\\
\text{s.t.} \quad & (m, w) \text{ solves \eqref{eq:disc:stationary}.}
\label{eq:biod:finite}
\end{aligned}  
\end{equation}
\subsubsection{Two-step Method}
\label{subsubsec:two-step}
In the one-step method, the dimension of the latent variable \(\boldsymbol{z}_F\) increases with the number of Monte Carlo samples used to approximate the expectations in the convex penalization term. Consequently, improving the accuracy of the Monte Carlo approximation by increasing the number of samples also increases the complexity of the optimization. To address this issue, we introduce a set of pseudo points to approximate \(F\), thereby reducing the number of parameters to optimize. Specifically, we select \(N_F \in \mathbb{N}\) pseudo points \(\{r_i\}_{i=1}^{N_F}\) within a sufficiently large domain in \(\mathbb{R}\) that covers the range of the density \(m\). Let \(\boldsymbol{z}_F\) denote the values of \(F\) at these pseudo points, which are optimized in the second step. We approximate \(F\) by the posterior mean of a Gaussian process \(\xi\sim\mathcal{GP}(0, K_{\mathscr{F}})\) conditioned on the observations \(\boldsymbol{z}_F\), i.e., \(F = \mathbb{E}[\xi \mid \xi(r_i) = \boldsymbol{z}_{F,i}, i=1,\dots, N_F]\) and define \(\boldsymbol{\psi}^F = (\delta_{r_1}, \dots, \delta_{r_{N_{\text{F}}}})\). In the second step, we draw a new bigger set of $N_{\text{MC}} \in \mathbb{N}$ samples \(\{\widetilde{r}_i\}_{i=1}^{N_{\text{MC}}}\)  to approximate the expectation in the convex penalization term and formulate the following optimization problem:
\begin{equation}
  \begin{aligned}
\inf_{\boldsymbol{v}, \boldsymbol{z}_F, \Lambda, q, \nu} \quad & \alpha_v\boldsymbol{v}^TK_{\mathcal{V}}(\boldsymbol{\psi}^V, \boldsymbol{\psi}^V)^{-1}\boldsymbol{v} + \alpha_{f}\boldsymbol{z}_F^TK_{\mathscr{F}}(\boldsymbol{\psi}^F, \boldsymbol{\psi}^F)^{-1}\boldsymbol{z}_F  + \alpha_{m^o}\left|\left[\boldsymbol{\phi}^o_m, m\right]-\boldsymbol{m}^o\right|^2\\& + \alpha_{v^o}|\boldsymbol{v}^{(2)}-\boldsymbol{V}^o|^2 -\alpha_{fp}\sum_{ij}\left[\langle F^{'}(\widetilde{r}_i)-F^{'}(\widetilde{r}_j), \widetilde{r}_i - \widetilde{r}_j\rangle\right]+\alpha_{\lambda}|\Lambda|^2+\alpha_{q}|q|^2 + \alpha_{\nu} |\nu|^2
\\
\text{s.t.} \quad & (m, w) \text{ solves \eqref{eq:disc:stationary}.}
\label{eq:biod:finite:two-step}
\end{aligned}  
\end{equation}
Hence, in \eqref{eq:biod:finite:two-step}, the size of \(\boldsymbol{z}_F\) depends only on the number of selected pseudo points and is independent of the accuracy of the Monte Carlo approximations.

To solve the minimization problems \eqref{eq:biod:finite} and \eqref{eq:biod:finite:two-step}, we adopt two methods: an unrolled differentiation approach and an adjoint method. In the unrolled differentiation approach, the inner-level minimization problem is first solved to approximate its solution as a function of the outer variables. This is done by implementing a dedicated subprogram for inner-level optimization. Next, automatic differentiation is used to compute the gradients of the outer-level objective with respect to the outer variables. These gradients are then incorporated into a stochastic optimization framework, such as Adam \cite{kingma2014adam}, to iteratively update the outer variables and minimize the outer objective.

In Section \ref{sec:adjoint_method}, we introduce an adjoint method to compute the gradient of the outer-level minimization problem. Unlike the auto-differentiation approach, the adjoint method derives the gradient analytically, offering two key advantages. First, the gradient computation is more accurate, as it is independent of the accuracy of the iterative solver for the inner problem.
 Second, the adjoint method is independent of the specific solver used for the inner minimization problem, making it robust and versatile for a wide range of applications.

\subsection{A Probabilistic Perspective}
 % Here, we give a probabilistic interpretation of the GP method for the potential MFG inverse problems. First, we put independent Gaussian priors on the functions $\left(V, F, Q, q\right)$ to \ref{Potentialstationary} such that $V \sim \mathcal{N}\left(0, K_V\right), F \sim \mathcal{N}\left(0, K_{f_z}\right), Q_{ij} \sim \mathcal{N}(0,1)$, and $q \sim \mathcal{N}\left(0, 1\right)$, where $K_V, K_{f_z}$ are kernels associated with the RKHSs $V, F$. Let $\phi^V$ be the vectors of linear operators defined in Subsection \ref{First typeV} and suppose $\boldsymbol{\delta}^{\vec{r}}=\left(\delta_{r_1}, \ldots, \delta_{r_{N_f}}\right)$. Let $\boldsymbol{v} $, and $\boldsymbol{f_z}$ be defined as

We propose a probabilistic interpretation of the GP framework for solving potential MFG inverse problems. This perspective enables the integration of prior knowledge about the functions \( V, F, \Lambda, q, \nu \) within a probabilistic framework, combining data-driven inference with regularization to ensure the well-posedness of the problem. Independent Gaussian priors are assigned to unknown functions and variables, such that \( V \sim \mathcal{GP}(0, K_{\mathcal{V}}) \), \( F \sim \mathcal{GP}(0, K_{\mathscr{F}}) \), \( \Lambda_{ij} \sim \mathcal{N}(0, 1) \), \( q \sim \mathcal{N}(0, 1) \), and \( \nu \sim \mathcal{N}(0, 1) \), where \( K_{\mathcal{
V}} \) and \( K_{\mathscr{F}} \) are kernels associated with the RKHSs of \( V \) and \( F \). 

Let \( \boldsymbol{\psi}^V \) and $\boldsymbol{\psi}^F$ denote the vectors of linear operators defined in Subsection \ref{subsec:Finite_Min_Prob}. Denote $\boldsymbol{v} := [\boldsymbol{\psi}^V, V]$ and $\boldsymbol{z}_F := [\boldsymbol{\psi}^F, F]$. Then,  the densities of \( \boldsymbol{v} \) and \( \boldsymbol{z}_F \) are given as:
\[
p(\boldsymbol{v}) = \frac{1}{C_V \sqrt{\det(K_{\mathcal{V}}(\boldsymbol{\psi}^V, \boldsymbol{\psi}^V))}} e^{-\frac{1}{2} \boldsymbol{v}^T K_{\mathcal{V}}(\boldsymbol{\psi}^V, \boldsymbol{\psi}^V)^{-1} \boldsymbol{v}},
\]
\[
p(\boldsymbol{z}_F) = \frac{1}{C_F \sqrt{\det(K_{\mathscr{F}}(\boldsymbol{\psi}^F, \boldsymbol{\psi}^F))}} e^{-\frac{1}{2} \boldsymbol{z}_F^T K_{\mathscr{F}}(\boldsymbol{\psi}^F, \boldsymbol{\psi}^F)^{-1} \boldsymbol{z}_F},
\]
where \( C_V \) and \( C_F \) are normalization constants ensuring unit mass of the probability densities.

To account for uncertainty in the observation of the potential function \( V \), the noise is modeled as \( \epsilon_V = \boldsymbol{v}^{(2)} - \boldsymbol{V}^o \), where \( \epsilon_V \sim \mathcal{N}(0, \gamma^2_V I) \). Similarly, the term \( \gamma^{-1}_m(\left[\boldsymbol{\phi}^o_m, m\right]-\boldsymbol{m}^o) \sim \mathcal{N}(0, I) \) represents uncertainty in the observations of population density. This probabilistic treatment of noise accounts for uncertainty in the observations, which is important for robustness against measurement errors. 

The potential MFG inverse problem is reformulated within this framework by combining the prior distributions, the likelihood of the observations, and regularization terms such as convexity enforcement. Hence, the inverse problem seeks to solve 
\begin{align*}
\min_{\boldsymbol{v}, \boldsymbol{z}_F, \Lambda, q, \nu} \;& -\alpha_v\log p(v) - \alpha_f\log p(\boldsymbol{z}_F) - \alpha_\lambda\sum_{ij} \log p(\Lambda_{ij}) - \alpha_q\log p(q) - \alpha_\nu\log p(\nu)\\
&-\alpha_{v^o}\log p\left(\gamma^{-1}_V(\boldsymbol{v}^{(2)} - \boldsymbol{V}^o)\right)- \alpha_{m^o}\log p\left(\gamma^{-1}_m(\left[\boldsymbol{\phi}^o_m, m\right]-\boldsymbol{m}^o)\right)\\
&
- \alpha_{fp}E_{\xi\sim\mu}E_{\zeta\sim\mu}\left[\langle F^{'}(\xi)-F^{'}(\zeta), \xi - \zeta\rangle\right],\\&
\text{s.t.} \quad  (m, w) \text{ solves \eqref{eq:disc:stationary}},
\end{align*}
which is equivalent to \eqref{eq:biod:finite}. 

This formulation integrates the observed data with prior knowledge while ensuring that the solution adheres to convexity constraints. By introducing Gaussian priors, the probabilistic framework highlights the potential to quantify uncertainty.

\section{Time-Dependent Potential MFG Inverse Problems}
\label{sec:time_dependent}
This section addresses the time-dependent inverse problem, Problem \ref{main_prob}, and extends the bilevel framework introduced in Section \ref{sec:bilevel_framework_stationary} to handle this setting. While the inf-sup approach could also be adapted for time-dependent inverse problems, we omit its formulation here to maintain clarity. Instead, we focus on the bilevel method, which can handle a broader range of potential MFGs.

\subsection{A Time Discretization-Based Bilevel Framework} In this subsection, we propose a bilevel framework for inverse problems of time-dependent potential MFGs.

We begin by dividing the interval $[0,T]$ into a uniform grid $t_0=0<t_1<\dots<t_{N_T}=T$. 
At each time step $t_k$, we introduce discrete approximations 
$m^k \approx m(t_k)$, $V_k \approx V(t_k)$, and ${F}_k \approx {F}\bigl(t_k, \cdot\bigr)$. 
We also model the entries \( \Lambda_{ij} \) of the matrix \( \Lambda \) as Gaussian variables and represent each  $F_k$ and $V_k$ using a GP. Specifically, we assume that for each \(k\), there exist RKHSs \(\mathcal{V}_k\) and \(\mathscr{F}_k\) with corresponding kernels \(K_{\mathcal{V}_k}\) and \(K_{\mathscr{F}_k}\), respectively, such that \(V_k \in \mathcal{V}_k\) and \(F_k \in \mathscr{F}_k\).   Likewise, the two unknown parameters $q$ and $\nu$ are modeled as Gaussian random variables.

Using this setup, we discretize the MFG PDE so that each pair $(m^k, w^k)$ must satisfy 
\[
   \frac{m^{k+1} - m^k}{\Delta t} \;-\;\nu\,\Delta m^{k+1}
   \;+\;\mathrm{div}\bigl(w^{k+1}\bigr) \;=\; 0
   \quad\text{on }\mathbb{T}^d,\quad k=0,\dots,N_T-1,
\]
with the initial condition $m^0(x) = m_0(x)$. Let \(\boldsymbol{m}_{k}^o\) and \(\boldsymbol{V}_k^o\) be the observed data for \(m^k\) and \(V_k\), respectively, and let \(\boldsymbol{\phi}_k^o\) and \(\boldsymbol{\phi}_{V_k}^o\) denote the corresponding observation operators for \(m^k\) and \(V_k\). 
We then embed these time-discretized PDE constraints into a bilevel optimization framework: 
\begin{equation}
  \begin{aligned}
&\inf_{\left(V_k, {F}_k, \Lambda,q,\nu\right)}\sum_{k=0}^{N_T}\bigg(\alpha_{m^o,k}|[\boldsymbol{\phi}^o_k, m^k]-\boldsymbol{m}^o_k|^2+\alpha_{v,k} \|V_k\|_{\mathcal{V}_k}^2 + \alpha_{v^o,k}|[\boldsymbol{\phi}_{V_k}^o, V_k]-\boldsymbol{V}^o_k|^2+\alpha_{f,k} \|{F}_k\|_{\mathscr{{F}}_k}^2\\&\quad\quad\quad\quad-\alpha_{fp,k}E_{\xi\sim\mu}E_{\zeta\sim\mu}\left[\langle {F_k}'(\xi) - {F_k}'(\zeta), \xi - \zeta\rangle\right]\bigg)+\alpha_{\lambda}|\Lambda|^2+\alpha_q|q|^2 + \alpha_\nu |\nu|^2
\\
&\text{s.t. } (m^k, w^k)_{k=0}^{N_T} \text{ solve } \begin{cases}
\min\limits_{m^k, w^k} \Delta t\sum\limits_{k=1}^{N_T}\int_{\mathbb{T}^d} b(m^k, w^k) + V_km^k + F_k(m^k)\dif x + \int_{\mathbb{T}^d}\mathcal{G}(x, m^{N_T})\dif x  \\
\text{s.t.} \quad\; \frac{m^{k+1} - m^k}{\Delta t} - \nu\Delta m^{k+1} + \div(w^{k+1}) = 0 \text{ on } \mathbb{T}^d \text{ for } k=0,\dots, N_T - 1, \\
 \quad\quad\quad m^0(x) = m_0(x) \text{ on } \mathbb{T}^d. 
\end{cases}
\label{eq:biorpt}
\end{aligned}  
\end{equation}

In the lower-level subproblem, we solve the time-discretized PDE-constrained minimization problem. 
Meanwhile, in the upper-level problem, we seek the parameters $\{V_k,F_k,\Lambda,q,\nu\}$ 
that achieve the best fit to observed data (through terms like 
$\alpha_{m^o,k}|\left[\boldsymbol{\phi}^o_k, m^k\right]-\boldsymbol{m}^o_k|^2$), 
enforce convexity in $F_k$ via expressions such as 
$-\alpha_{fp,k}E_{\xi\sim\mu}E_{\zeta\sim\mu}\left[\langle {F_k}'(\xi) - {F_k}'(\zeta), \xi - \zeta\rangle\right]$, 
and penalize large parameter norms (e.g.\ $|\Lambda|^2$, $|\nu|^2$). This design allows the method to learn unknown functions through GP approximations, 
respect the discrete-time PDE constraints at each step and control model complexity 
with regularization.

\subsection{A Space-Time Discretized Bilevel Framework in 2D}
In this section, we propose a space--time discretized bilevel optimization framework for solving
\eqref{eq:biorpt} in two dimensions. Let \(0 = t_0 < t_1 < \dots < t_{N_T} = T\) be a partition
of the time interval \([0,T]\) with \(\Delta t = t_{k+1} - t_k\). Let \(\{x_{i,j}\}\) be a spatial 
mesh on \(\mathbb{T}^2\). We denote by \(m^k_{i,j} \approx m(x_{i,j},t_k)\) and 
\(w^k_{i,j} \approx w(x_{i,j},t_k)\), collecting them into the arrays
\[
m = \bigl\{ m^k_{i,j} \bigr\}, 
\quad
w = \bigl\{ w^k_{i,j} \bigr\}.
\]
Define \(\mathcal{M}\) as the space of these discrete density arrays and \(\mathcal{W}\) as the 
space of the discrete flux arrays. Recalling the discrete time-dependent MFG formulation 
\eqref{eq:time-dependent:discr}, we now consider the following space--time discretization 
of \eqref{eq:biorpt}:
\begin{equation}
  \begin{aligned}
&\inf_{\left(V_k, {F}_k, \Lambda,q,\nu\right)}\sum_{k=0}^{N_T}\bigg(\alpha_{m^o,k}|[\boldsymbol{\phi}^o_k, m^k]-\boldsymbol{m}^o_k|^2+\alpha_{v,k} \|V_k\|_{\mathcal{V}_k}^2 + \alpha_{v^o,k}|[\boldsymbol{\phi}_{V_k}^o, V_k]-\boldsymbol{V}^o_k|^2+\alpha_{f,k} \|{F}_k\|_{\mathscr{{F}}_k}^2\\&\quad\quad\quad\quad-\alpha_{fp,k}E_{\xi\sim\mu}E_{\zeta\sim\mu}\left[\langle {F_k}'(\xi) - {F_k}'(\zeta), \xi - \zeta\rangle\right]\bigg)+\alpha_{\lambda}|\Lambda|^2+\alpha_q|q|^2 + \alpha_\nu |\nu|^2\\
&\text{s.t. }(m, w) \text{ solves \eqref{eq:time-dependent:discr}.}
\label{eq:biorpt:sdisc}
\end{aligned}  
\end{equation}
\subsection{A Finite-Dimensional Minimization Problem} 
\label{Time dependent Finite}
In \eqref{eq:biorpt:sdisc}, we minimize functions in RKHSs, giving rise to an infinite-dimensional optimization problem. To address this, we employ a two-step method similar to the stationary MFG setting in Subsection~\ref{subsubsec:two-step} that introduces latent variables and thereby reduces the problem to a finite-dimensional formulation. As before, we use pseudo points to approximate \(F_k\) and rely on Monte Carlo sampling for the convex penalization term.

We select \(N_{F_k}\in \mathbb{N}\) pseudo points \(\{r_i^k\}_{i=1}^{N_{F_k}}\) within a domain large enough to cover the range of \(m^k\). Denote by \(\boldsymbol{z}_{F_k}\) the values of \(F_k\) at these pseudo points; we then approximate \(F_k\) by the posterior mean of a GP \(\xi_k \sim \mathcal{GP}(0, K_{\mathscr{F}_k})\) conditioned on \(\boldsymbol{z}_{F_k}\), i.e.\ 
$F_k = \mathbb{E}\bigl[\xi_k \,\bigm\vert\, \xi_k\bigl(r_i^k\bigr) = \boldsymbol{z}_{F_{k}, i}\bigr],$
and let \(\boldsymbol{\psi}^{F}_k = (\delta_{r_1^k}, \dots, \delta_{r_{N_{F_k}}^k})\). We draw a new larger set of \(N_{\mathrm{MC}}^k \in \mathbb{N}\) samples \(\{\widetilde{r}_i^k\}_{i=1}^{N_{\mathrm{MC}}^k}\) for approximating the expectation in the convex penalization term. As in Subsection~\ref{subsec:Finite_Min_Prob}, we introduce latent vectors \(\boldsymbol{v}_k\) to represent \(V_k\) on the grid and observation points, with \(\boldsymbol{\psi}^V_k\) denoting the corresponding operator.  Under these approximations, \eqref{eq:biorpt:sdisc} is reformulated as the following finite-dimensional optimization problem:
\begin{equation}
  \begin{aligned}
\inf_{\left(\boldsymbol{v}_k, {\boldsymbol{z}_{F,k}}, \Lambda,q,\nu\right)}&\sum_{k=0}^{N_T}\bigg(\alpha_{m^o,k}|[\boldsymbol{\phi}^o_k, m^k]-\boldsymbol{m}^o_k|^2+ \alpha_{v^o,k}|[\boldsymbol{\phi}_{V_k}^o, V_k]-\boldsymbol{V}^o_k|^2\\&\quad-\alpha_{fp,k}\sum_{ij}[\langle {F_k}'(\widetilde{r}_i^k) - {F_k}'(\widetilde{r}_j^k), \widetilde{r}_i^k - \widetilde{r}_j^k\rangle]+\alpha_{v,k}\boldsymbol{v}_k^TK_{\mathcal{V}_k}(\boldsymbol{\psi}^V_k, \boldsymbol{\psi}^V_k)^{-1}\boldsymbol{v}_k \\
&\quad+ \alpha_{f,k}\boldsymbol{z}_{F,k}^TK_{\mathscr{F}_k}(\boldsymbol{\psi}^F_k, \boldsymbol{\psi}^F_k)^{-1}\boldsymbol{z}_{F,k}
\bigg)+\alpha_{\lambda}|\Lambda|^2+\alpha_q|q|^2 + \alpha_\nu |\nu|^2\\
&\text{s.t. }(m, w) \text{ solves \eqref{eq:time-dependent:discr}.}\label{eq:biorpt:sdisctwostep}
\end{aligned}  
\end{equation}
Following a similar strategy as in the stationary case, we employ an unrolled differentiation method procedure to solve \eqref{eq:biorpt:sdisc} and \eqref{eq:biorpt:sdisctwostep}. First, we solve the inner-level problem to obtain its solution as a function of the outer variables. We then compute the gradient of the outer-level objective, either via automatic differentiation or through the adjoint method introduced in Section \ref{sec:adjoint_method}. The adjoint approach provides an analytic gradient that does not depend on the accuracy of the inner solver and is solver-agnostic, making it particularly robust. We subsequently apply a stochastic optimizer (e.g., Adam) to update the outer variables.

\section{The Adjoint Method}
\label{sec:adjoint_method}
\noindent
As discussed previously, one way to solve the bilevel minimization problems \(\eqref{eq:biod:finite:two-step}\) and \(\eqref{eq:biorpt:sdisctwostep}\) is via an unrolled differentiation method. First, the inner-level problem is solved to express its solution as a function of the outer variables. Second, the gradient of the outer-level objective is computed by automatic differentiation, and a stochastic optimizer (e.g., Adam) is used to update the outer variables. Although straightforward, this method requires differentiating through an iterative inner solver. Increasing the number of inner iterations improves accuracy but substantially raises the differentiation cost, whereas fewer iterations reduce that cost at the expense of inner-solution accuracy.

We employ the adjoint method to compute the gradient of the outer objective with respect to the outer parameters, treating the bilevel minimization problems \eqref{eq:biod:finite:two-step} and \eqref{eq:biorpt:sdisctwostep} in an abstract Euclidean setting. The bilevel problem can be written as
\[
\min_{\theta} \, J(\theta)
\;=\;
\min_{\theta} \, Q\bigl(v(\theta), \theta\bigr),
\]
where \(v(\theta)\) is obtained from the inner minimization
\[
v(\theta)
\;=\;
\arg\min_{v}\,\mathcal{L}(v, \theta)
\quad
\text{subject to}
\quad
P(v,\theta)=0,
\]
and \(P(v,\theta) = 0\) represents a PDE constraint. 

For instance, in \eqref{eq:biod:finite:two-step}, the parameter vector \(\theta\) may include the unknowns \(\bigl(\boldsymbol{v}, \boldsymbol{z}_F, \Lambda, q, \nu\bigr)\), while \(J(\theta)\) is the outer objective function. The solution \(v(\theta)\), viewed as \((m, w)\) in a potential MFG problem, is determined by minimizing \(\mathcal{L}\), where \(\mathcal{L}(\theta) = \mathcal{B}(m, w) + \mathcal{F}(m)+\sum_{i,j}V(x_{i,j})m_{i,j}\), and the constraint \(P(v,\theta)=0\) encodes, for example, the  system \(G(m,w)=0\). Because \(J(\theta)\) depends on the solution \(v(\theta)\), gradient computations become nontrivial. The adjoint method offers an efficient mechanism to compute these gradients by systematically incorporating the PDE constraint into the differentiation process.

A direct application of the chain rule to compute \(\frac{dJ}{d\theta}\) yields
\[
\label{eq:dJdtheta}
\frac{dJ}{d\theta}
\;=\;
\frac{dv^\top}{d\theta}\,\frac{\partial Q}{\partial v}\bigl(v(\theta),\theta\bigr)
\;+\;
\frac{\partial Q}{\partial \theta}\bigl(v(\theta),\theta\bigr).
\]
Here, \(\tfrac{dv^\top}{d\theta}\) arises from the dependence of \(v\) on \(\theta\). Computing \(\tfrac{dv}{d\theta}\) explicitly is often prohibitive in large-scale settings, especially when \(v\) is a high-dimensional PDE solution. Instead, we use the stationarity condition for the inner problem:
\[
\frac{\partial \mathcal{L}}{\partial v}(v(\theta),\theta)
\;+\;
\Bigl(\frac{\partial P}{\partial v}(v(\theta),\theta)\Bigr)^\top \lambda
\;=\;
0,
\]
where \(\lambda\) is the Lagrange multiplier enforcing \(P(v,\theta)=0\). Differentiating with respect to \(\theta\) and isolating \(\tfrac{dv}{d\theta}\) gives
\[
\left(
\frac{\partial^2 \mathcal{L}}{\partial v\,\partial v}
\;+\;
\left(\frac{\partial^2 P}{\partial v\,\partial v}\right)^\top \!\lambda
\right)
\frac{dv}{d\theta}
\;=\;
-\,
\left(
\frac{\partial^2 \mathcal{L}}{\partial v\,\partial \theta}
\;+\;
\left(\frac{\partial^2 P}{\partial v\,\partial \theta}\right)^\top \!\lambda
\right).
\]
To avoid explicitly solving for \(\tfrac{dv}{d\theta}\), we introduce an adjoint vector \(p\) satisfying
\[
\left(
\frac{\partial^2 \mathcal{L}}{\partial v\,\partial v}
\;+\;
\left(\frac{\partial^2 P}{\partial v\,\partial v}\right)^\top \!\lambda
\right)^\top
p
\;=\;
-\,\frac{\partial Q}{\partial v}\bigl(v(\theta),\theta\bigr).
\]
Substituting this adjoint relation into the chain-rule expression eliminates \(\tfrac{dv}{d\theta}\), yielding
\[
\label{eq:final_derivative}
\frac{dJ}{d\theta}
\;=\;
\left(
\frac{\partial^2 \mathcal{L}}{\partial v\,\partial \theta}
\;+\;
\left(\frac{\partial^2 P}{\partial v\,\partial \theta}\right)^\top \!\lambda
\right)^\top
p
\;+\;
\frac{\partial Q}{\partial \theta}\bigl(v(\theta),\theta\bigr).
\]
The final expression  efficiently computes \(\frac{dJ}{d\theta} \), avoiding the need to compute \(\frac{dv}{d\theta} \). This makes the adjoint method highly suitable for large-scale optimization problems and can be extended to other lower-level optimization algorithms for inverse problems besides the primal-dual algorithm.

% Then we can update the 
% $$
% \mathbf{\theta}_{k+1}=\theta_k-\tilde{\alpha}\frac{dJ}{d\theta}|_{\theta_k}
% $$

\section{Numerical Experiments}
\label{sec:Experiment}
In this section, we present numerical experiments on a variety of potential MFG inverse problems to validate our proposed frameworks. These experiments focus on reconstructing entire population profiles, agent strategies, and environmental configurations from partial, noisy data on both populations and environments. In Subsection~\ref{subsec:num:inf-sup}, we highlight the effectiveness of the inf-sup framework. Subsections~\ref{subsec:num:bilevel:VF} explore identifying the spatial cost and coupling functions in potential MFGs using the bilevel approach under different prior assumptions. Subsection~\ref{subsec:num:LambdaVF} considers recovering all unknowns simultaneously, including an Euclidean metric for kinetic energy, while Subsection~\ref{subsec:num:NLambdaVF} addresses a non-Euclidean metric. Finally, Subsection~\ref{subsec:num:td} applies our methods to a time-dependent MFG scenario.

We test two methods for solving the bilevel minimization problem. The first relies on automatic differentiation, implemented in Python with PyTorch. The second uses the adjoint-based approach from Section~\ref{sec:adjoint_method}, computing gradients independently of the solver used for the inner minimization problem. We adopt a coordinate descent strategy for training all unknown parameters in the  inverse problem recovering all unkowns. For instance, in the stationary MFG setting where \(V\), \(F\), \(q\), \(\Lambda\), and \(\nu\) are jointly recovered, we first mainly perform gradient-based updates on \(V\) (which has the largest parameter dimension and deviation) and then update the remaining parameters together. This partitioned approach helps manage the computational complexity and enhances convergence stability. In the experiments, by assigning infinite values to the penalization parameters, we effectively transform the corresponding constraints into hard constraints that are enforced exactly.

In the bilevel framework, we employ the Chambolle--Pock splitting (CP) algorithm \cite{chambolle2011first, briceno2018proximal, briceno2019implementation} to solve the inner minimization problem. This choice is motivated by its superior speed and accuracy compared to other methods for potential MFGs, as reported in \cite{briceno2018proximal, briceno2019implementation}.

% Chambolle-Pock's (CP) splitting algorithm \cite{chambolle2011first} is an efficient iterative primal-dual algorithm used to solve convex optimization problems, particularly effective for problems involving total variation minimization. We will use the CP algorithm to solve the lower-level optimization problem for both the inverse problem of stationary and time-dependent potential MFGs for all the experiments similar to the papers \cite{briceno2018proximal, briceno2019implementation}.

All experiments measure discrepancies between the recovered density and a reference density using a discretized \(L^2\) norm. Specifically, let \(u\) and \(v\) be functions on \([a, b]^2\). We discretize this domain with grid sizes \(h_x\) and \(h_y\) along the \(x\)- and \(y\)-axes, respectively, forming arrays \(\{u_{ij}\}\) and \(\{v_{ij}\}\). The discretized \(L^2\) error is given by
\begin{equation}
\label{eq:l2disc}
\mathcal{E}(u, v) 
\;=\; 
\sqrt{\,h_x\,h_y\,\sum_{i,j}\,\bigl|\,u_{ij} - v_{ij}\bigr|^2}\,.
\end{equation}
% \begin{align}
% \label{eq:l2disc}
% \mathcal{E}(u, v) = \sqrt{h^2 \sum_{i,j} |u_{ij} - v_{ij}|^2}.
% \end{align}

Although these inverse problems can be ill-posed when insufficient prior knowledge is available, our frameworks consistently yield surrogate MFG models that closely match distribution data, offering insights into underlying dynamics and supporting practical tasks such as forecasting and scenario analysis.

\subsection{Recovering  Spatial Costs Using The Inf-Sup Framework.} 
\label{subsec:num:inf-sup}
This experiment investigates a first-order stationary MFG system defined as follows:
\begin{equation}
\begin{cases}
\frac{1}{2}|\nabla u(x, y)|^2 - \lambda = \log m(x, y) + V(x, y), & \forall (x, y) \in \mathbb{T}^2, \\
\operatorname{div}(m(x, y) \nabla u(x, y)) = 0, & \forall (x, y) \in \mathbb{T}^2, \\
\int_{\mathbb{T}^2} m(x, y) \, dx \, dy = 1, \quad \int_{\mathbb{T}^2} u(x, y) \, dx \, dy = 0,
\end{cases}
\label{eq:MFG_system}
\end{equation}
where \begin{equation*}
V(x, y) = -\sin(2\pi x) - \sin(2\pi y).
\end{equation*}
The explicit solution to the above system is given by:
\begin{equation}
\label{eq:rec:pot:explicit}
u^*(x, y) = 0, \quad m^*(x, y) = e^{-V(x, y) - \lambda^*}, \quad 
\lambda^* = \log \left( \int_{\mathbb{T}^2} e^{-V(x, y)} \, dx \, dy \right).
\end{equation}
The objective is to reconstruct the MFG distribution \( m \) and the function \(V\) from partial observations of both \(m\) and \(V\), employing the inf-sup framework proposed in Section~\ref{sec:inf_sup_framework}.

\textbf{Experimental Setup.} In this experiment, the spatial domain is discretized with a grid spacing of \( h = \frac{1}{50} \) along both the \(x\)- and \(y\)-axes.
 For observations, 39 points are selected from the total 2,500 sample points of \(m\). Meanwhile, 36 observation points for \(V\) are randomly generated in space and are not restricted to grid points. The regularization parameters are set to \( \alpha_{m^o} = 10^5 \), $a_v=2$, and \( \alpha_{v^o} = \infty \), where \( \alpha_{v^o} = \infty \) ensures that the exact values of \( V \) are imposed at the observation points.
 Gaussian noise \( \mathcal{N}(0, \gamma^2 I) \) with a standard deviation \( \gamma = 10^{-3} \) is added to the observations. 
 To approximate \( V \), a GP with a periodic kernel is employed in all subsequent experiments.
%  :
% \begin{equation*}
% K(x, y; \sigma) = \exp\left( -\frac{\|x - y\|^2}{2 \sigma^2} \right),
% \end{equation*}
% where \( \sigma = 0.6 \). 
% Reducing the grid size and increasing observation points improves parameter accuracy and reduces \( m \)'s recovery error.

\begin{figure}[h]
    \centering
    \begin{subfigure}{0.33\textwidth}
        \centering
        \includegraphics[width=\linewidth]{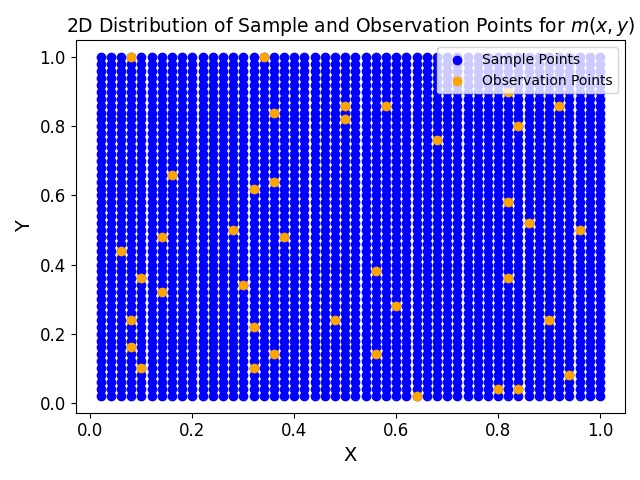}
        \caption{Samples \& Observations of $m$}
        \label{fig:msample1}
    \end{subfigure}%
    \begin{subfigure}{0.33\textwidth}
        \centering
        \includegraphics[width=\linewidth]{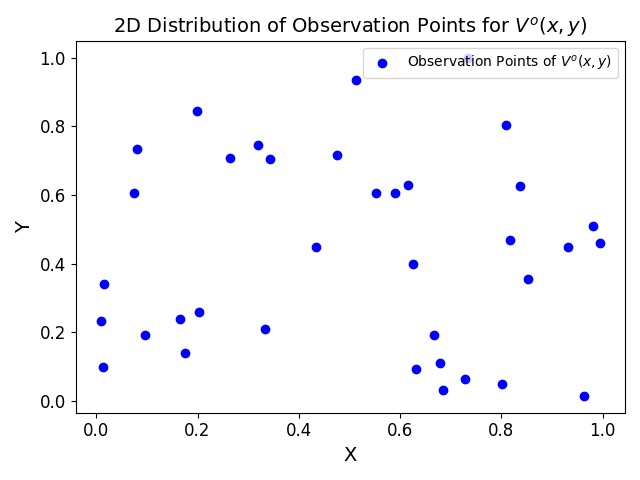}
        \caption{Observations of $V$}
        \label{fig:v01}
    \end{subfigure}%
    \begin{subfigure}{0.33\textwidth}
        \centering
        \includegraphics[width=\linewidth]{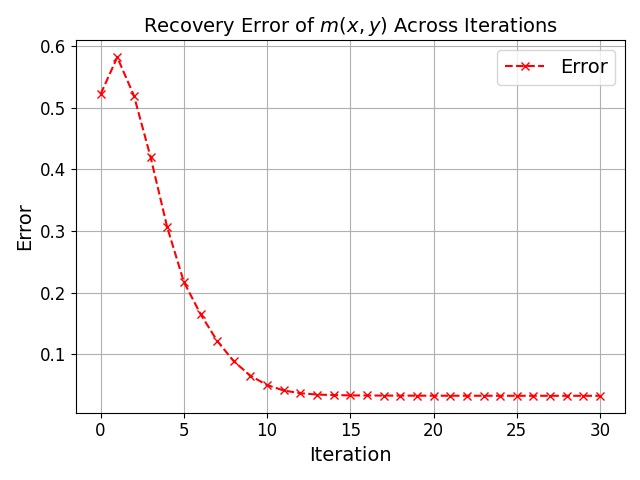}
        \caption{Error \( \mathcal{E}(m^k, m^*) \) vs.  Iteration  \( k \)}
        \label{fig:error2}
    \end{subfigure}

    \begin{subfigure}{0.33\textwidth}
        \centering
        \includegraphics[width=\linewidth]{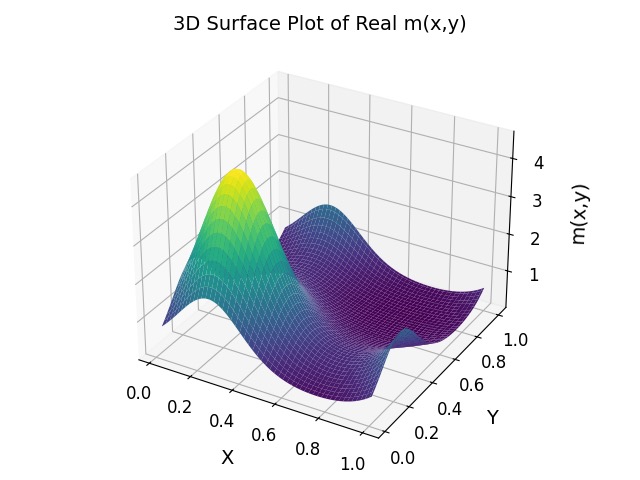}
        \caption{Reference Solution $m^*$}
        \label{fig:real1}
    \end{subfigure}%
    \begin{subfigure}{0.33\textwidth}
        \centering
        \includegraphics[width=\linewidth]{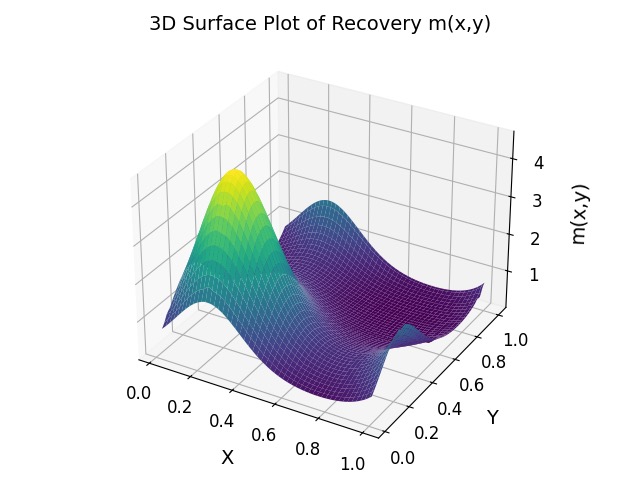}
        \caption{Recovered $m$}
        \label{fig:recover1}
    \end{subfigure}%
    \begin{subfigure}{0.33\textwidth}
        \centering
        \includegraphics[width=\linewidth]{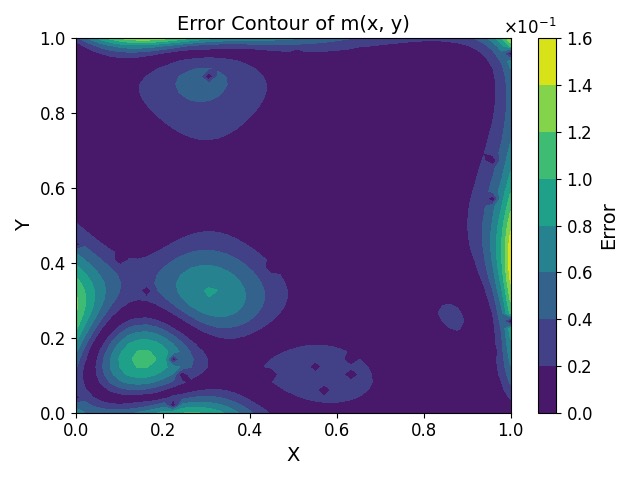}
        \caption{Error Contour of $m$}
        \label{fig:errorcontour1}
    \end{subfigure}

    \begin{subfigure}{0.33\textwidth}
        \centering
        \includegraphics[width=\linewidth]{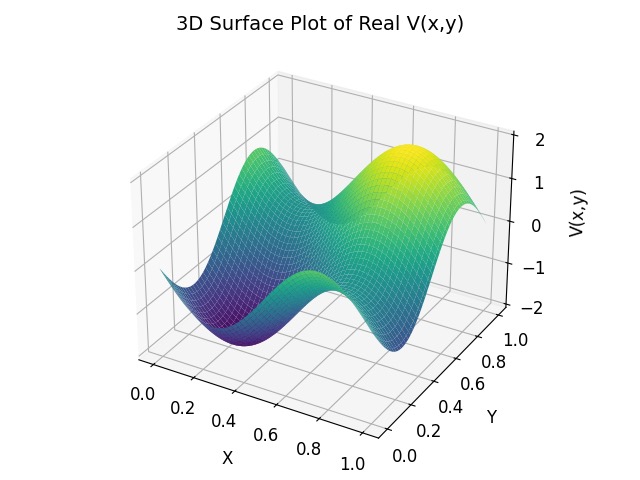}
        \caption{Ground Truth $V$.}
        \label{fig:msample1}
    \end{subfigure}%
    \begin{subfigure}{0.33\textwidth}
        \centering
        \includegraphics[width=\linewidth]{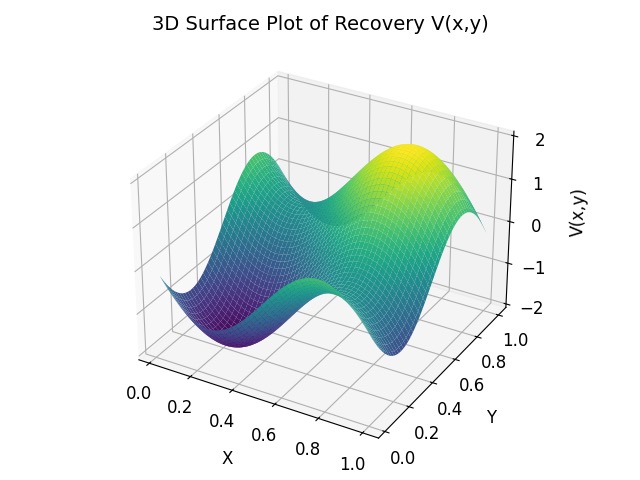}
        \caption{Recovered $V$}
        \label{fig:v01}
    \end{subfigure}%
    \begin{subfigure}{0.33\textwidth}
        \centering
        \includegraphics[width=\linewidth]{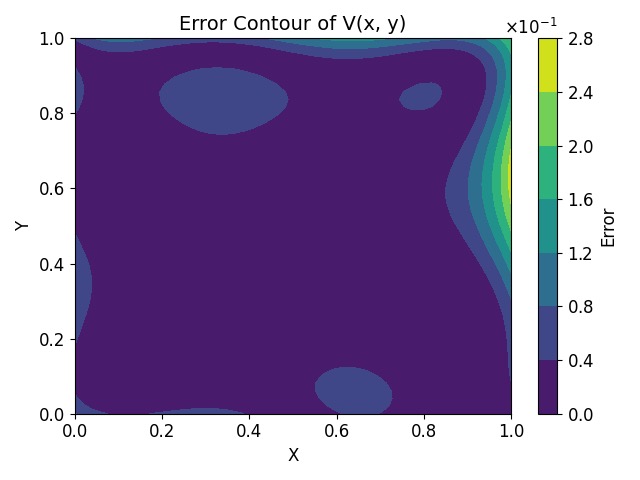}
        \caption{Error Contour of $V$}
        \label{fig:error2}
    \end{subfigure}
    \caption{Numerical results for solving the inverse problem of the MFG system in \eqref{eq:MFG_system} using the inf-sup framework: (a) the sample (grid) points and observation points of \( m \); (b) the observation points of \( V \); (c) the discretized \( L^2 \) error \( \mathcal{E}(m^k, m^*) \) versus the iteration number \( k \); (d) the exact solution \( m^* \); (e) the recovered \( m \); (f) the pointwise error between the recovered \( m \) and the exact \( m^* \); (g) the ground truth $V$; (h) the recovered \( V \); (i) the pointwise error between the recovered values and the exact solution of \( V \).}
    \label{figV}
\end{figure}

\textbf{Experimental Results.} Figure \ref{figV} shows the sample points (collocation grid points) for \( m \) and the observation points of both \( m \) and \( V \). It also displays the discretized \( L^2 \) errors \( \mathcal{E}(m^k, m^*) \) defined in \eqref{eq:l2disc}, which represent the error between the approximated \( m^k \) at the \( k \)-th iteration and the exact solution \( m^* \), as defined in \eqref{eq:rec:pot:explicit}, during the CP iterations. The figure includes the true solutions, the recovered results, and the pointwise error contours of approximated functions of both \( m \) and \( V \). These calculations are consistent across all experiments.

\begin{figure}[h]
    \centering
    \begin{subfigure}{0.33\textwidth}
        \centering
        \includegraphics[width=\linewidth]{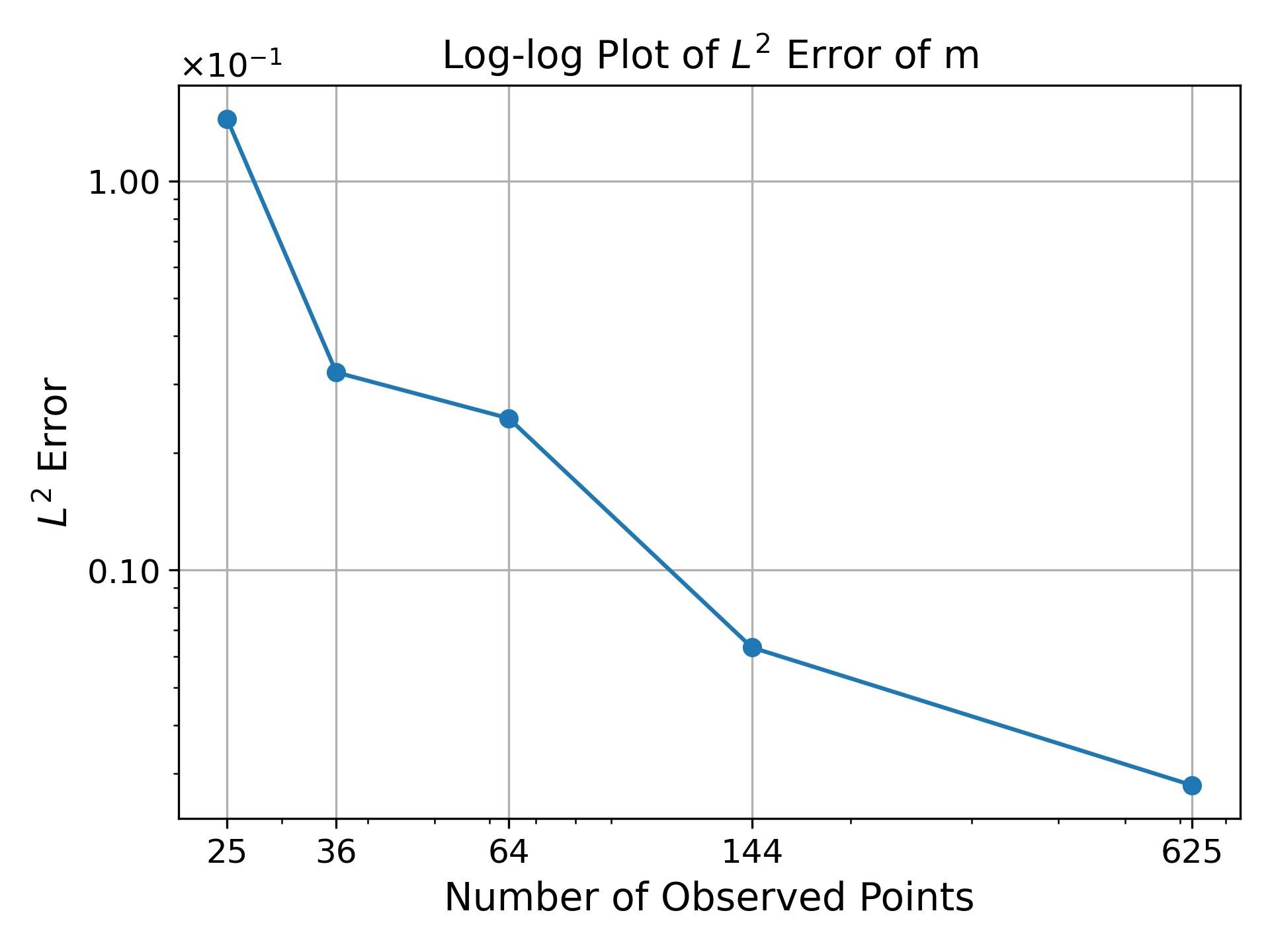}
        \caption{Log-log Plot of $L^2$ Error of $m$}
        \label{log1}
    \end{subfigure}%
    \hspace{0.01\textwidth}% Adjust the space as needed or leave it out for automatic spacing
    \begin{subfigure}{0.33\textwidth}
        \centering
        \includegraphics[width=\linewidth]{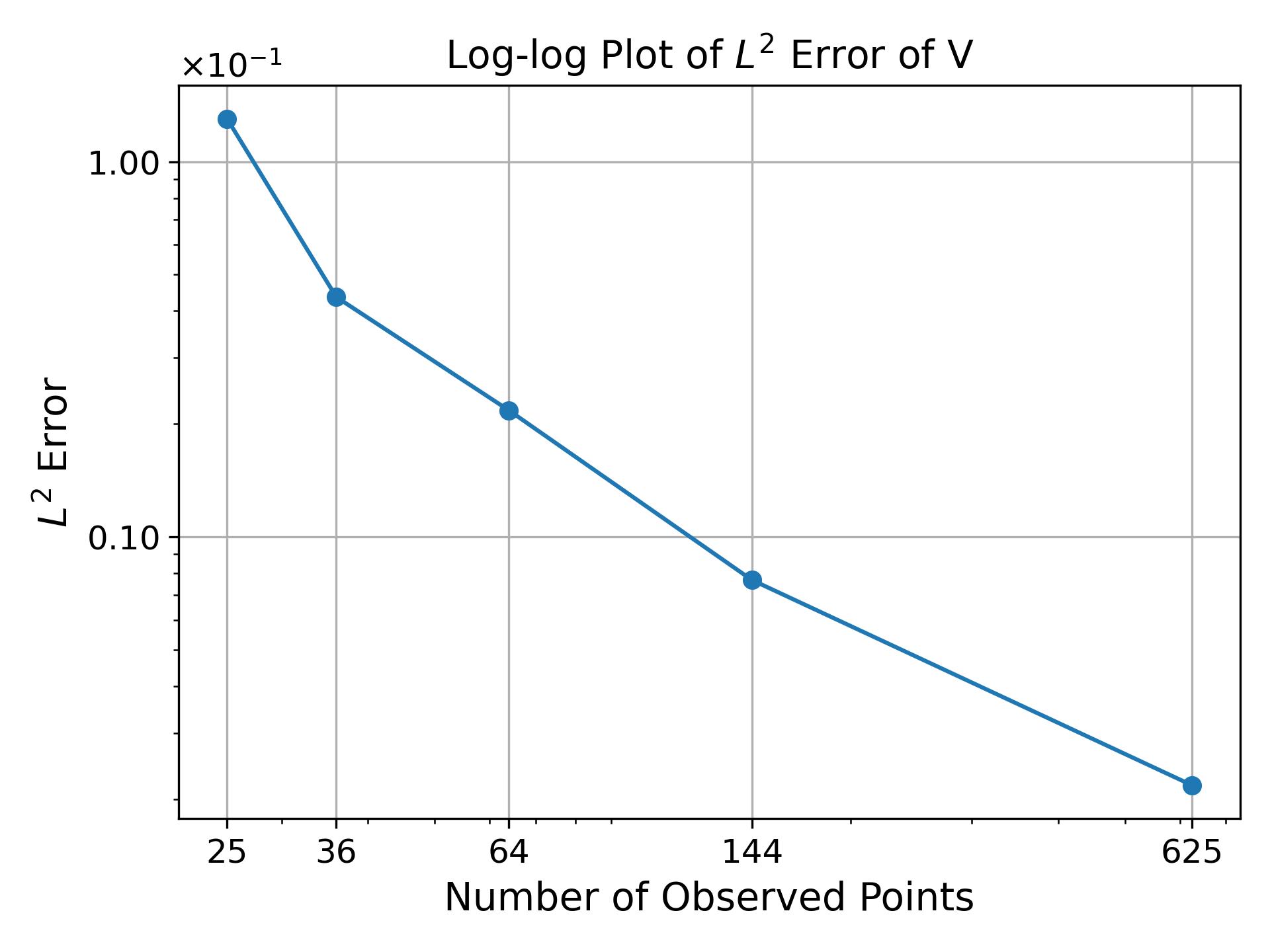}
        \caption{Log-log Plot of $L^2$ Error of $V$}
        \label{log2}
    \end{subfigure}
\caption{Numerical results for solving the inverse problem of the MFG system in \eqref{eq:MFG_system} using the inf-sup framework:  
(a) log-log plot of \( L^2 \) errors for \( m \) versus \( m^* \) as the number of observation points increases.  
(b) log-log plot of \( L^2 \) errors for \( V \) versus its exact values as the number of observation points increases.
}
\end{figure}

The results demonstrate that it is possible to recover the unknown function \( V \) and the distribution \( m \) with a limited number of observations. Figures \ref{log1} and \ref{log2} show that as the number of observation points increases, the average recovery errors for both \( m \) and \( V \) decrease. This illustrates the convergence of the recovery framework with respect to the number of observation points and highlights the importance of observation density in achieving better approximation accuracy.


\subsection{Recovering Spatial Costs and Couplings Using The Bilevel Framework } 
\label{subsec:num:bilevel:VF}
In this subsection, we consider the following MFG system:
\begin{align}
\begin{cases}
\frac{1}{2}|\nabla u(x, y)|^2 - \lambda = f(m(x, y)) + V(x, y), & \forall (x, y) \in \mathbb{T}^2, \\
\operatorname{div}(m(x, y) \nabla u(x, y)) = 0, & \forall (x, y) \in \mathbb{T}^2, \\
\int_{\mathbb{T}^2} m(x, y) \, dx \, dy = 1, \quad \int_{\mathbb{T}^2} u(x, y) \, dx \, dy = 0,
\end{cases}
\label{General}    
\end{align}
where \( V(x, y) = -\frac{1}{2}(\sin(2\pi x) + \sin(2\pi y)) \) and \( f(m) = m^3 \). We focus on recovering the distribution \( m \), the coupling function \( f \), and the spatial cost function \( V \) based on partial noisy observations of \( m \) and \( V \), using the bilevel framework proposed in Section \ref{sec:bilevel_framework_stationary}.

\begin{figure}[h]
    \centering
    \begin{subfigure}[b]{0.33\textwidth}
        \centering
        \includegraphics[width=\linewidth]{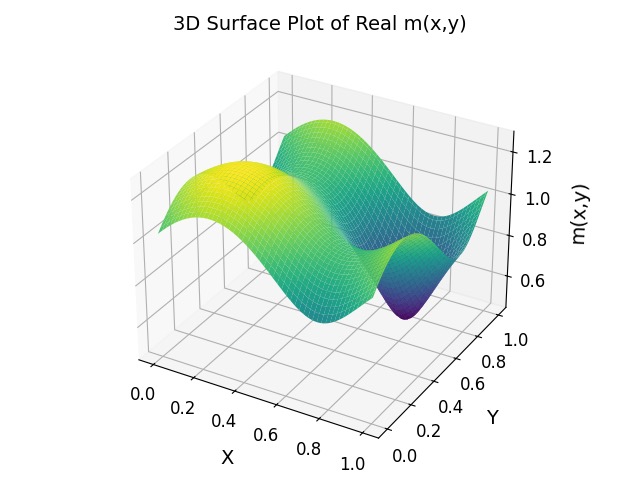}
        \caption{Reference Solution $m^*$}
        \label{fig:real1}
    \end{subfigure}%
    \begin{subfigure}[b]{0.33\textwidth}
        \centering
        \includegraphics[width=\linewidth]{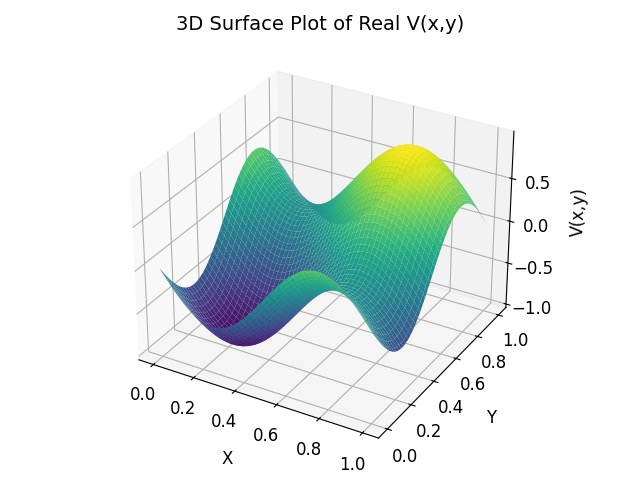}
        \caption{Ground Truth $V$}
        \label{fig:msample1}
    \end{subfigure}%    \end{subfigure}
\caption{Reference results of $m$ and $V$ for the MFG \eqref{General}
}
\label{ReferenceVf}
\end{figure}

For the system in \eqref{General}, the explicit solution is:
\begin{align}
\label{eq:general:explcit}
u^*(x, y) = 0, \quad m^*(x, y) = \left( -V(x, y) - \lambda \right)^{\frac{1}{3}},
\end{align}
where \( \lambda \) satisfies:
\begin{align*}
\int_{\mathbb{T}^2} \left( \frac{1}{2}(\sin(2\pi x) + \sin(2\pi y)) - \lambda \right)^{\frac{1}{3}} \, dx \, dy = 1.
\end{align*}
We treat the explicit solutions for \(m\) and \(V\) as reference solutions, which are illustrated in Figure~\ref{ReferenceVf}.

\subsubsection{Estimation of \(F\) as a Power Function }
\label{Power}
 Here , we assume that the prior assumption of $f(m)$ is the power function $f(m)=m^\alpha$. Hence, the primitive function  $F$ of $f$ is $F(m)=m^\alpha/\alpha$. 
 The goal is to recover \( m \), \( V \), and \( \alpha \) from noisy partial observations of \( m \) and \( V \) using the framework described in Subsection \ref{Powerfunctionmethod}. 

\textbf{Experimental Setup.} The grid resolution is set to \( h = \frac{1}{40} \), with the observation points \( \boldsymbol{m}^o \) consisting of 100 samples selected from the total of 1600 grid points of \( m \). We randomly choose 100 observation points for \( V \) in the spatial domain. The regularization parameters are set as \( \alpha_{m^o} = 10^5 \) and \( \alpha_{v^o} = \infty \). The Gaussian regularization coefficient \( \alpha_v \) is set to 1, and \( \alpha_{\beta} = 10 \). To simulate measurement noise, Gaussian noise \( \mathcal{N}(0, \gamma^2 I) \) with a standard deviation \( \gamma = 10^{-3} \) is added to the observations.

\begin{figure}[h]
    \centering
    \begin{subfigure}[b]{0.33\textwidth}
        \centering
        \includegraphics[width=\linewidth]{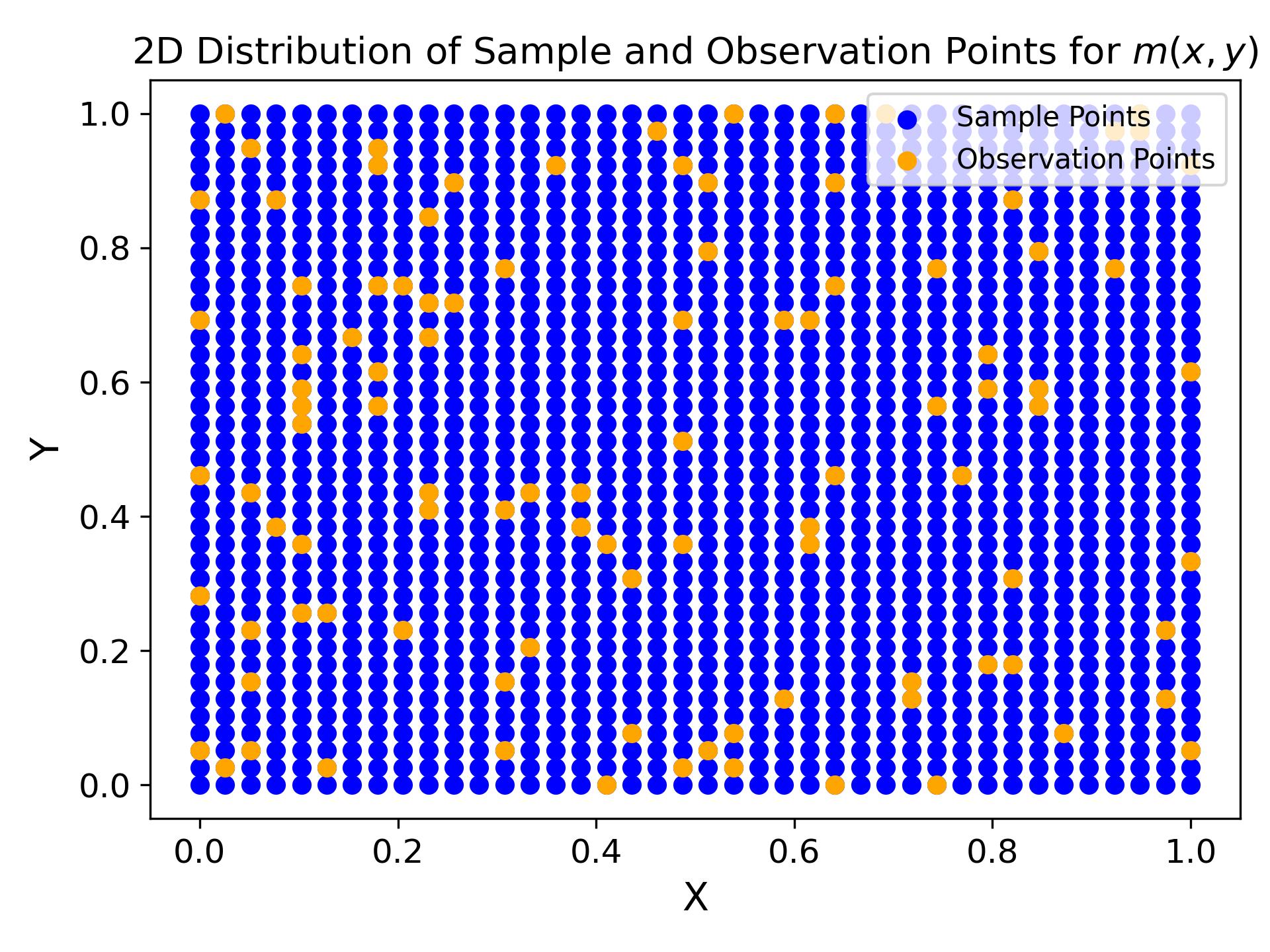}
        \caption{Samples \& Observations of $m$}
        \label{fig:msample1}
    \end{subfigure}%
    \begin{subfigure}[b]{0.33\textwidth}
        \centering
        \includegraphics[width=\linewidth]{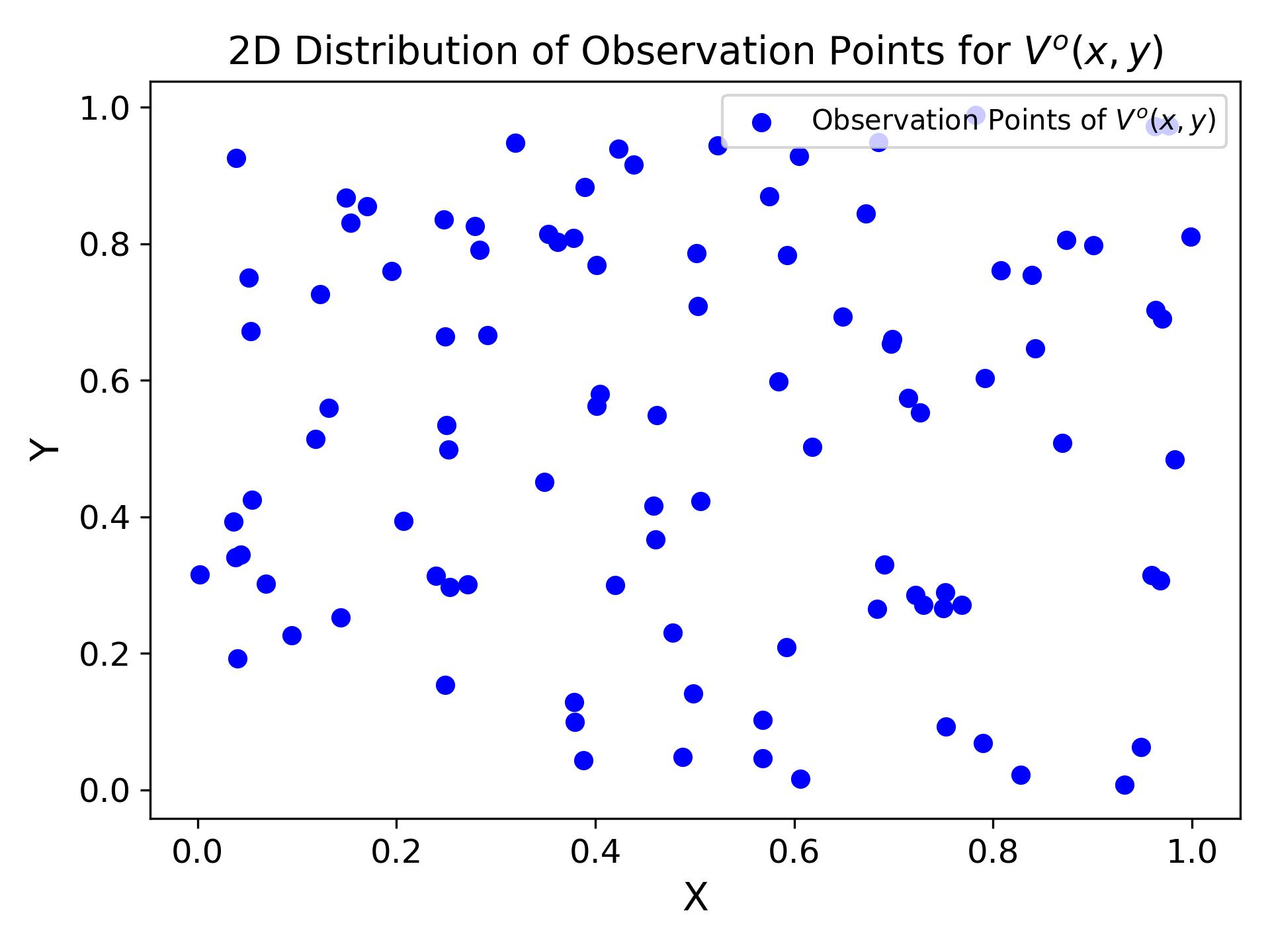}
        \caption{Observations of $V$}
        \label{fig:v01}
    \end{subfigure}%
    \begin{subfigure}[b]{0.33\textwidth}
        \centering
        \includegraphics[width=\linewidth]{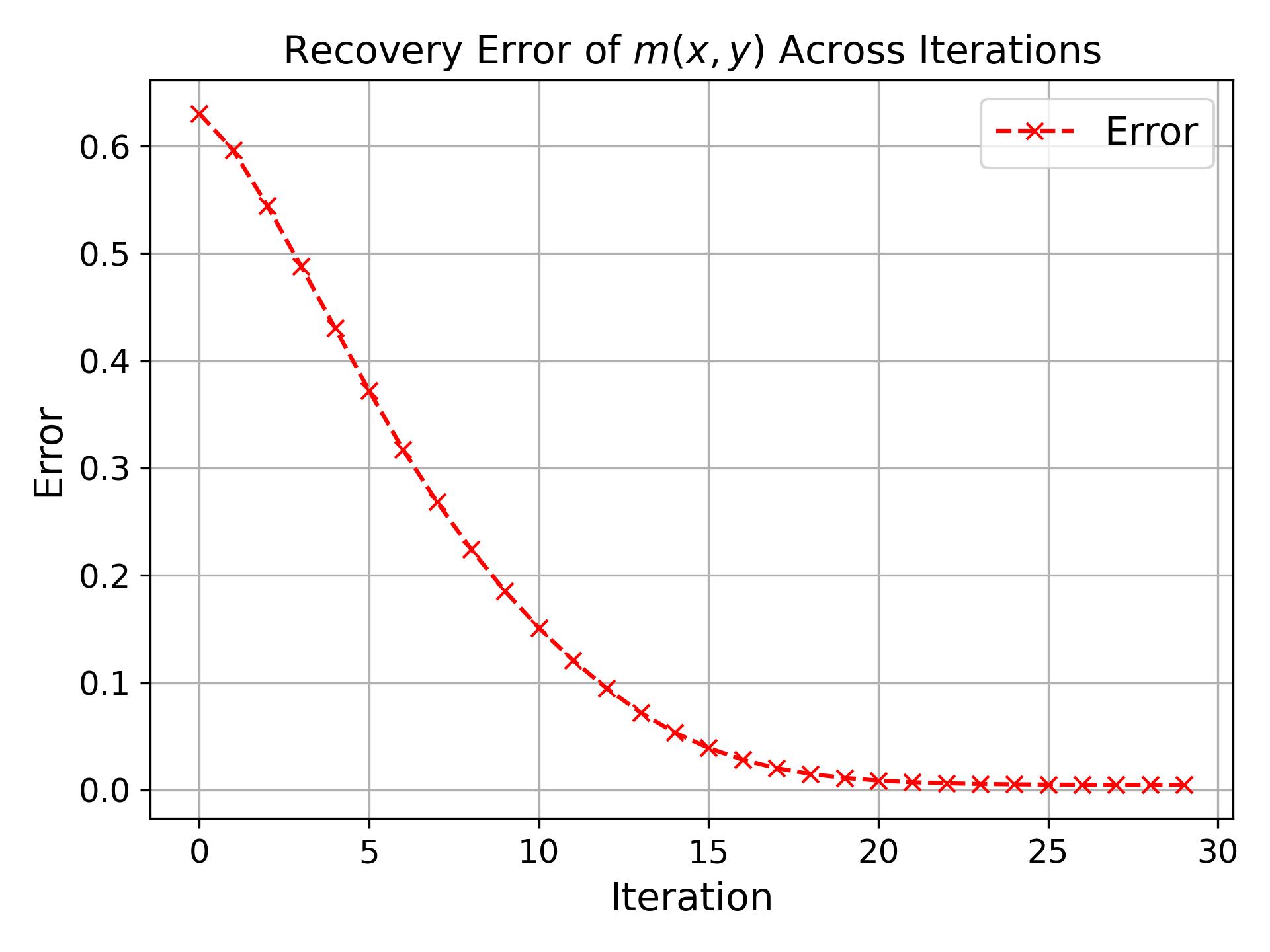}
        \caption{Error \( \mathcal{E}(m^k, m^*) \) vs.  Iteration  \( k \)}
        \label{fig:error2}
    \end{subfigure}
  \begin{subfigure}[b]{0.33\textwidth}
        \includegraphics[width=\linewidth]{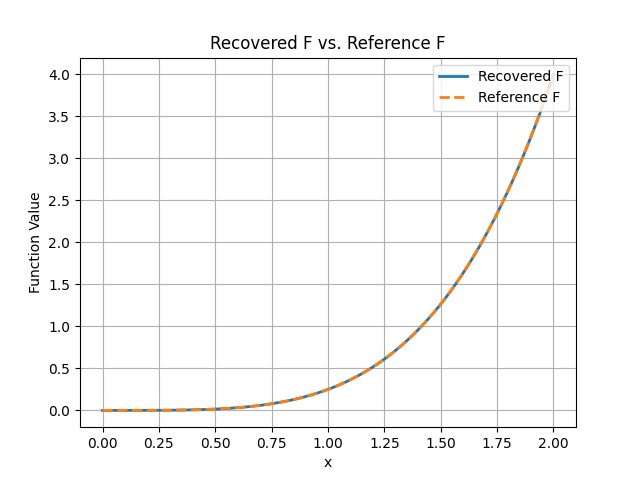}
        \caption{Recovered F vs. Reference F}
    \label{VfpolyF}
    \end{subfigure}
    \begin{subfigure}[b]{0.33\textwidth}
        \centering
        \includegraphics[width=\linewidth]{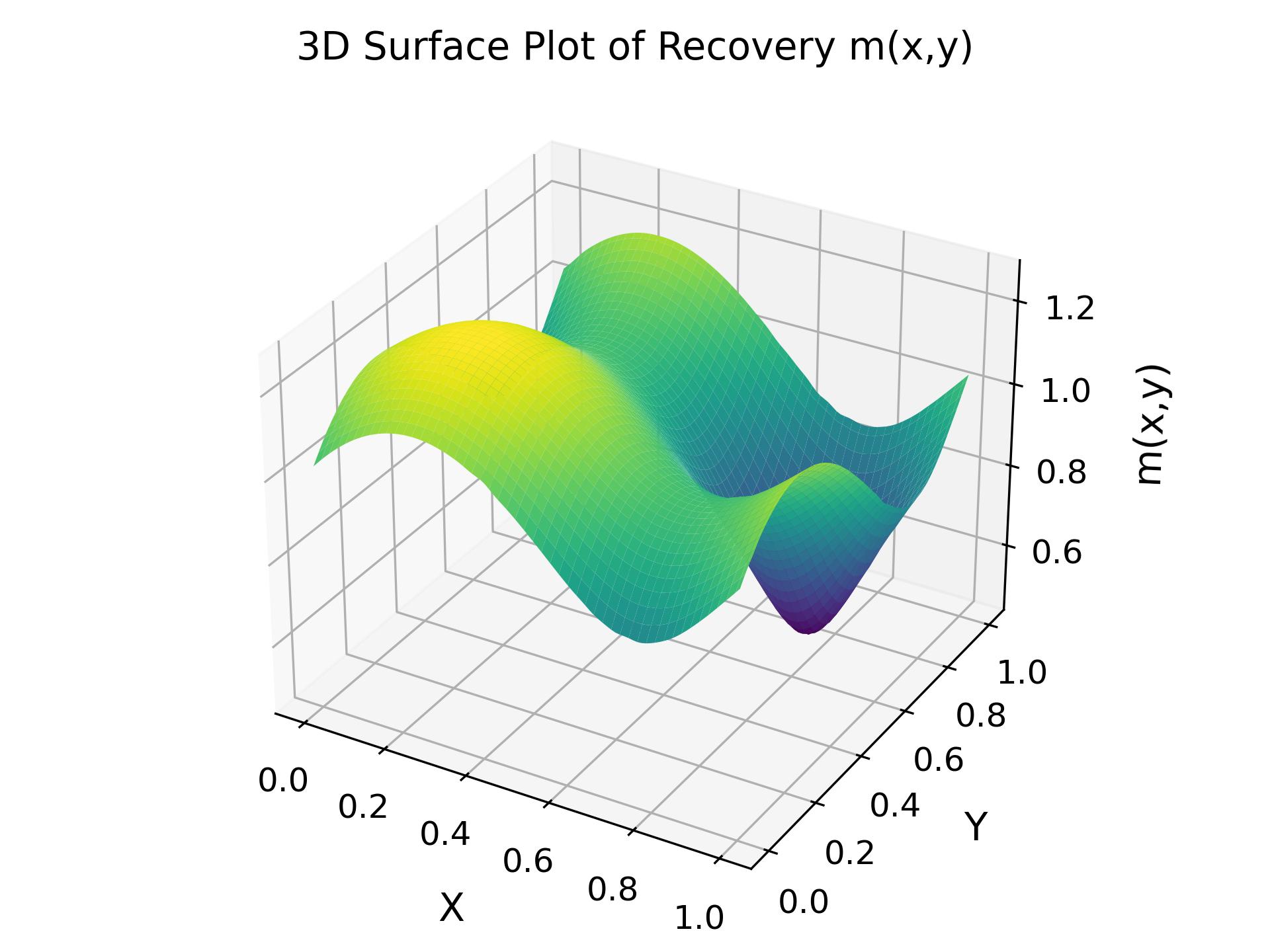}
        \caption{Recovered $m$}
        \label{fig:recover1}
    \end{subfigure}%
    \begin{subfigure}[b]{0.33\textwidth}
        \centering
        \includegraphics[width=\linewidth]{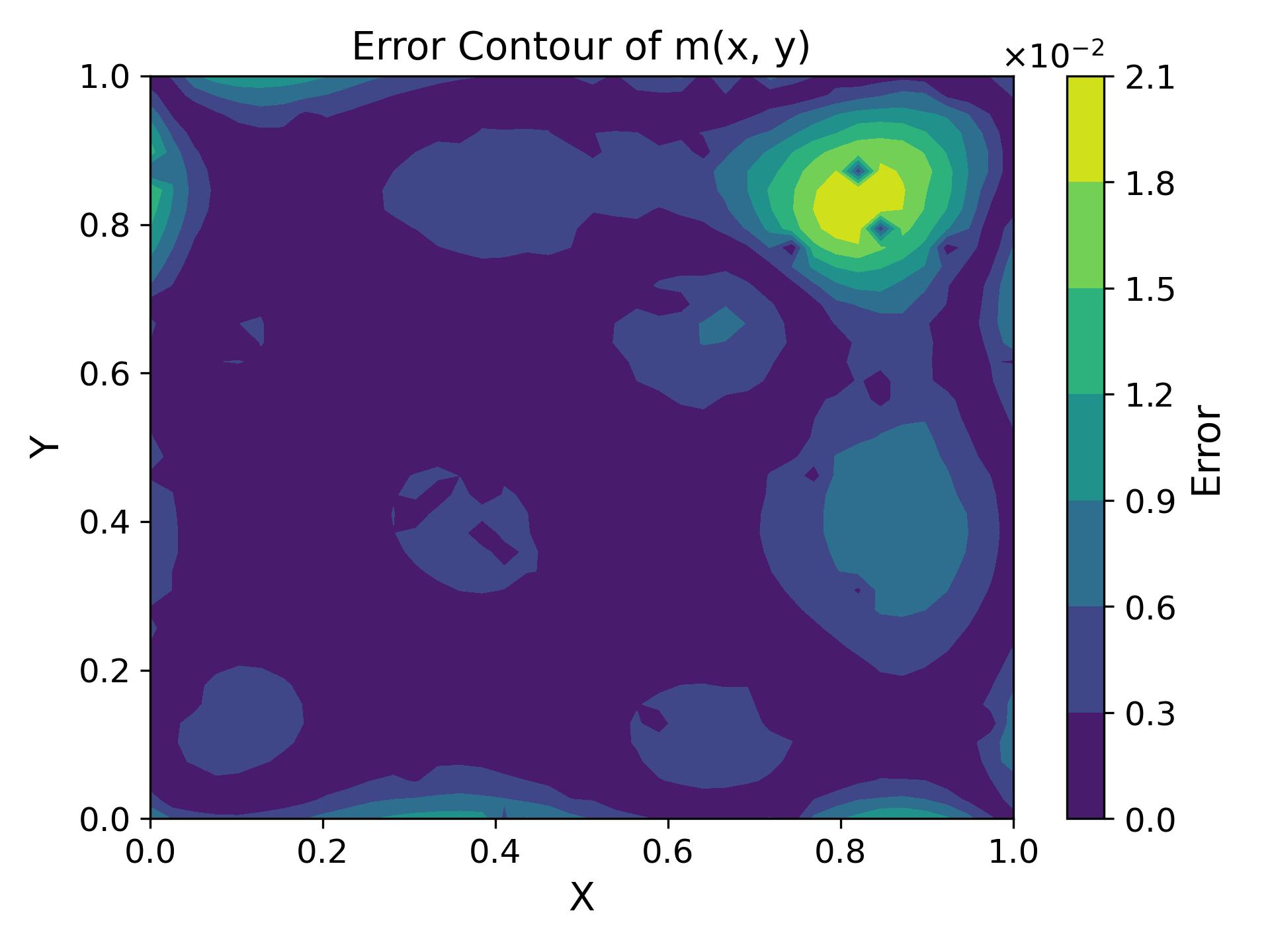}
        \caption{Error Contour of $m$}
        \label{fig:errorcontour1}
    \end{subfigure}
    \begin{subfigure}[b]{0.33\textwidth}
        \centering
        \includegraphics[width=\linewidth]{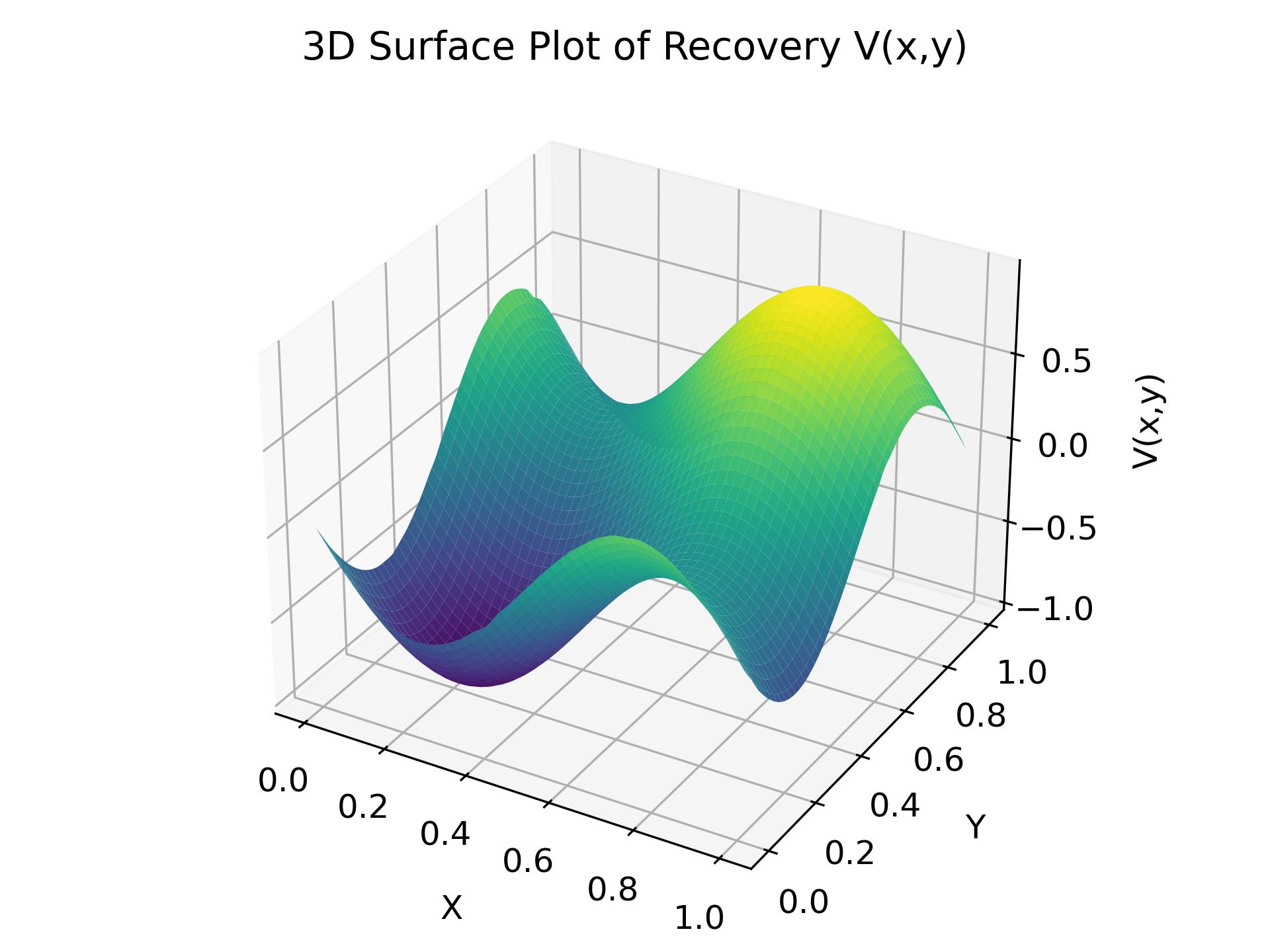}
        \caption{Recovered $V$}
        \label{fig:v01}
    \end{subfigure}%
    \begin{subfigure}[b]{0.33\textwidth}
        \centering
        \includegraphics[width=\linewidth]{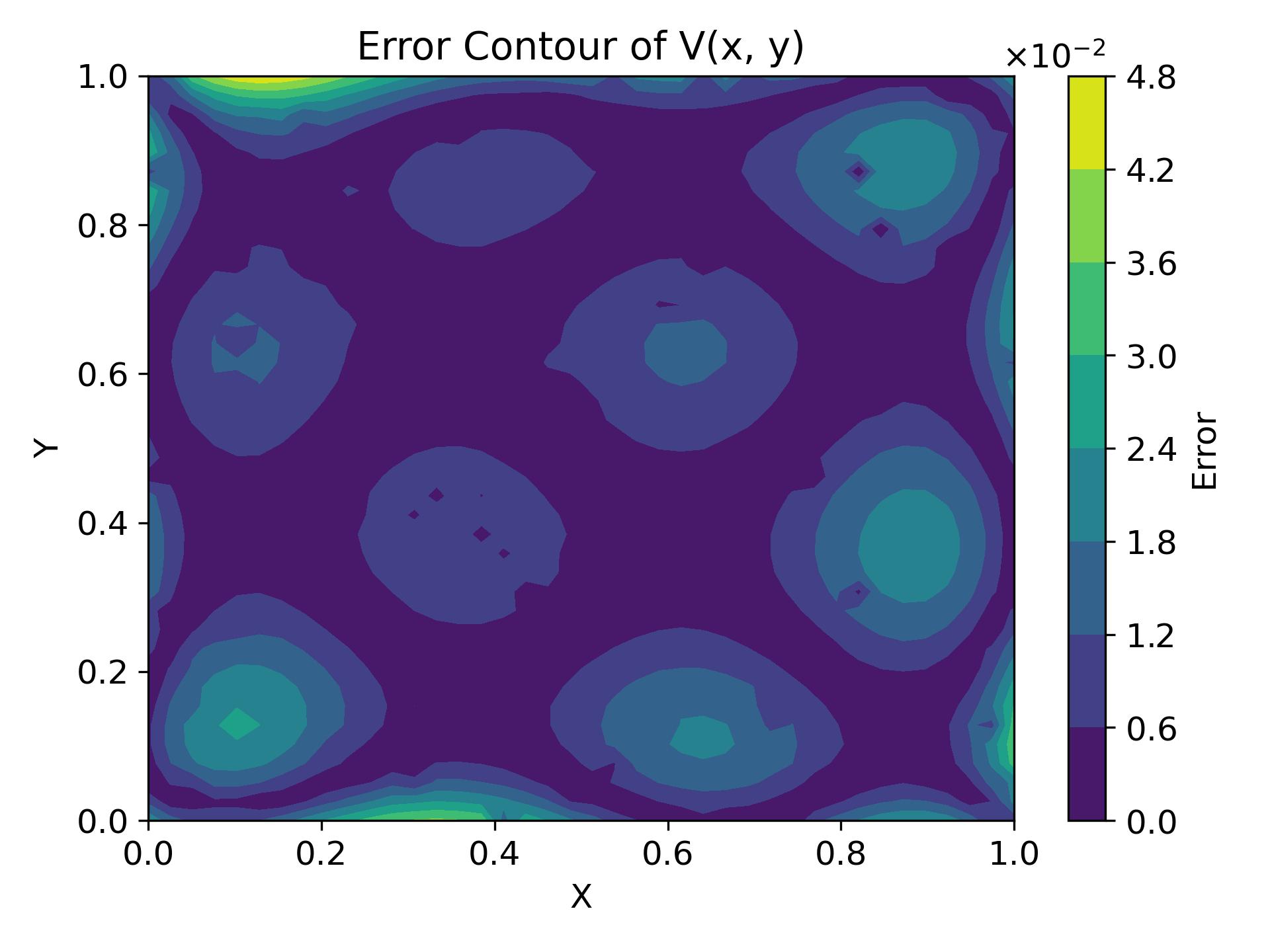}
        \caption{Error Contour of $V$}
        \label{fig:error2}
    \end{subfigure}
    \caption{Numerical results for solving the inverse problem  of the MFG system in \eqref{General} using a power function to approximate the coupling function \( F \):  
(a) sample (grid) points and observation points of \( m \);  
(b) observation points of \( V \);  
(c) discretized \( L^2 \) error \( \mathcal{E}(m^k, m^*) \) versus iteration number \( k \); 
(d) recovered \( F \) vs. reference \( F \); 
(e) recovered \( m \);  
(f) pointwise error between the recovered \( m \) and the exact \( m^* \);   
(g) recovered \( V \);  
(h) pointwise error between the recovered \( V \) and its exact solution.
}
    \label{figVfbeta}
\end{figure}

\begin{figure}[h]
    \centering
    \begin{subfigure}{0.33\textwidth}
        \centering
        \includegraphics[width=\linewidth]{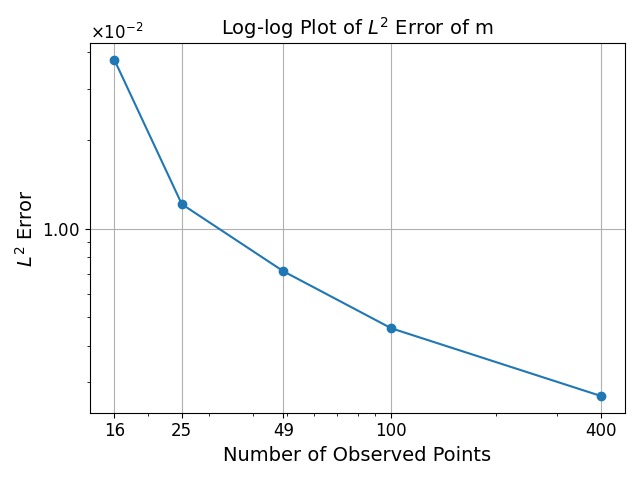}
        \caption{Log-log Plot of $L^2$ Error of $m$}
        \label{log5}
    \end{subfigure}%
    \hspace{0.01\textwidth}% Adjust the space as needed or leave it out for automatic spacing
    \begin{subfigure}{0.33\textwidth}
        \centering
        \includegraphics[width=\linewidth]{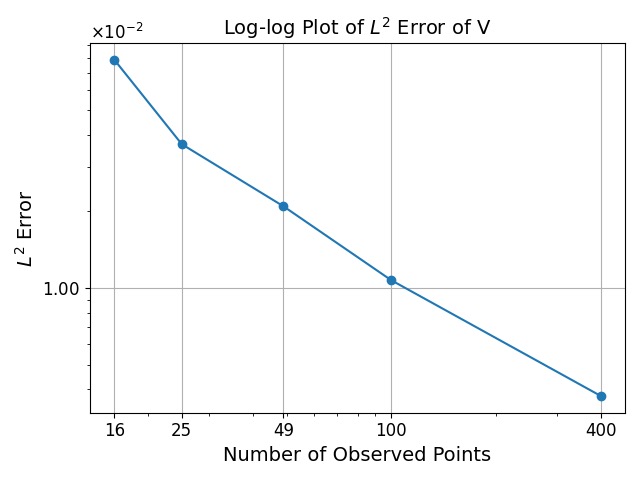}
        \caption{Log-log Plot of $L^2$ Error of $V$}
        \label{log6}
    \end{subfigure}
\caption{Numerical results for solving the inverse problem of the MFG system in \eqref{General} using a power function approximation for the coupling function \( F \):  
(a) log-log plot of \( L^2 \) errors for \( m \) versus \( m^* \) as the number of observation points increases.  
(b) log-log plot of \( L^2 \) errors for \( V \) versus its exact values as the number of observation points increases.
}
\end{figure}

\textbf{Experiment Results.} 
Figure \ref{figVfbeta} presents the sample points (collocation grid points) for \( m \) and the observation points for both \( m \) and \( V \). It also shows the discretized \( L^2 \) errors \( \mathcal{E}(m^k, m^*) \) defined in \eqref{eq:l2disc}, representing the error between the approximated solution \( m^k \) at the \( k \)-th iteration and the exact solution \( m^* \), as defined in \eqref{eq:general:explcit}, during the CP iterations. The figure includes the true solutions, the recovered results, and the pointwise error contours of the approximated functions of both \( m \) and \( V \). To ensure that \( \alpha \geq 1 \) for the monotonicity of \( m^{\alpha} \), we parameterize \( \alpha \) as \( \alpha = \ln(e^{\beta} + 1) + 1 \) and optimize \( \beta \) instead. The recovered value of \( \beta \) is 1.8524, which closely matches the true value of 1.8523 when \( \alpha = 3 \). Figures \ref{log5} and \ref{log6} illustrate the recovery error for \( m \) and \( V \) under varying numbers of observation points, highlighting the importance of observation density in achieving better approximation accuracy.

\subsubsection{Convex Function Library Method}  
\label{poltnomial method}
In this approach, we assume that the prior assumption of $F(m)$ is only a general convex function approximated as a polynomial sequence \( F(m) = \sum_k \gamma_k m^k \), assuming the coefficients \( \gamma_k \) are Gaussian variables. As before, we model the function \( V \) using a GP and follow the framework proposed in Subsection \ref{Function group}. The objective is to infer the optimal values \( m^*, w^* \), \( V \), and \( F \) from the observations \( m^o \) and \( V^o \).

\begin{comment}
During the training phase, we initially reduce the learning rate of \( z \) to prioritize learning the parameter \( V \) since the Gaussian loss dominates at the beginning. Once the total loss decreases to a certain level and the estimate of \( V \) becomes more accurate, we increase the learning rate of \( z \) and continue optimizing both \( z \) and \( V \) simultaneously.  
\end{comment}

\textbf{Experimental Setup.} The grid size is set to \( h = \frac{1}{50} \), and 625 observation points \( \boldsymbol{m}^o \) are sampled from the 2500 grid points of \( m \). The observation points for \( m \) are selected from the grid, while the 625 observation points for \( V \) are randomly distributed in space without grid constraints. We set \( \alpha_{m^o} = 10^6 \) and \( \alpha_{v^o} = \infty \). The Gaussian regularization coefficient is \( \alpha_v = 10^4 \), with \( \alpha_{\gamma_k} = 2 \times 10^4 \) and \( \alpha_{fp} = 0.01 \). Gaussian noise \( \mathcal{N}(0, \gamma^2 I) \) with a standard deviation \( \gamma = 10^{-3} \) is added to the observations. For the polynomial method, we approximate the coupling function \( F \) as a sum of eight polynomials, \( \sum_{k=1}^8 \gamma_k x^k \), where the coefficients \( (\gamma_k)_{k=1}^8 \) are modeled as Gaussian variables. The initial values for both \( m \) and \( V \) are set to 1. In our experiments, the fixed points \( \{\widetilde{r}_i\}_{i=1}^{N_{\text{MC}}} \) are uniformly spaced between 0.5 and 1.2, with \( N_{\text{MC}} = 500 \) points used in the Monte Carlo approximation for the convexity penalization.

\begin{figure}[h]
    \centering
    \begin{subfigure}[b]{0.33\textwidth}
        \centering
        \includegraphics[width=\linewidth]{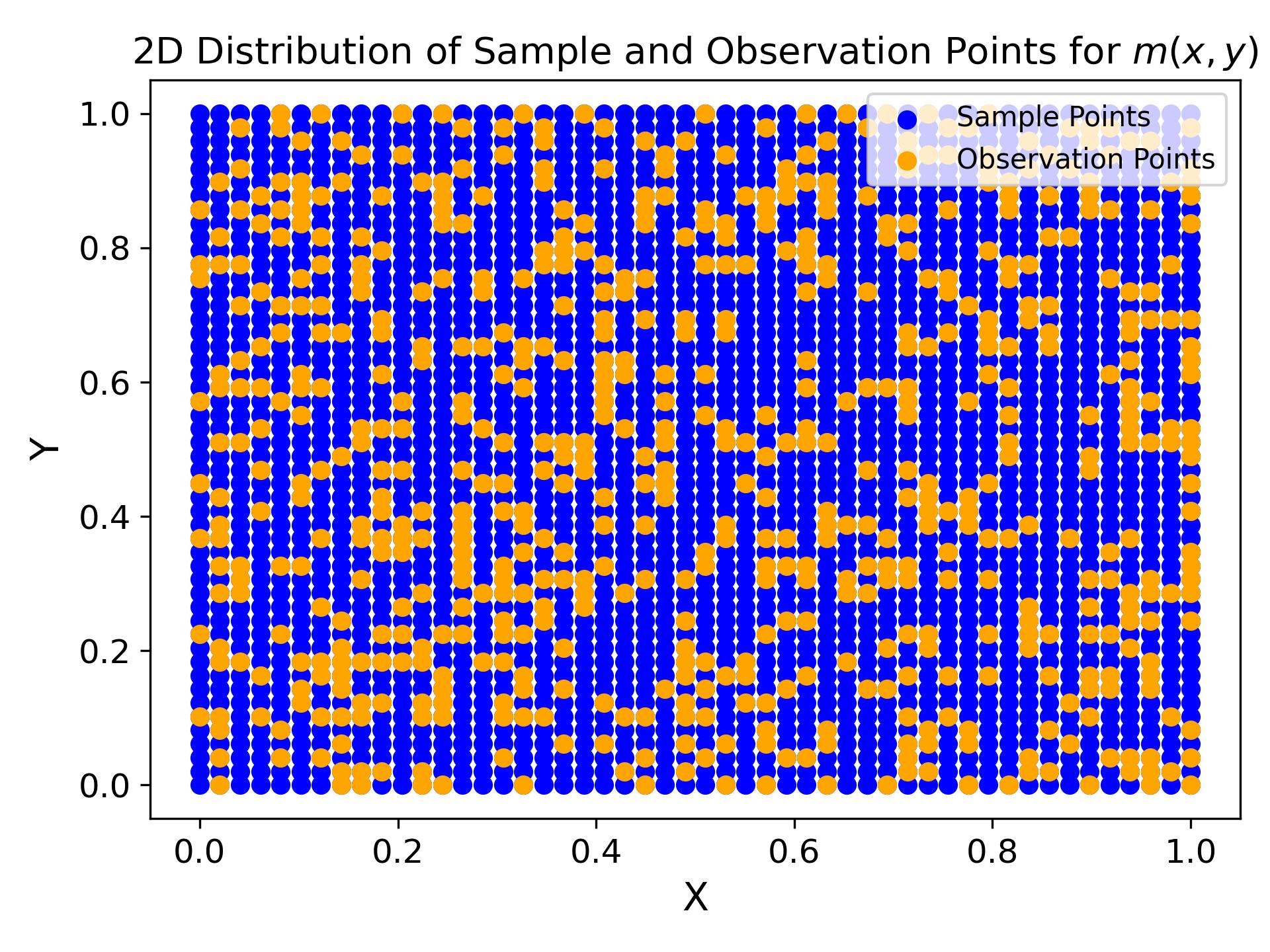}
        \caption{Samples \& Observations of $m$}
        \label{fig:msample1}
    \end{subfigure}%
    \begin{subfigure}[b]{0.33\textwidth}
        \centering
        \includegraphics[width=\linewidth]{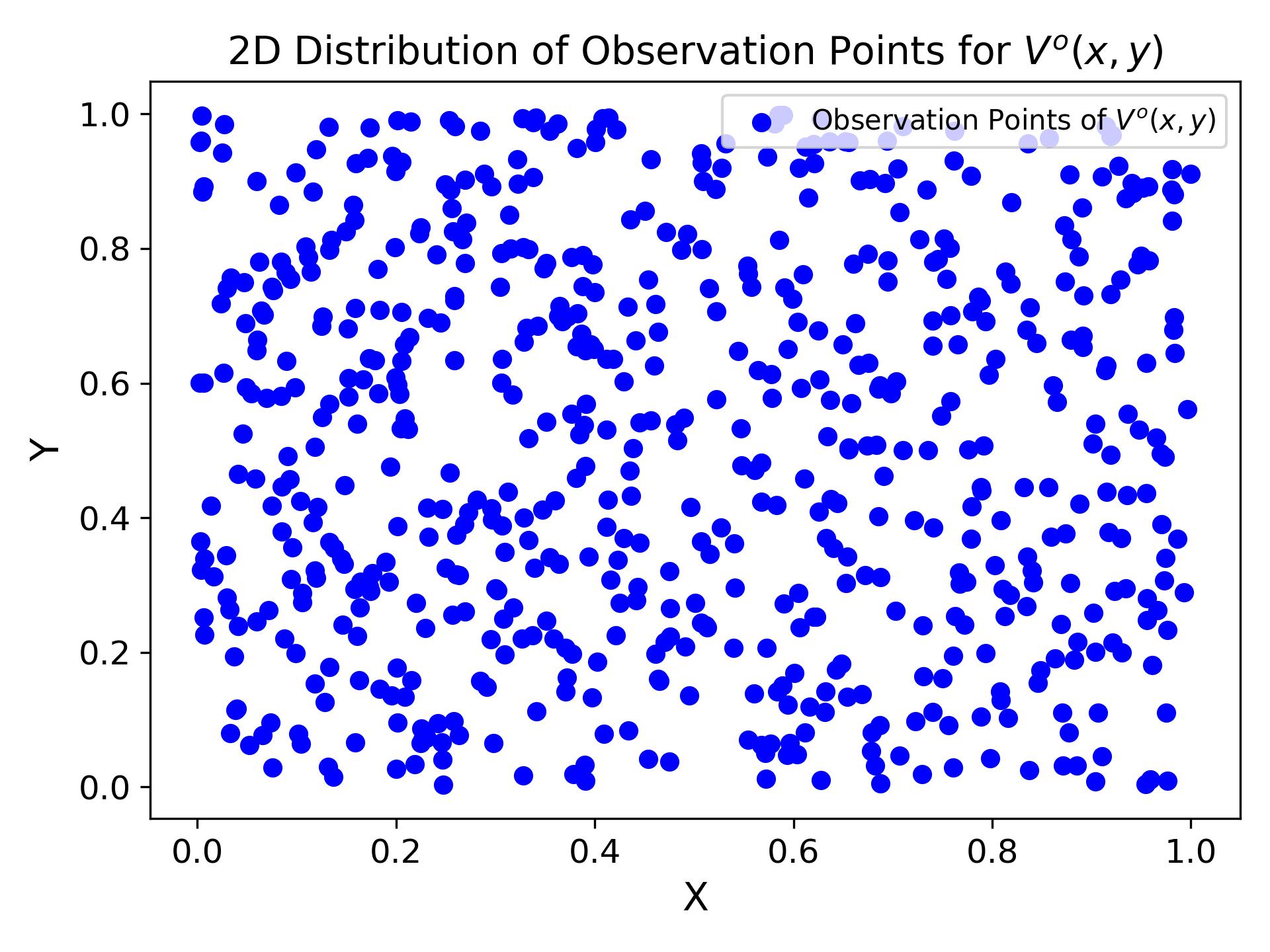}
        \caption{Observations of $V$}
        \label{fig:v01}
    \end{subfigure}%
    \begin{subfigure}[b]{0.33\textwidth}
        \centering
        \includegraphics[width=\linewidth]{FigureVfbeta/afig1Vfbeta.jpg}
        \caption{Error \( \mathcal{E}(m^k, m^*) \) vs.  Iteration  \( k \)}
        \label{fig:error2}
    \end{subfigure}
    \begin{subfigure}[b]{0.33\textwidth}
        \includegraphics[width=\linewidth]{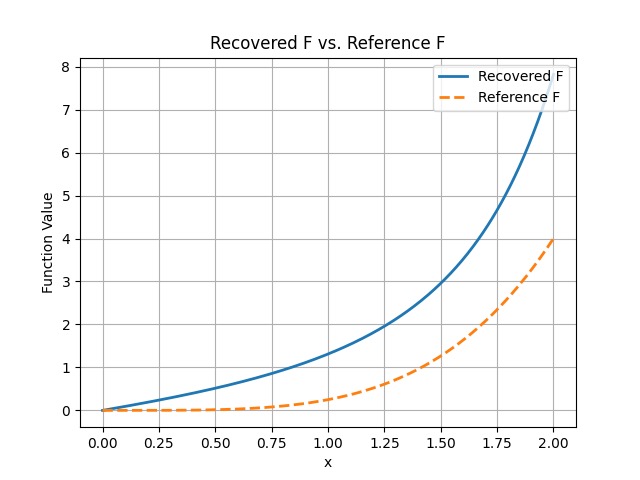}
        \caption{Recovered F vs. Reference F}
    \label{VfpolyF}
    \end{subfigure}
\begin{subfigure}[b]{0.33\textwidth}
        \centering
        \includegraphics[width=\linewidth]{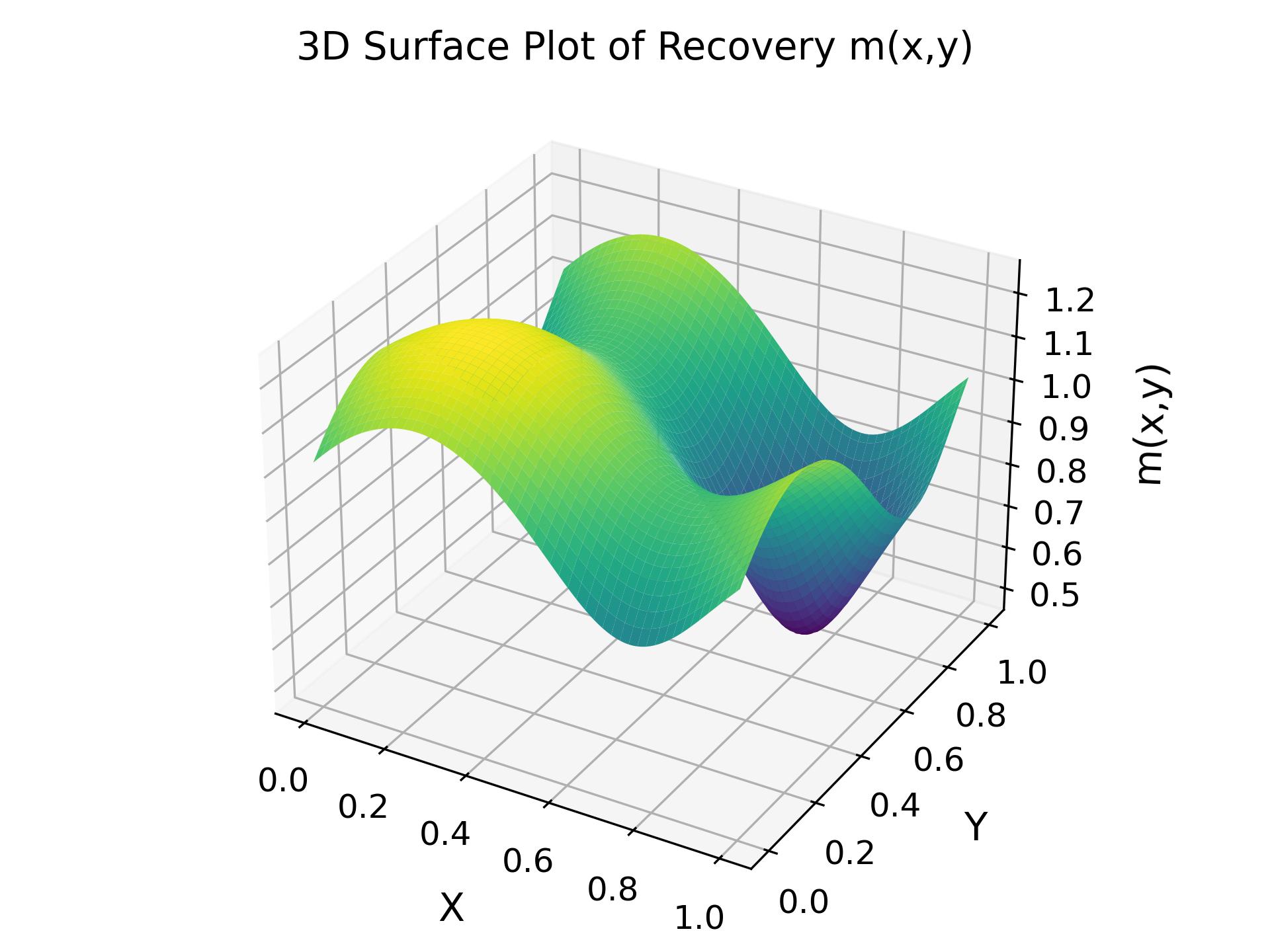}
        \caption{Recovered $m$}
        \label{fig:recover1}
    \end{subfigure}%
    \begin{subfigure}[b]{0.33\textwidth}
        \centering
        \includegraphics[width=\linewidth]{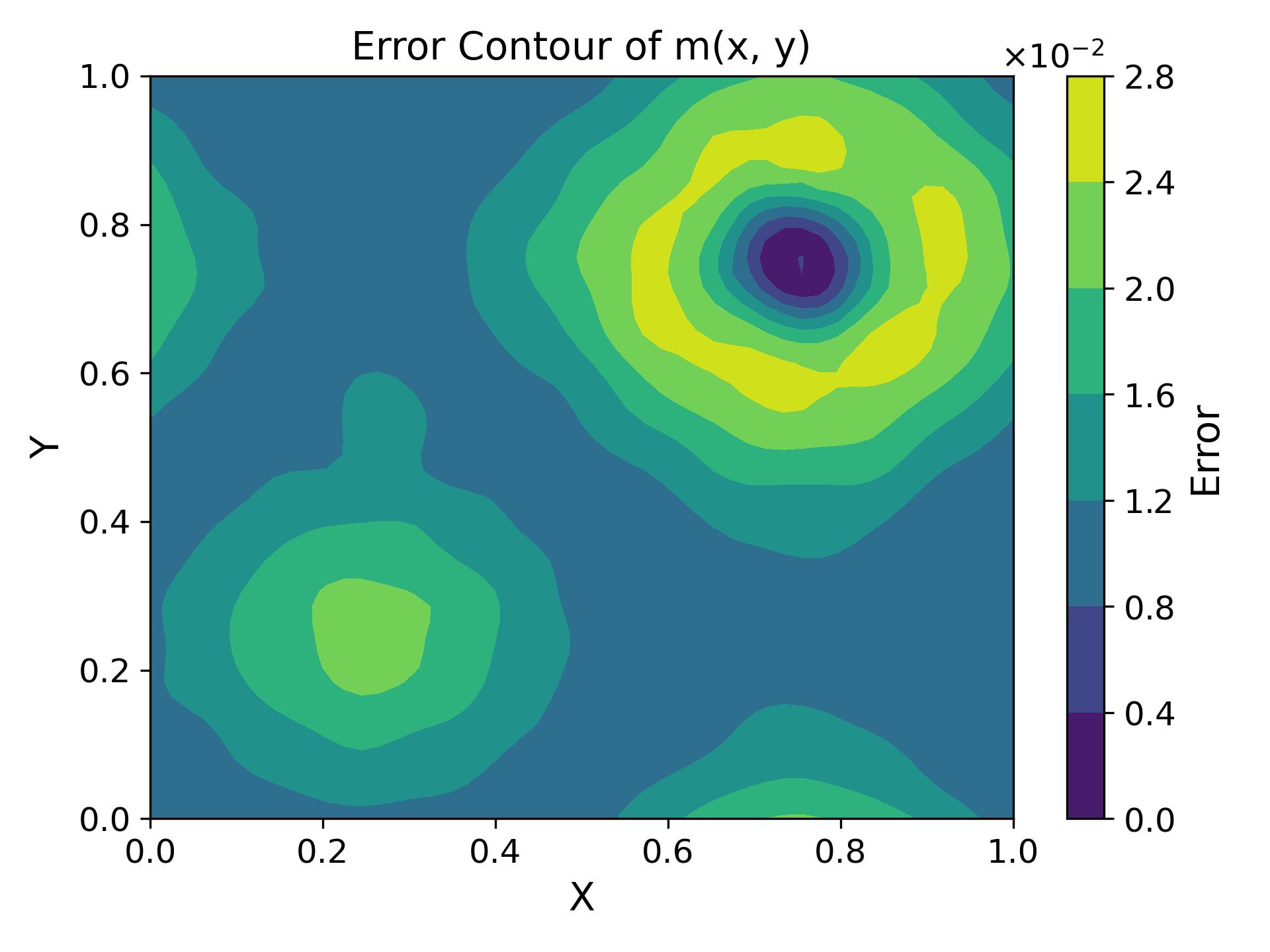}
        \caption{Error Contour of $m$}
        \label{fig:errorcontour1}
    \end{subfigure}
     \begin{subfigure}[b]{0.33\textwidth}
        \centering
        \includegraphics[width=\linewidth]{FigureVfbeta/afig4Vfbeta.jpg}
        \caption{Recovered $V$}
        \label{fig:v01}
    \end{subfigure}%
    \begin{subfigure}[b]{0.33\textwidth}
        \centering
        \includegraphics[width=\linewidth]{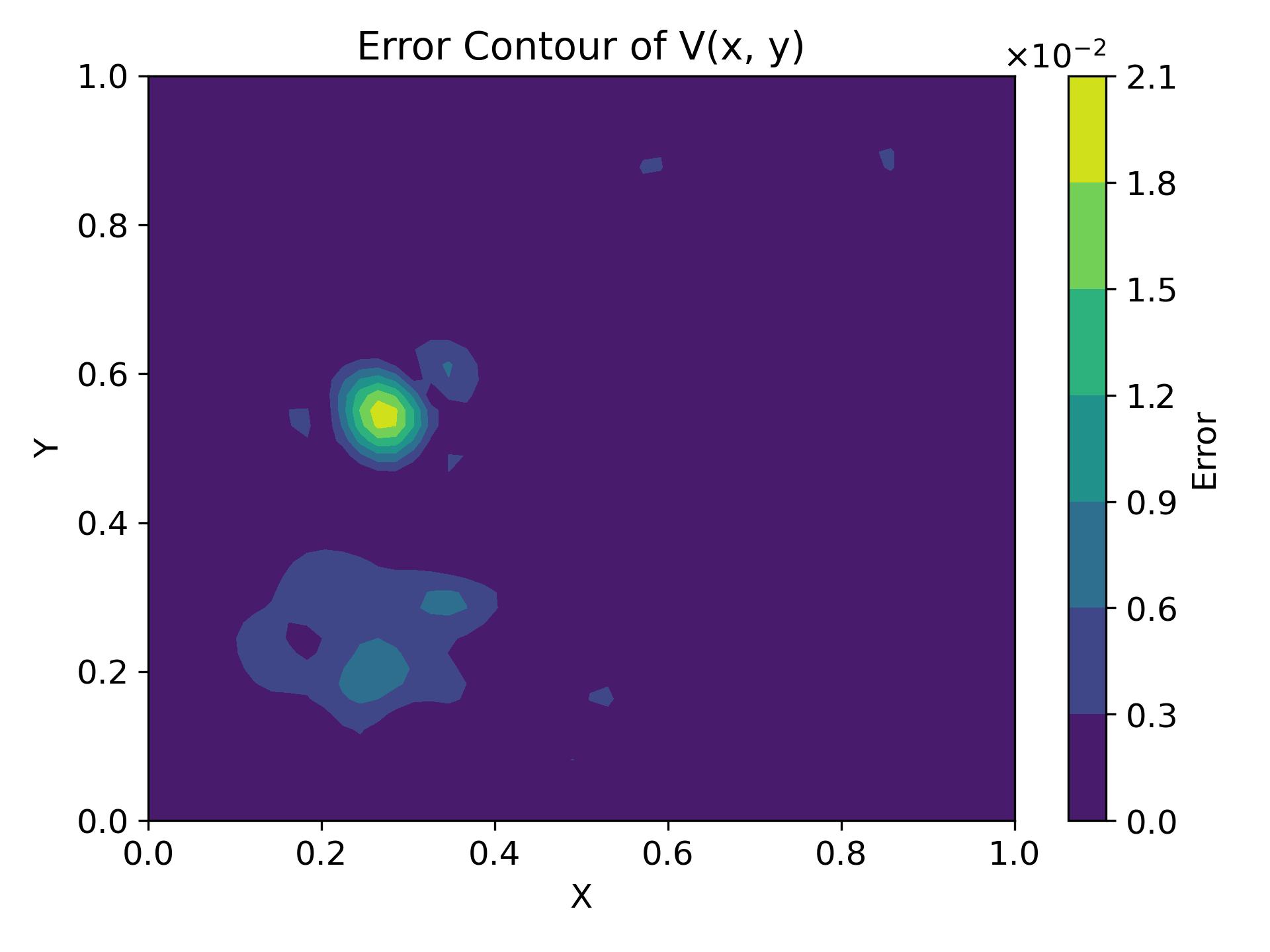}
        \caption{Error Contour of $V$}
        \label{fig:error2}
    \end{subfigure}
    
    \caption{Numerical results for recovering the coupling \(F\) using polynomial approximations for the MFG system in \eqref{General} are shown as follows: (a) sample (grid) points and observation points of \( m \);  
(b) observation points of \( V \);  
(c) discretized \( L^2 \) error \( \mathcal{E}(m^k, m^*) \) versus iteration number \( k \); 
(d) recovered \( F \) vs. reference \( F \); 
(e) recovered \( m \);  
(f) pointwise error between the recovered \( m \) and the exact \( m^* \);   
(g) recovered \( V \);  
(h) pointwise error between the recovered \( V \) and its exact solution. Although the recovered \(F\) differs from the ground truth, it remains convex, and the corresponding MFG generates solutions that closely fit the observed data.}
    \label{figVfpoly}
\end{figure}

\textbf{Experiment Results.} 
Figure \ref{figVfpoly} shows the collocation grid points for \( m \) and the observation points for both \( m \) and \( V \). It also depicts the discretized \( L^2 \) errors \(\mathcal{E}(m^k, m^*)\) from \eqref{eq:l2disc}, which quantify the discrepancy between the approximate solution \( m^k \) at iteration \( k \) and the exact solution \( m^* \) defined in \eqref{eq:general:explcit}, during the CP iterations. The figure includes pointwise error contours for the approximated \( m \) and \( V \). 

However, as shown in Figure~\ref{VfpolyF}, the recovered function \(F\) does not match the ground truth, likely because the polynomial ansatz for approximating \(F\) discards more prior information than the power-function approach, which only estimates an exponential term. Recovering \(m\), \(V\), and \(F\) together is inherently ill-posed, as multiple combinations of \(F\) and \(V\) can produce the same MFG profiles. Consequently, additional observations of \(m\) and \(V\) are needed to identify the underlying MFG, since more data narrows the search space and enhances identifiability. Still, the inner minimization in the bilevel formulation always admits some MFG solution, allowing alternative parameter sets that fit the data without necessarily reproducing the true parameters. This underscores the ill-posedness of the inverse problem and motivates further investigation into uniqueness.

Nevertheless, finding any data-consistent model remains valuable in practice: multiple valid MFGs may exist, each offering insights into potential underlying dynamics and supporting tasks such as forecasting or scenario analysis. Incorporating tailored priors, adaptive loss weighting, and advanced optimization strategies could help address these challenges in both data-rich and data-scarce settings. We leave the exploration of these directions to future work.

\subsubsection{Gaussian Process Approximation}
\label{GaussianMatern}
In this case, we follow the framework described in Subsection \ref{General Convex} and use the two-step method as Subsection \ref{subsubsec:two-step}. 

\noindent
\textbf{Experimental Setup.} In this example, we use the Mat\'{e}rn kernel 
% $K(x, y)=(1+\frac{\sqrt{5} d}{\rho}+\frac{5 d^2}{3 \rho^2}) \exp (-\frac{\sqrt{5} d}{\rho})$ 
$K(x, y)=\sigma^2(1+\frac{\sqrt{5} d}{\rho}+\frac{5 d^2}{3 \rho^2}) \exp (-\frac{\sqrt{5} d}{\rho})$ 
to approximate the unknown functions $F$ when $(\rho=1, \sigma=1,d=\sqrt{(x-y)^2+\epsilon})$, and remains unchanged across all subsequent experiments. The grid size is set to \( h = \frac{1}{50} \). For observations, 625 points \( \boldsymbol{m}^o \) are selected from the total 2500 sample points of \( m \). These observation points are sampled from the grid, while the 625 observation points for \( V \) are randomly distributed in space and are not restricted to the grid. We set the regularization parameters as \( \alpha_{m^o} = 10^5 \) and \( \alpha_{v^o} = \infty \), indicating that the exact values of \( V = V^o \) are known at the observation points. The Gaussian regularization coefficient is set to \( \alpha_v = 10 \), with \( \alpha_f = 0.7 \) and \( \alpha_{fp} = 8 \times 10^2 \). Gaussian noise \( \mathcal{N}(0, \gamma^2 I) \) with a standard deviation of \( \gamma = 10^{-3} \) is added to the observations. The initial value for \( V \) is set to zero. 

In our experiments, the fixed points \( \{r_i\}_{i=1}^{N_F} \) are evenly spaced between 0.5 and 2, with \( N_F = 10 \) points. Similarly, the fixed points \( \{\widetilde{r}_i\}_{i=1}^{N_{MC}} \) are distributed between 0 and 1.5, with \( N_{MC} = 200 \) points. Since we expect \( \boldsymbol{z}_F \) to be positive, we set \( \boldsymbol{z}_F = \widetilde{\boldsymbol{z}}_F^2 \) and optimize for \( \widetilde{\boldsymbol{z}}_F \). The initial condition for \( \widetilde{\boldsymbol{z}}_F \) consists of equally spaced points over an interval chosen to ensure that its squared range covers the range of \( m \). 
% The length scale of the Mat\'ern kernel is set to be one and remains unchanged across all subsequent experiments.

\begin{figure}[h]
    \centering
    \begin{subfigure}[b]{0.33\textwidth}
        \centering
        \includegraphics[width=\linewidth]{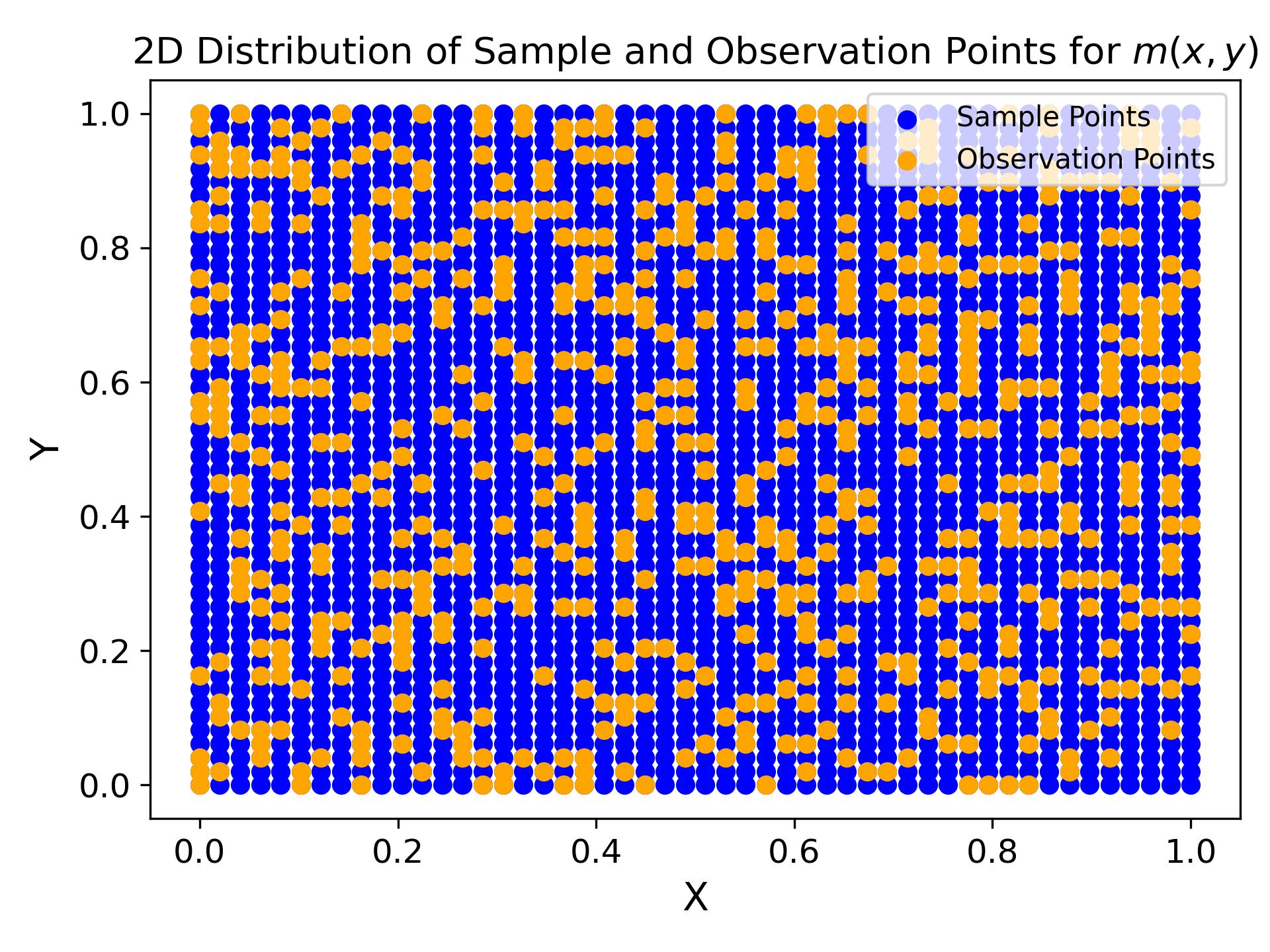}
        \caption{Samples \& Observations of $m$}
        \label{fig:msample1}
    \end{subfigure}%
    \begin{subfigure}[b]{0.33\textwidth}
        \centering
        \includegraphics[width=\linewidth]{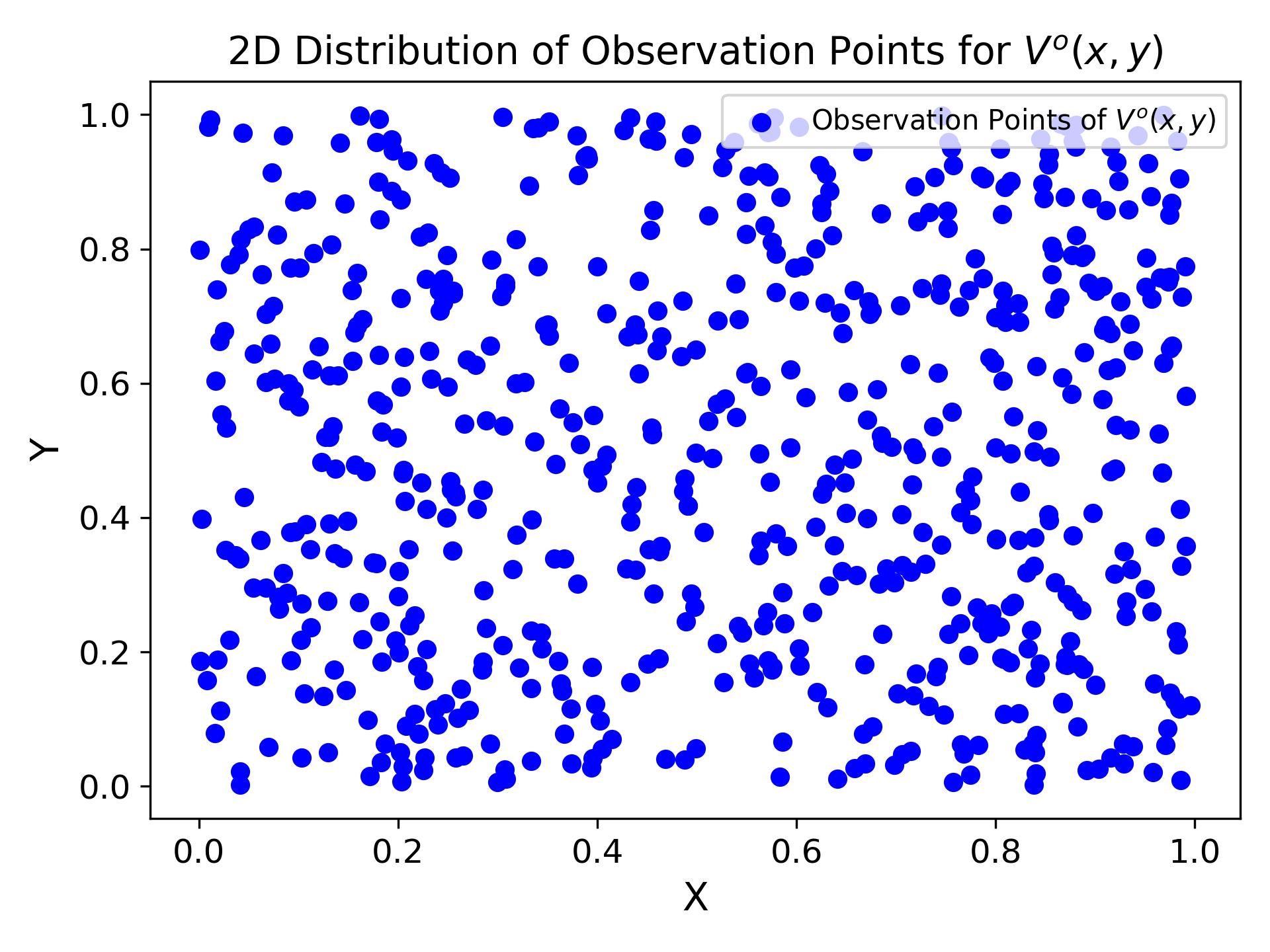}
        \caption{Observations of $V$}
        \label{fig:v01}
    \end{subfigure}%
    \begin{subfigure}[b]{0.33\textwidth}
        \centering
        \includegraphics[width=\linewidth]{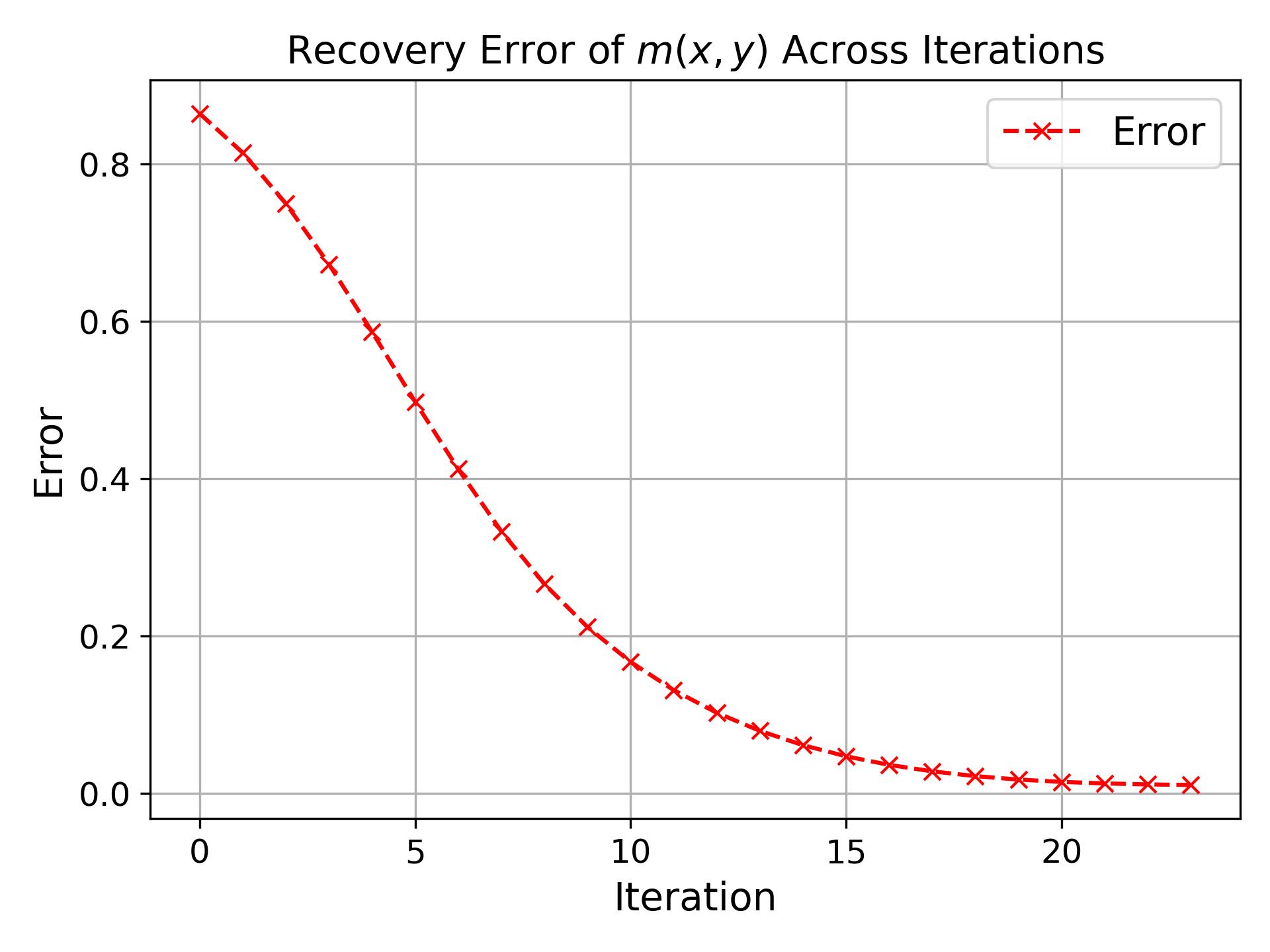}
        \caption{Error \( \mathcal{E}(m^k, m^*) \) vs.  Iteration  \( k \)}
        \label{fig:error2}
    \end{subfigure}
    \begin{subfigure}[b]{0.33\textwidth}
        \centering
        \includegraphics[width=\linewidth]{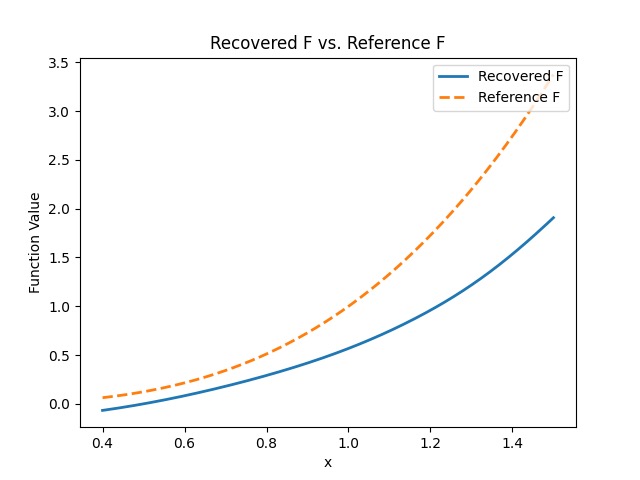}
        \caption{Recovered F vs. Reference F}
        \label{VfmaternF}
    \end{subfigure}
    \begin{subfigure}[b]{0.33\textwidth}
        \centering
        \includegraphics[width=\linewidth]{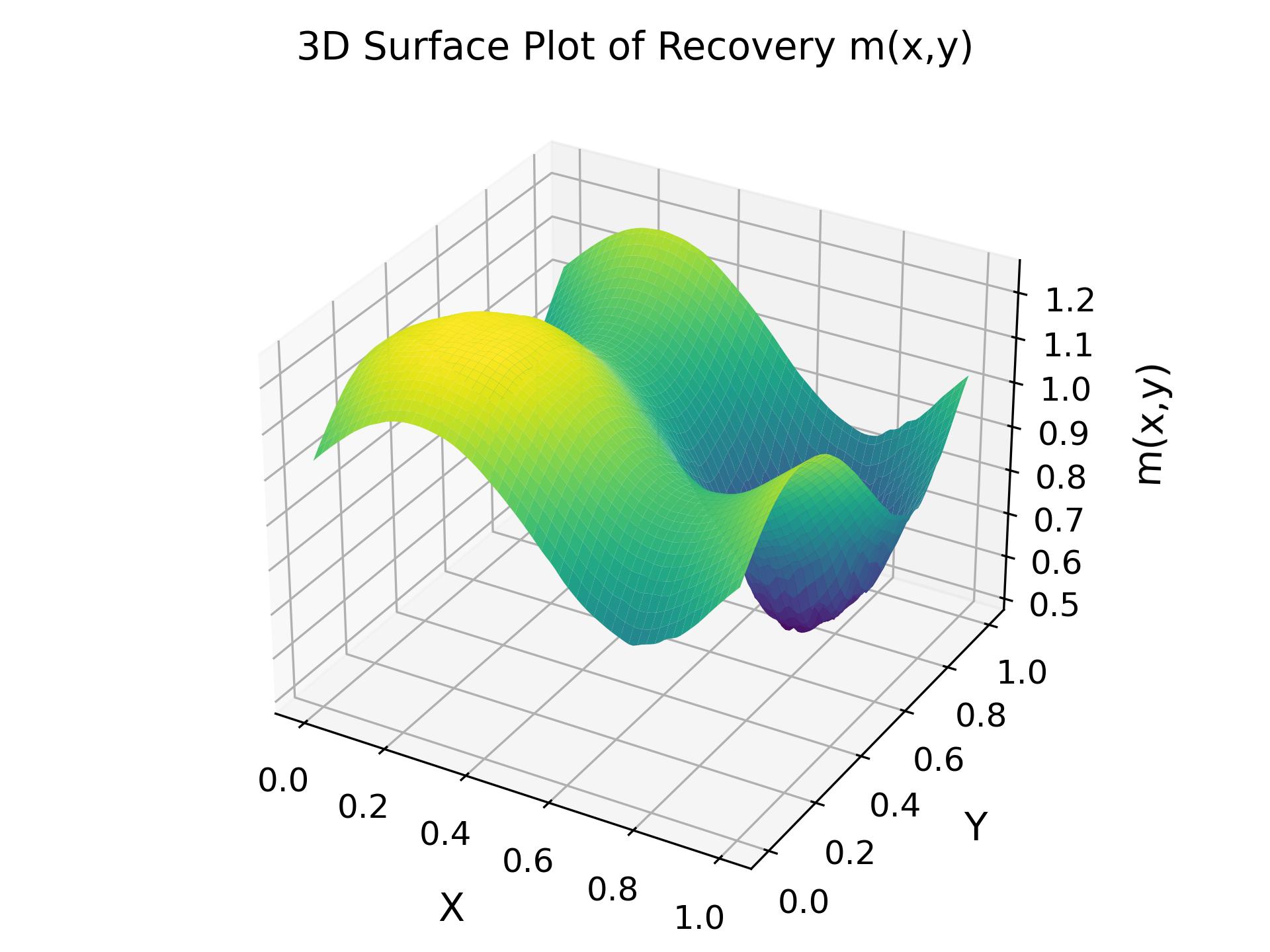}
        \caption{Recovered $m$}
        \label{fig:recover1}
    \end{subfigure}%
    \begin{subfigure}[b]{0.33\textwidth}
        \centering
        \includegraphics[width=\linewidth]{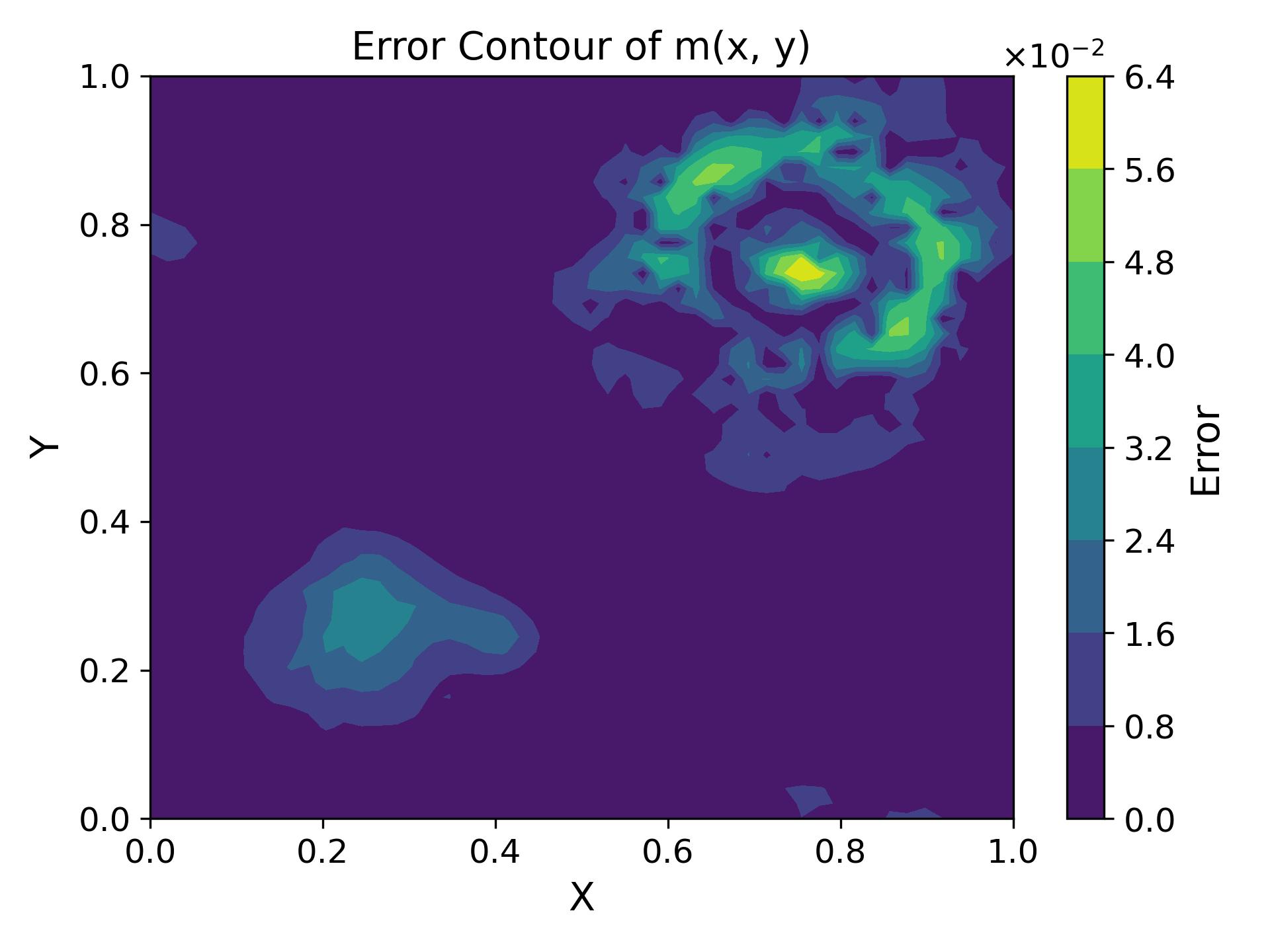}
        \caption{Error Contour of $m$}
        \label{fig:errorcontour1}
    \end{subfigure}
    \begin{subfigure}[b]{0.33\textwidth}
        \centering
        \includegraphics[width=\linewidth]{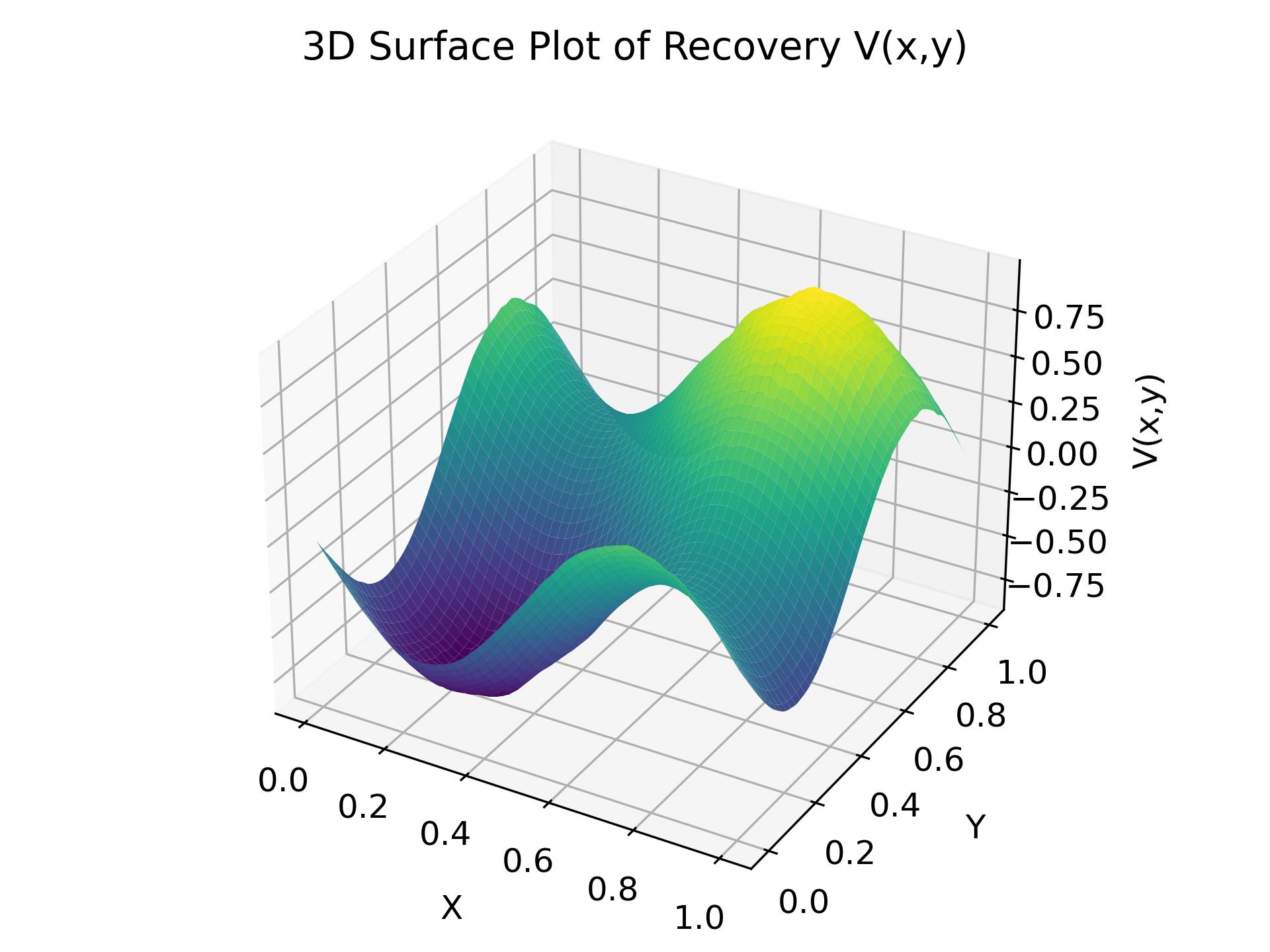}
        \caption{Recovered $V$}
        \label{fig:v01}
    \end{subfigure}%
    \begin{subfigure}[b]{0.33\textwidth}
        \centering
        \includegraphics[width=\linewidth]{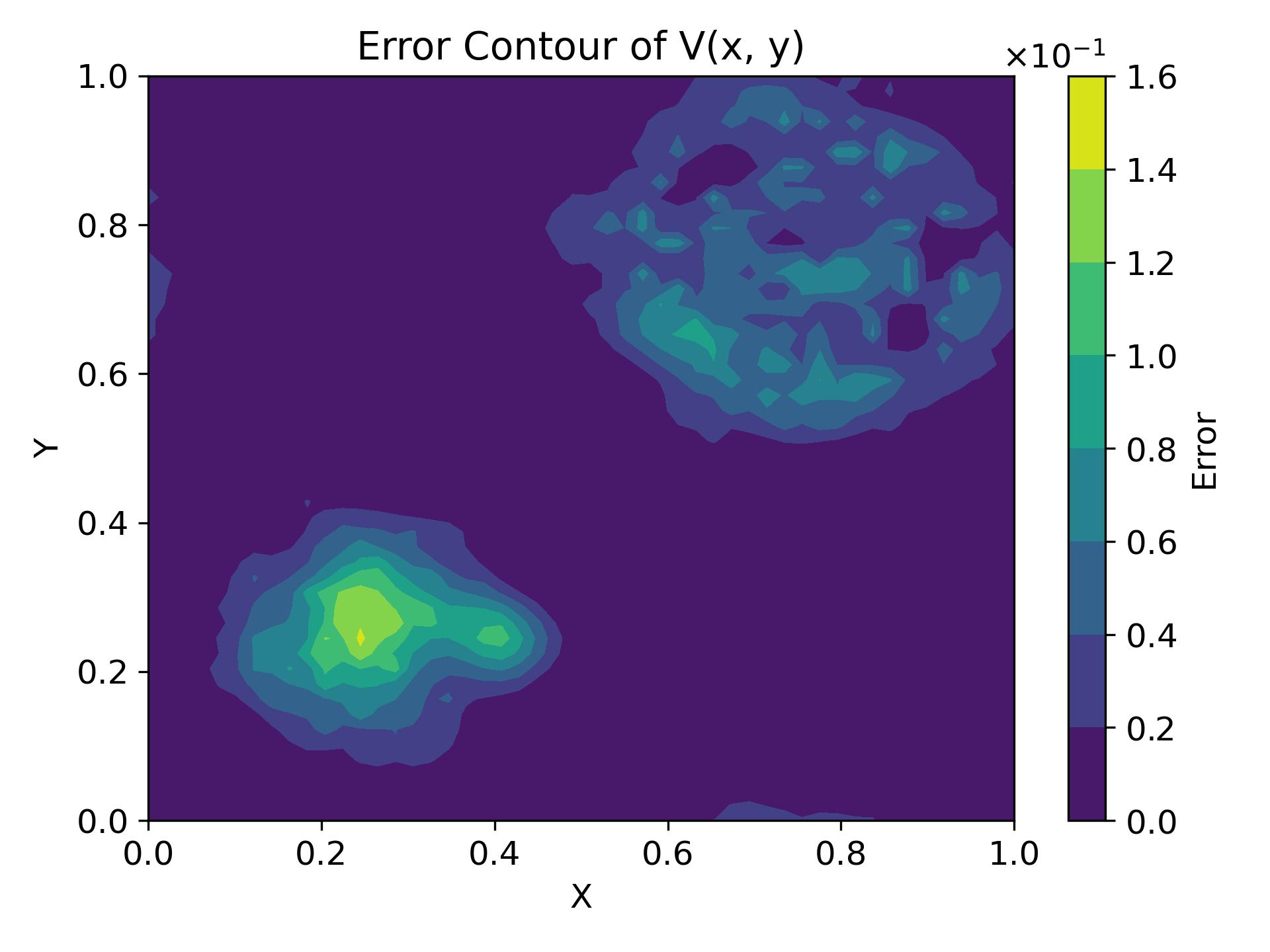}
        \caption{Error Contour of $V$}
        \label{fig:error2}
    \end{subfigure}
    
    \caption{Numerical results of recovering the coupling \(F\) using a GP for the MFG system in \eqref{General} are summarized as follows:  
(a) sample (grid) points and observation points of \( m \);  
(b) observation points of \( V \);  
(c) discretized \( L^2 \) error \( \mathcal{E}(m^k, m^*) \) versus iteration number \( k \); 
(d) recovered \( F \) vs. reference \( F \); 
(e) recovered \( m \);  
(f) pointwise error between the recovered \( m \) and the exact \( m^* \);   
(g) recovered \( V \);  
(h) pointwise error between the recovered \( V \) and its exact solution. 
Although the recovered \(F\) differs from the ground truth, the corresponding MFG solution closely matches the observed data. Furthermore, due to the introduction of the convexity penalization term, the GP-parametrized coupling function remains convex. 
}
    \label{figgeneralmatern}
\end{figure}

\textbf{Experiment Results.} 
Figure \ref{figgeneralmatern} presents the collocation grid points for \( m \) and the observation points for both \( m \) and \( V \). It also displays the discretized \( L^2 \) errors \(\mathcal{E}(m^k, m^*)\) from \eqref{eq:l2disc}, measuring the discrepancy between the approximate solution \( m^k \) and the exact solution \( m^* \) during CP iterations. Additionally, the figure includes pointwise error contours for the approximated \( m \) and \( V \).

As shown in Figure \ref{VfmaternF}, similar to the results from the convex function library method, The recovered function \(F\) does not match the ground truth. This is because the GP model for approximating \( F \) assumes only convexity, without enforcing any polynomial structure, thereby losing even more prior information than the convex function library approach. However, the inner minimization in the bilevel formulation always yields an MFG solution, meaning the recovered quantities fit the data well but do not necessarily match the true parameters. 

Meanwhile, Figure \ref{VfmaternF} also illustrates that the convexity penalization term in \eqref{biorp} effectively enforces the convexity of \( F \) and ensures that the recovered coupling satisfies the Lasry--Lions monotonicity condition. Consequently, the recovered quantities provide a surrogate MFG that is consistent with the observed data and helps characterize the system's behavior.

\subsection{Recovering Euclidean Metrics, Spatial Costs, and Couplings Using The Bilevel Framework}
\label{subsec:num:LambdaVF}
\begin{figure}[h]
    \centering
    \begin{subfigure}[b]{0.33\textwidth}
        \centering
        \includegraphics[width=\linewidth]{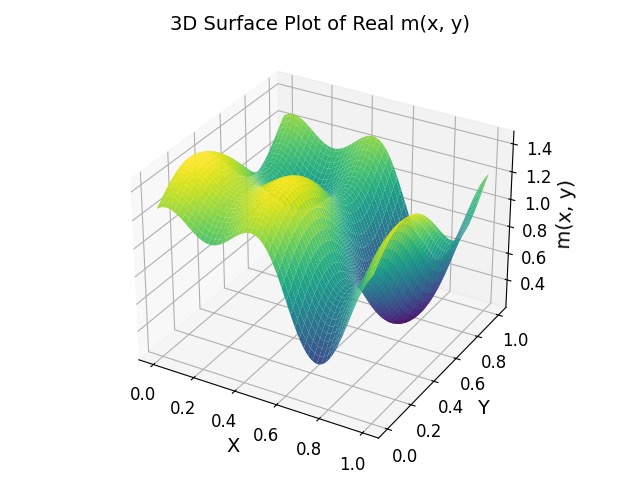}
        \caption{Reference Solution $m^*$}
        \label{figrealvfQ}
    \end{subfigure}%
    \begin{subfigure}[b]{0.33\textwidth}
        \centering
        \includegraphics[width=\linewidth]{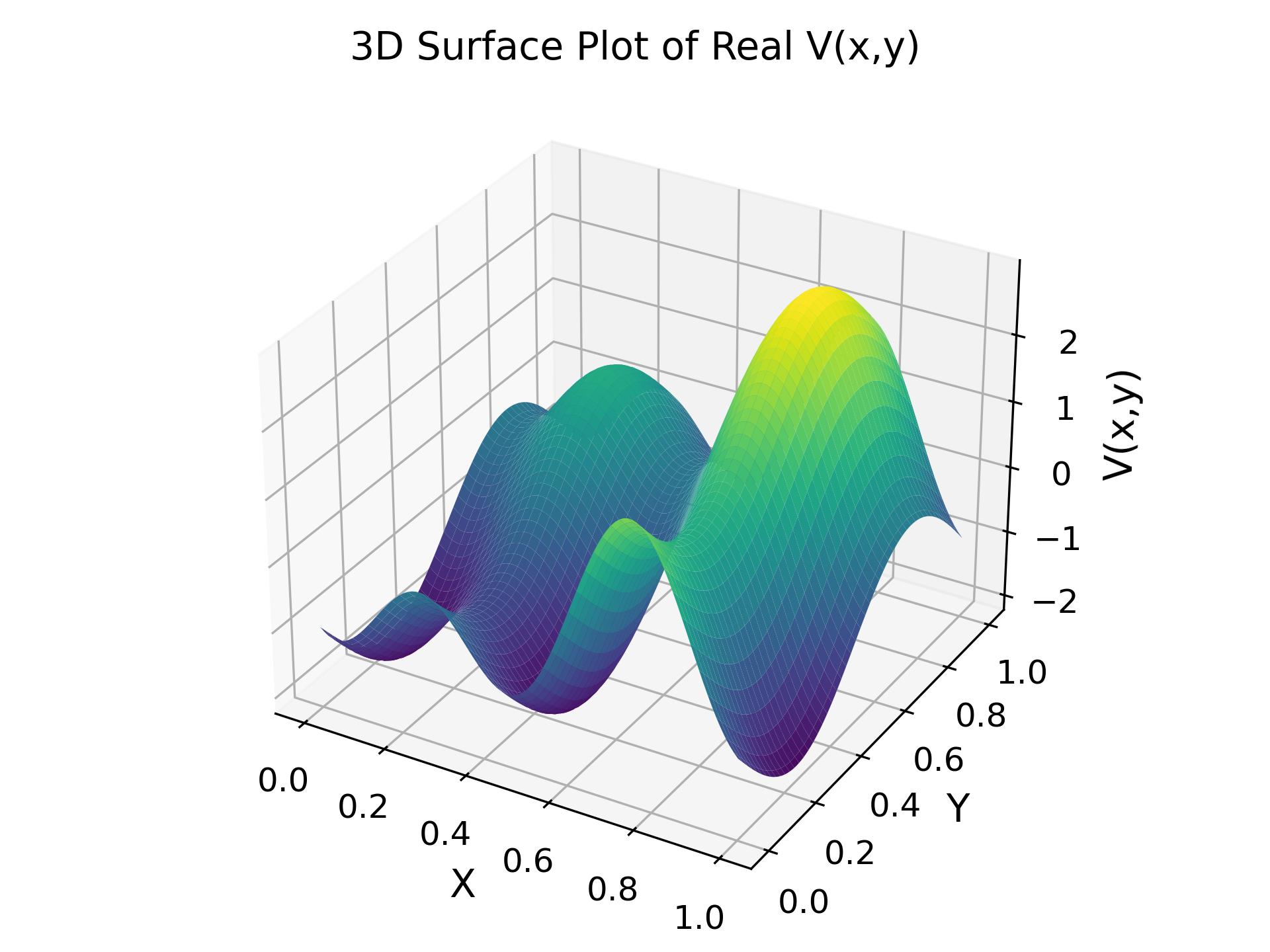}
        \caption{Ground Truth $V$}
        \label{fig:msample1}
    \end{subfigure}%%    \end{subfigure}
\caption{Reference results of $m$ and $V$ for the MFG \eqref{Viscosity}. 
}
\label{ReferenceViscosity}
\end{figure}

In this example, we study inverse problems related to the following stationary MFG system  
\begin{equation}
\begin{cases}
-\nu \Delta u + H(\nabla u) - \lambda = f(m) + V(x, y), & \forall (x, y) \in \mathbb{T}^2, \\ 
-\nu \Delta m - \operatorname{div}\left( D_p H(\nabla u) \, m \right) = 0, & \forall (x, y) \in \mathbb{T}^2, \\
\int_{\mathbb{T}^2} u \, \dif x \, \dif y = 0, \quad \int_{\mathbb{T}^2} m \, \dif x \, \dif y = 1.
\end{cases}
\label{Viscosity}
\end{equation}
Here, the Legendre transform of the Hamiltonian is given by \( H^*(q) = \frac{1}{2} |\Lambda q|^2 \), where \( \Lambda = I \), corresponding to a Euclidean metric for measuring kinetic energy. The spatial cost function is defined as \( V(x, y) = -(\sin(2 \pi x) + \cos(4 \pi x) + \sin(2 \pi y)) \). The coupling function is \( f(m) = m^3 \), and the viscosity coefficient is set to \( \nu = 0.1 \). For accuracy comparison, the reference solution \( m^* \) is computed using the CP algorithm from \cite{briceno2018proximal} as shown in Figure \ref{figrealvfQ}.
In this subsection, we focus on recovering the distribution \( m \) along with the unknown Euclidean metric matrix \( \Lambda \) in the Hamiltonian and the unknown functions \( f \) and \( V \), based on partial noisy observations of \( m \) and \( V \).

% $$
% \begin{cases}-\nu \Delta u+H(x, \nabla u)-\lambda=logm+V(x), & \forall x \in \mathbb{T}^2, \\ -\nu \Delta m-\operatorname{div}\left(D_p H(x, \nabla u) m\right)=0, & \forall x \in \mathbb{T}^2, \\ \int_{\mathbb{T}^2} u d x=0, \int_{\mathbb{T}^2} m d x=1, & \end{cases}
% $$

\subsubsection{Estimation of \(F\) as a Power Function}
\label{PowerQ}
For the initial approach to solving the inverse problem, we follow the framework proposed in Subsection  \ref{Powerfunctionmethod}. 

\textbf{Experimental Setup.} The grid size is set to \( h = \frac{1}{59} \). From the total 3,481 sample points of \( m \), 386 observation points are selected from the grid, while the 400 observation points for $V$ are randomly distributed in space without grid constraints. The regularization parameters are set to \( \alpha_{m^o} = 10^6 \) and \( \alpha_{v^o} = \infty \). The Gaussian regularization coefficient is \( \alpha_v = 0.1 \), with \( \alpha_{\beta} = 1 \) and \( \alpha_{\lambda} = 2 \times 10^5 \). Gaussian noise with a standard deviation \( \gamma = 10^{-3} \) is added to these observations, modeled as \( \mathcal{N}(0, \gamma^2 I) \). 

Note that in Section \ref{sec:Prerequisites}, when discretizing the Hamiltonian, the momentum \( w \) is represented using four components. Consequently, the matrix \( \Lambda \) is approximated as a \( 4 \times 4 \) matrix, with each element modeled as a GP variable.

\begin{figure}[h]
    \centering
    \begin{subfigure}[b]{0.33\textwidth}
        \centering
        \includegraphics[width=\linewidth]{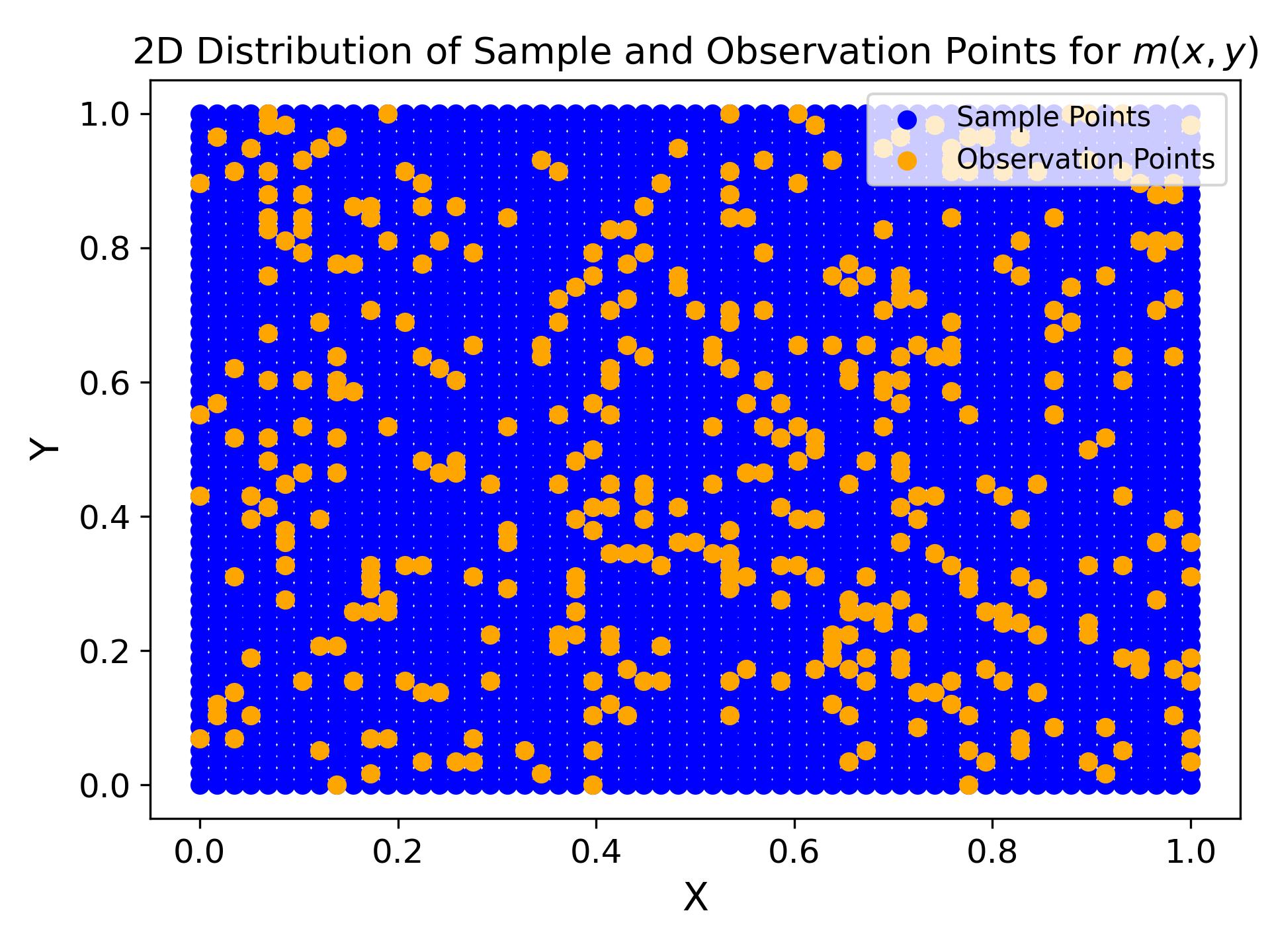}
        \caption{Samples \& Observations of $m$}
        \label{fig:msample1}
    \end{subfigure}%
    \begin{subfigure}[b]{0.33\textwidth}
        \centering
        \includegraphics[width=\linewidth]{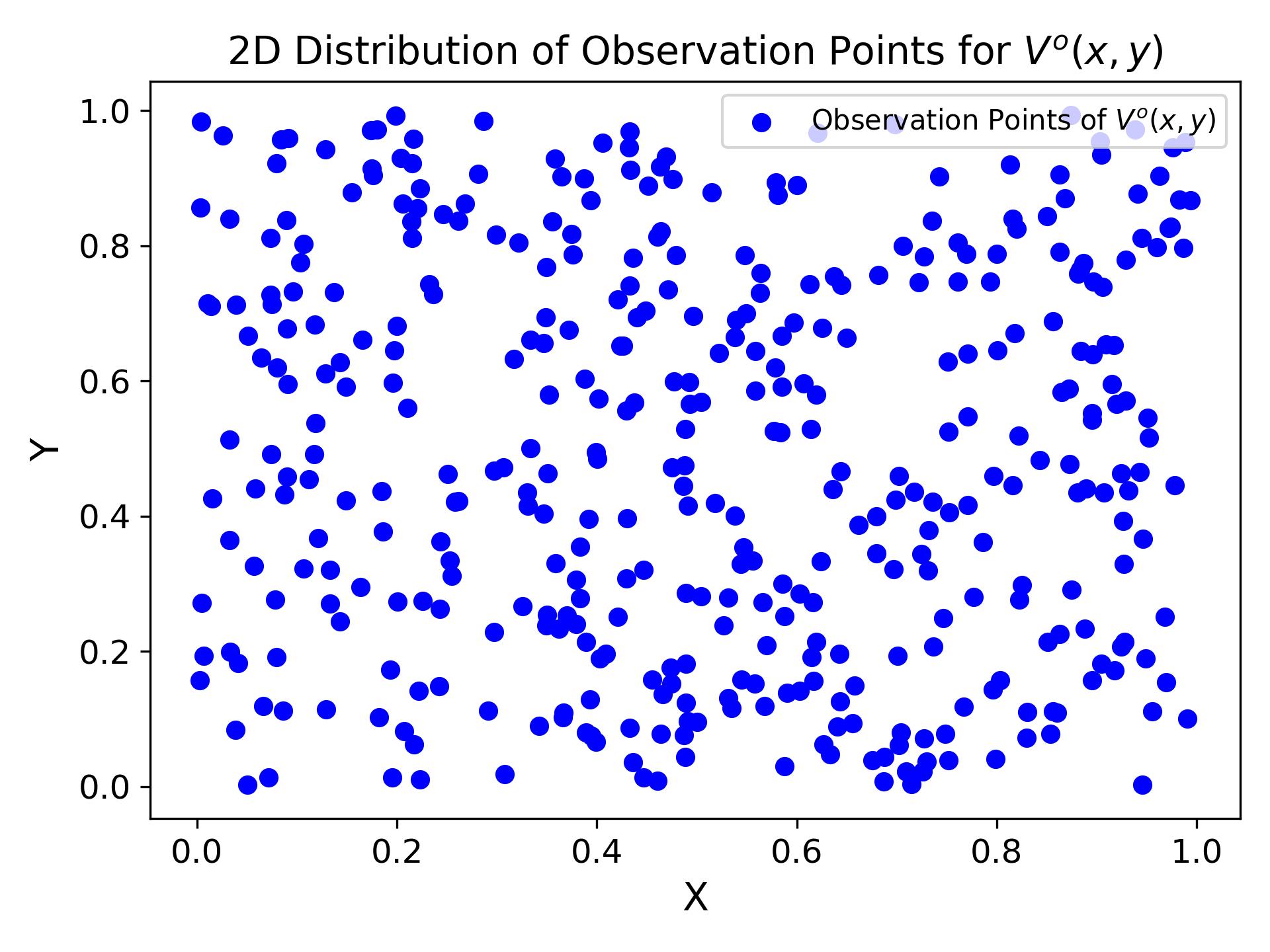}
        \caption{Observations of $V$}
        \label{fig:v01}
    \end{subfigure}%
    \begin{subfigure}[b]{0.33\textwidth}
        \centering
        \includegraphics[width=\linewidth]{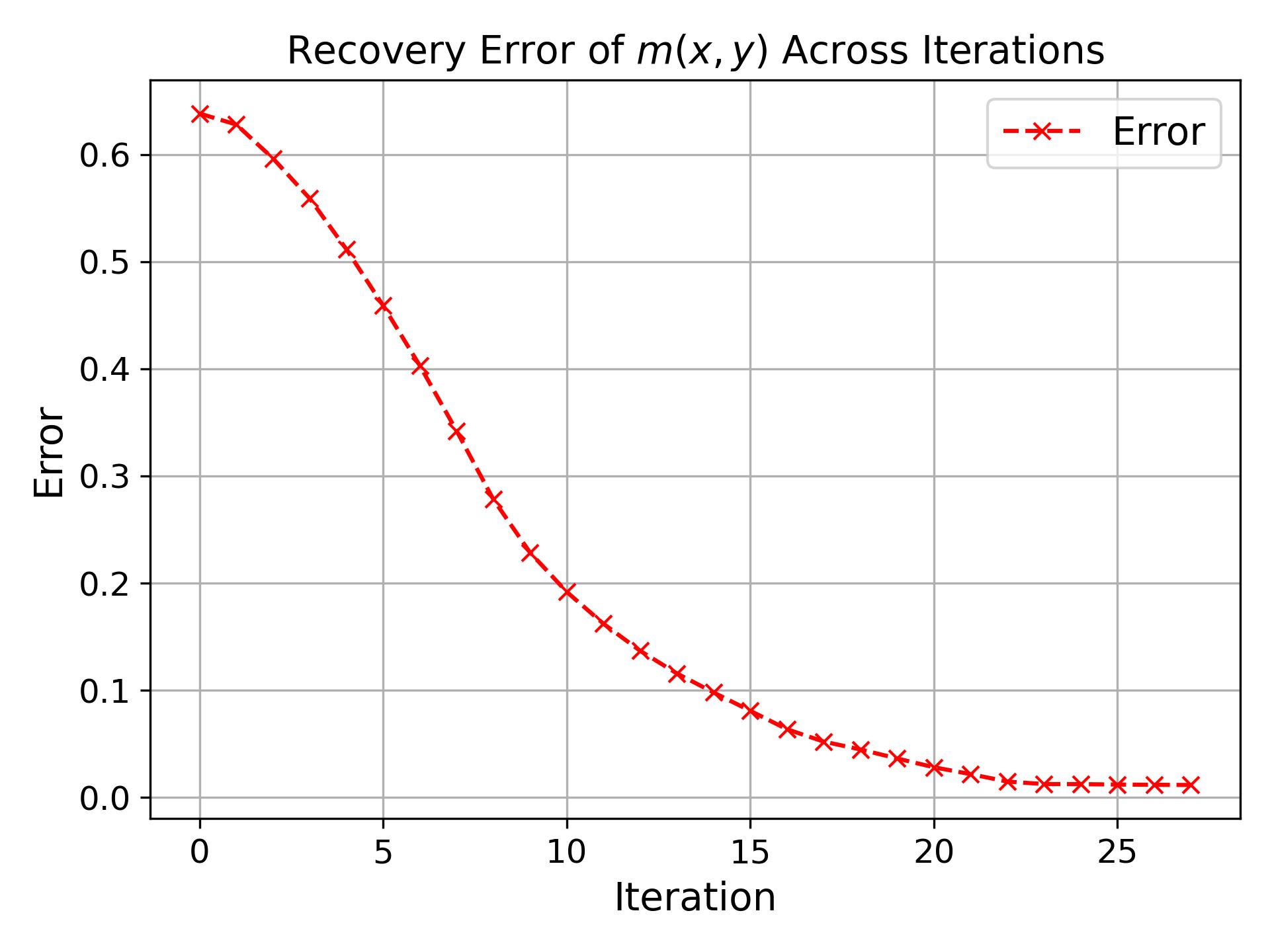}
        \caption{Error \( \mathcal{E}(m^k, m^*) \) vs.  Iteration  \( k \)}
        \label{fig:error2}
    \end{subfigure}
  \begin{subfigure}[b]{0.32\textwidth}
        \includegraphics[width=\linewidth]{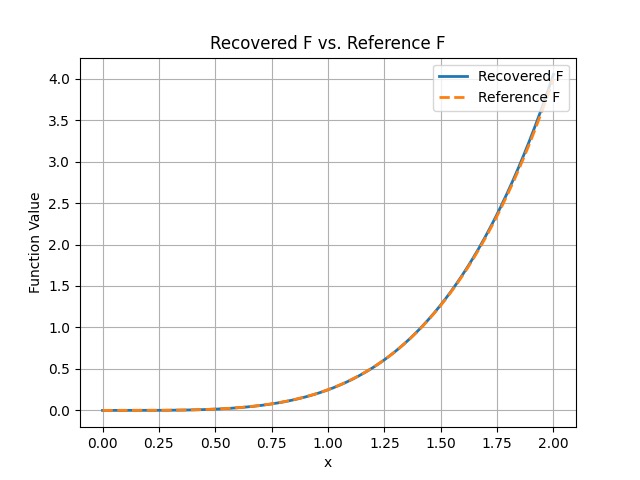}
    \caption{Recovered F vs. Reference F.}
    \label{VfbetaQF}
    \end{subfigure}
    \begin{subfigure}[b]{0.33\textwidth}
        \centering
        \includegraphics[width=\linewidth]{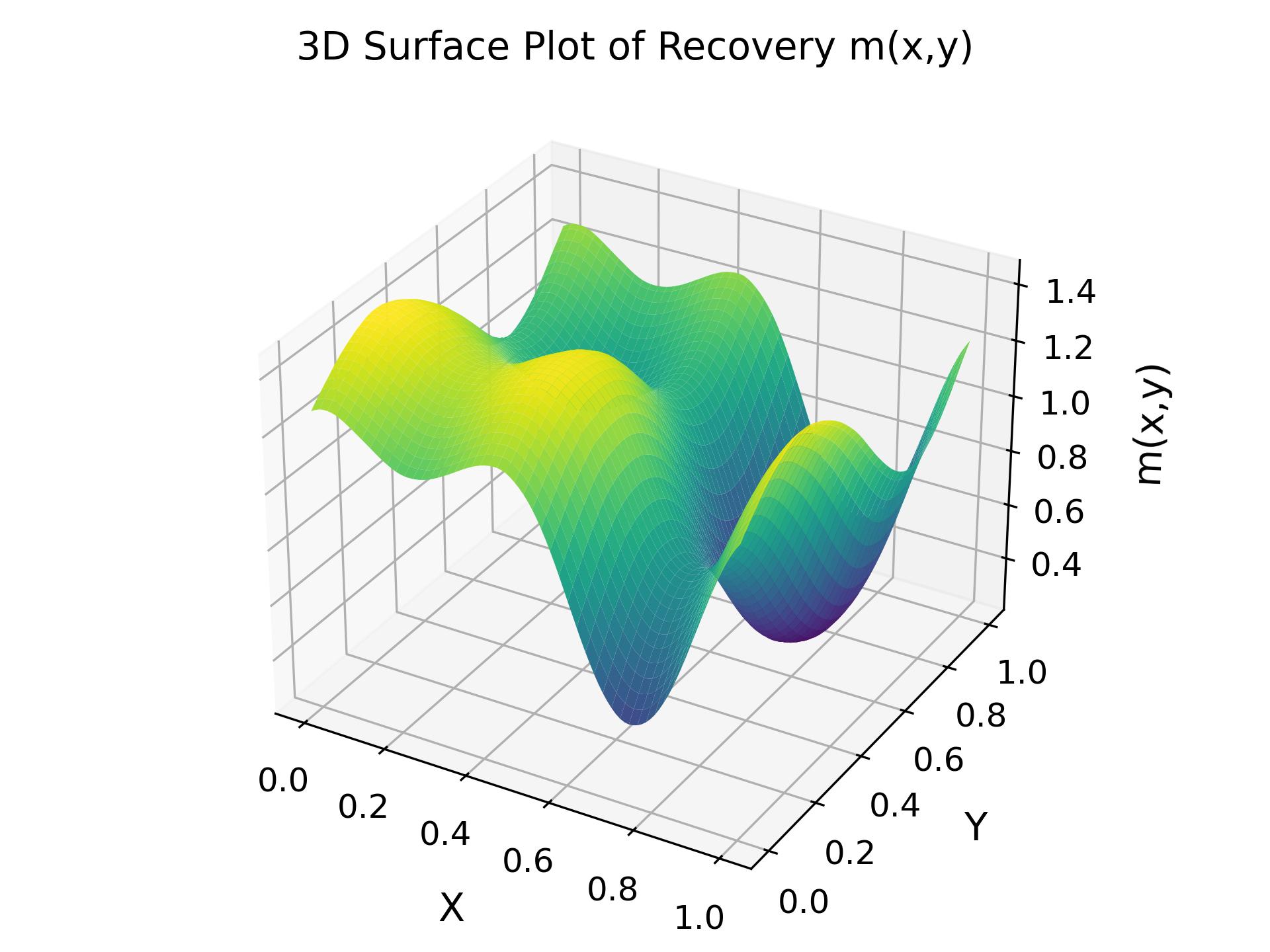}
        \caption{Recovered $m$}
        \label{fig:recover1}
    \end{subfigure}
    \begin{subfigure}[b]{0.33\textwidth}
        \centering
        \includegraphics[width=\linewidth]{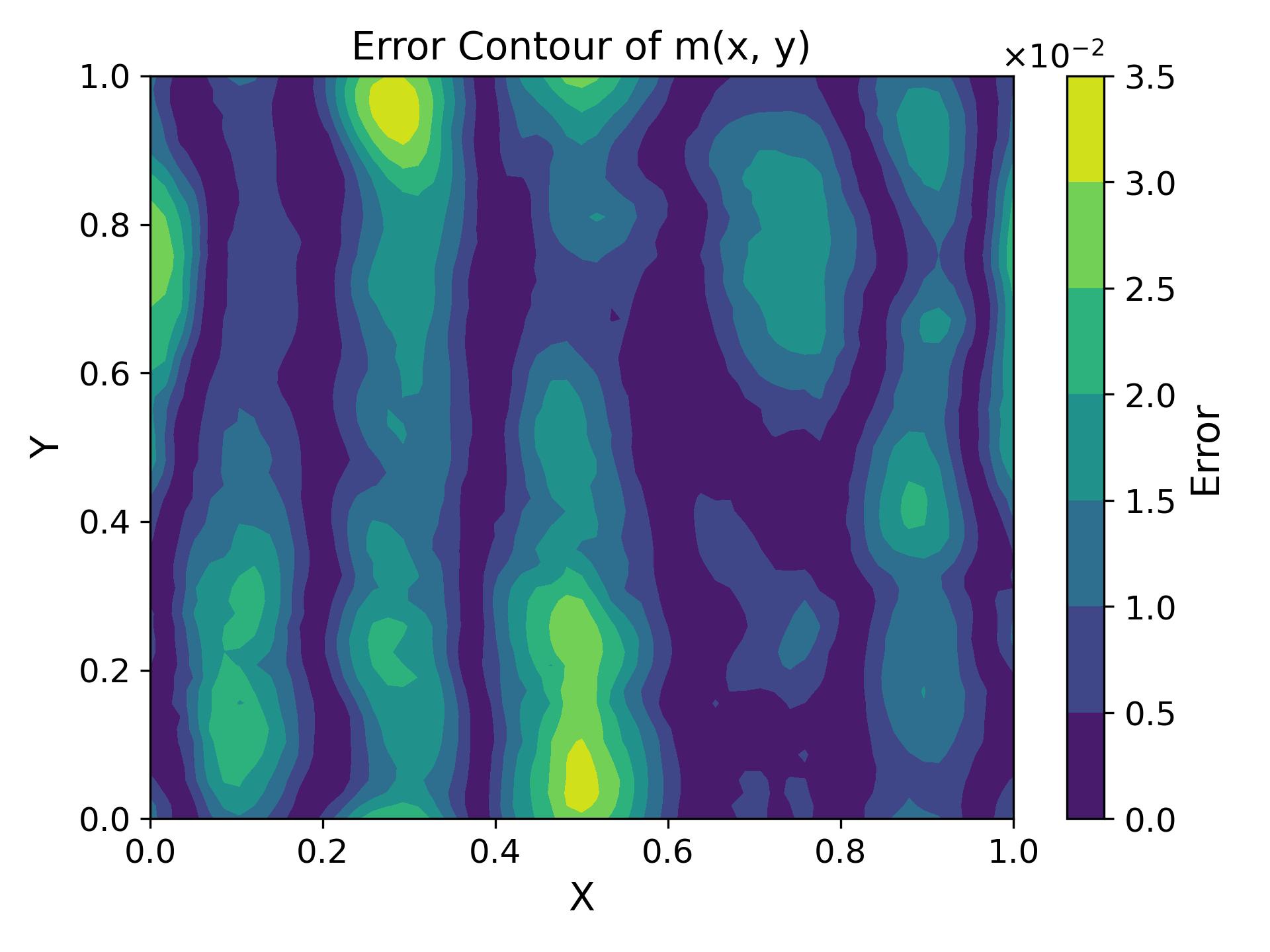}
        \caption{Error Contour of $m$}
        \label{fig:errorcontour1}
    \end{subfigure}
   \begin{subfigure}[b]{0.33\textwidth}
        \centering
        \includegraphics[width=\linewidth]{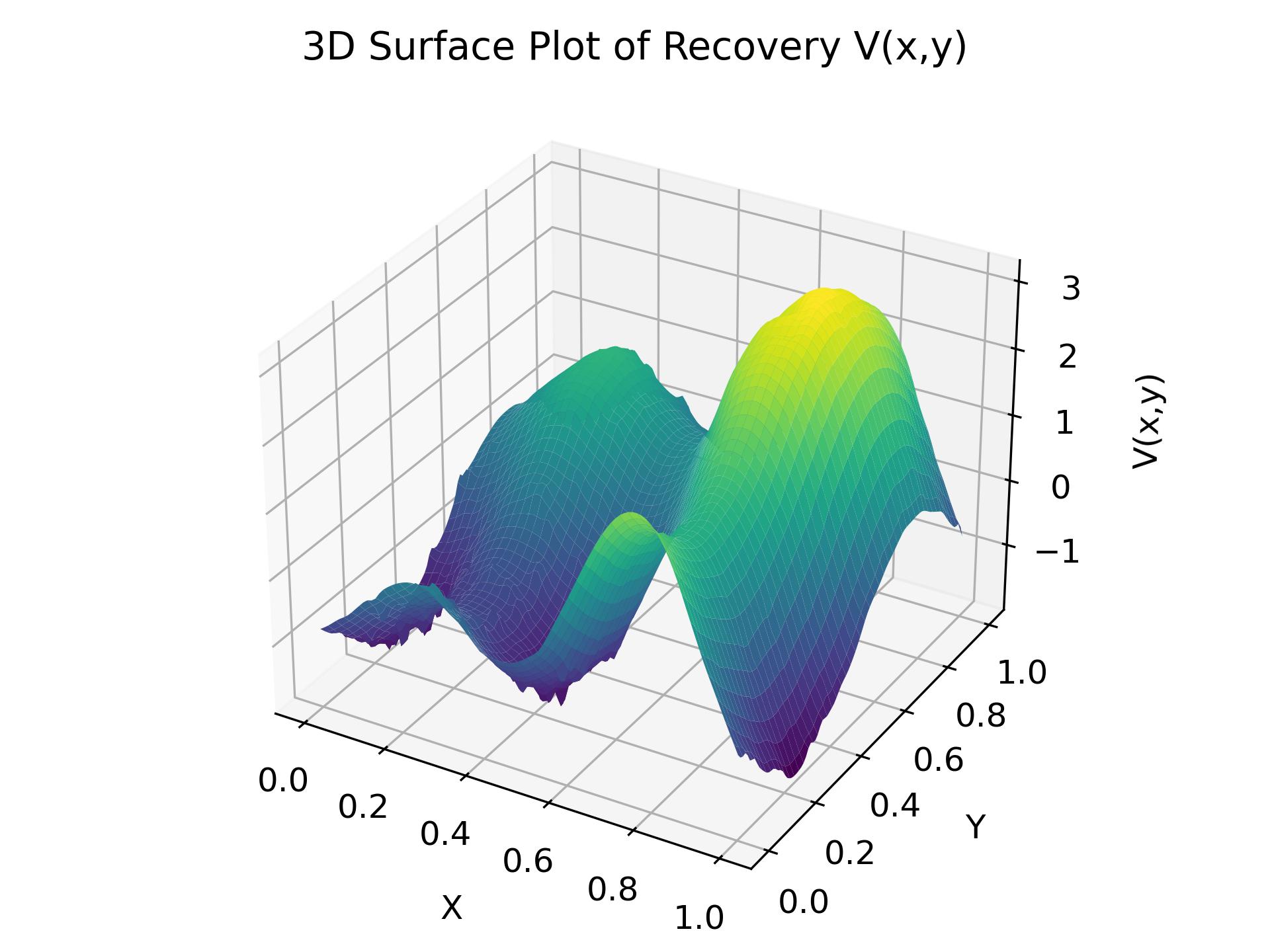}
        \caption{Recovered $V$}
        \label{fig:v01}
    \end{subfigure}%
    \begin{subfigure}[b]{0.33\textwidth}
        \centering
        \includegraphics[width=\linewidth]{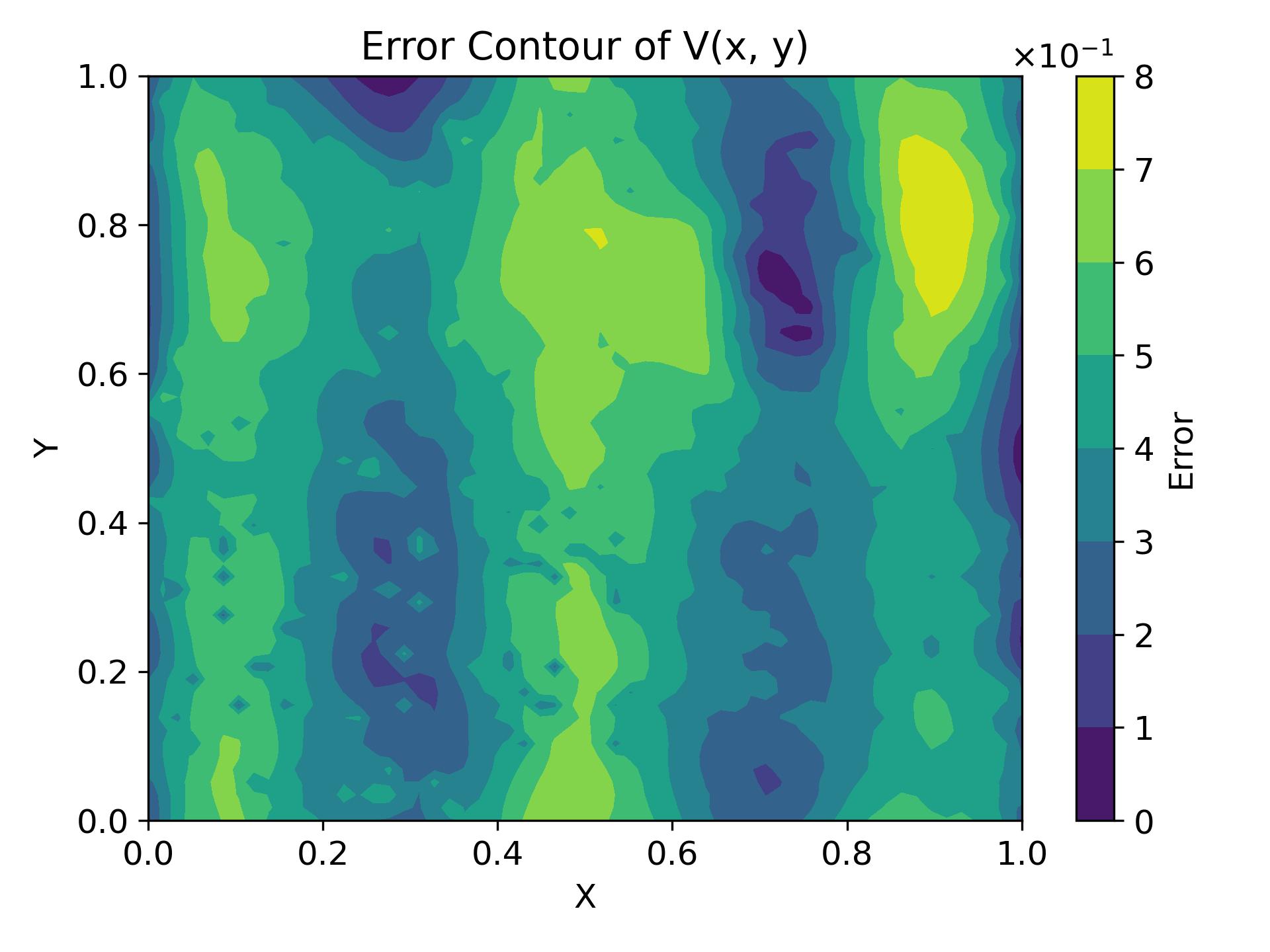}
        \caption{Error Contour of $V$}
        \label{fig:error2}
    \end{subfigure}
    \caption{Numerical results for recovering the Euclidean metric and the coupling \( F \) using a power function for the MFG system in \eqref{Viscosity}:  
(a) sample (grid) points and observation points of \( m \);  
(b) observation points of \( V \);  
(c) discretized \( L^2 \) error \( \mathcal{E}(m^k, m^*) \) versus iteration number \( k \);  
(d) recovered \( F \) vs. reference \( F \);  
(e) the recovered \( m \);  
(f) pointwise error between the recovered \( m \) and the exact \( m^* \);  
(g) the recovered \( V \); and  
(h) pointwise error between the recovered \( V \) and its exact solution.
}
    \label{figfbetaQ}
\end{figure}

\textbf{Experiment Results.} 
Figure \ref{figfbetaQ} presents the sample points (collocation grid points) for \( m \) and the observation points for both \( m \) and \( V \). It also shows the discretized \( L^2 \) errors \( \mathcal{E}(m^k, m^*) \) defined in \eqref{eq:l2disc}, representing the error between the approximated solution \( m^k \) at the \( k \)-th iteration and the exact solution \( m^* \), as defined in \eqref{eq:general:explcit}, during the CP iterations. The figure includes the true solutions, the recovered results, and the pointwise error contours of the approximated functions of both \( m \) and \( V \). 

To ensure that \( \alpha \geq 1 \) for the monotonicity of \( m^{\alpha} \), we parameterize \( \alpha \) as \( \alpha = \ln(e^{\beta} + 1) + 1 \) and optimize \( \beta \) instead. The recovered value of \( \beta \) is 1.8873, which closely matches the true value of 1.8523 when \( \alpha = 3 \). Figure \ref{VfbetaQF} illustrates the recovered \( F \), demonstrating that \( F \) is accurately reconstructed. 

However, the recovered matrix \( \Lambda \) deviates from the ground truth, suggesting that alternative parameterizations can yield solutions that fit the data well. The recovered \( \Lambda \) is:
\[
\Lambda=\begin{bmatrix}
0.9104 & 0.0956 & 0.1975 & -0.0017 \\
0.2263 & 0.9670 & -0.0432 & 0.2084 \\
0.1778 & -0.0056 & 0.6477 & 0.0667 \\
-0.0117 & 0.1939 & 0.0766 & 0.6891
\end{bmatrix}.
\]
The discrepancy in \( \Lambda \) likely arises from the fact that multiple metric matrices can produce similar observed dynamics in the MFG system. This underscores a fundamental challenge in recovering \( \Lambda \), as its influence on the system may not be uniquely identifiable given the available observations.

\subsubsection{The Convex Function Library Method With Known Spatial Costs}
\label{PolynomialQ}
In this experiment, we follow the framework outlined in Subsection \ref{General Convex} with known spatial costs. 

\textbf{Experimental Setup.} The grid size is set to \( h = \tfrac{1}{59} \). Among the 3,481 total grid points for \( m \), 20 points are selected as observation points. The regularization parameters are set to \(\alpha_{m^o} = 10^5\), with additional coefficients  \(\alpha_{fp} = 0.01\), \(\alpha_{\gamma_k} = 8 \times 10^3\), and \(\alpha_{\lambda} = 10^4\). Gaussian noise \(\mathcal{N}(0, \gamma^2 I)\) with \(\gamma = 10^{-3}\) is added to the observations. In this example, we assume the spatial cost function \( V \) is known.

Here, we approximate the coupling \( F \) as a sum of six polynomials, \(\sum_{k=1}^6 \gamma_k \, x^k\), initializing each \(\gamma_k\) to 1. The initial value for \(\Lambda\) is set as \(\Lambda = I + \boldsymbol{\xi}\), where \(\boldsymbol{\xi}\) is a matrix sampled from standard Gaussian variables. In our experiments, the points \(\{\widetilde{r}_i\}_{i=1}^{N_{\text{MC}}}\) are uniformly spaced between 0.3 and 1.5, with \( N_{\text{MC}} = 500 \) samples used to approximate the expectations in the convexity penalization term via Monte Carlo integration.

\begin{figure}[h]
    \centering
    \begin{subfigure}[b]{0.33\textwidth}
        \centering
        \includegraphics[width=\linewidth]{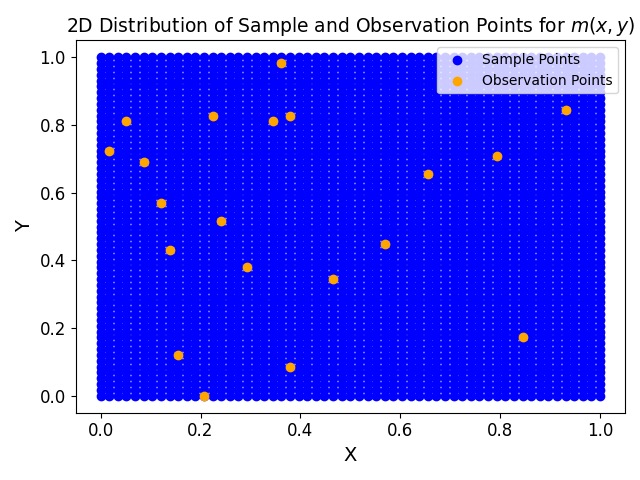}
        \caption{Samples \& Observations of $m$}
        \label{fig:msample1}
    \end{subfigure}%
    % \begin{subfigure}[b]{0.33\textwidth}
    %     \centering
    %     \includegraphics[width=\linewidth]{FigureVfQpoly/afig8VfQpoly0.jpg}
    %     \caption{Observations of $V$}
    %     \label{fig:v01}
    % \end{subfigure}%
    \begin{subfigure}[b]{0.33\textwidth}
        \centering
        \includegraphics[width=\linewidth]{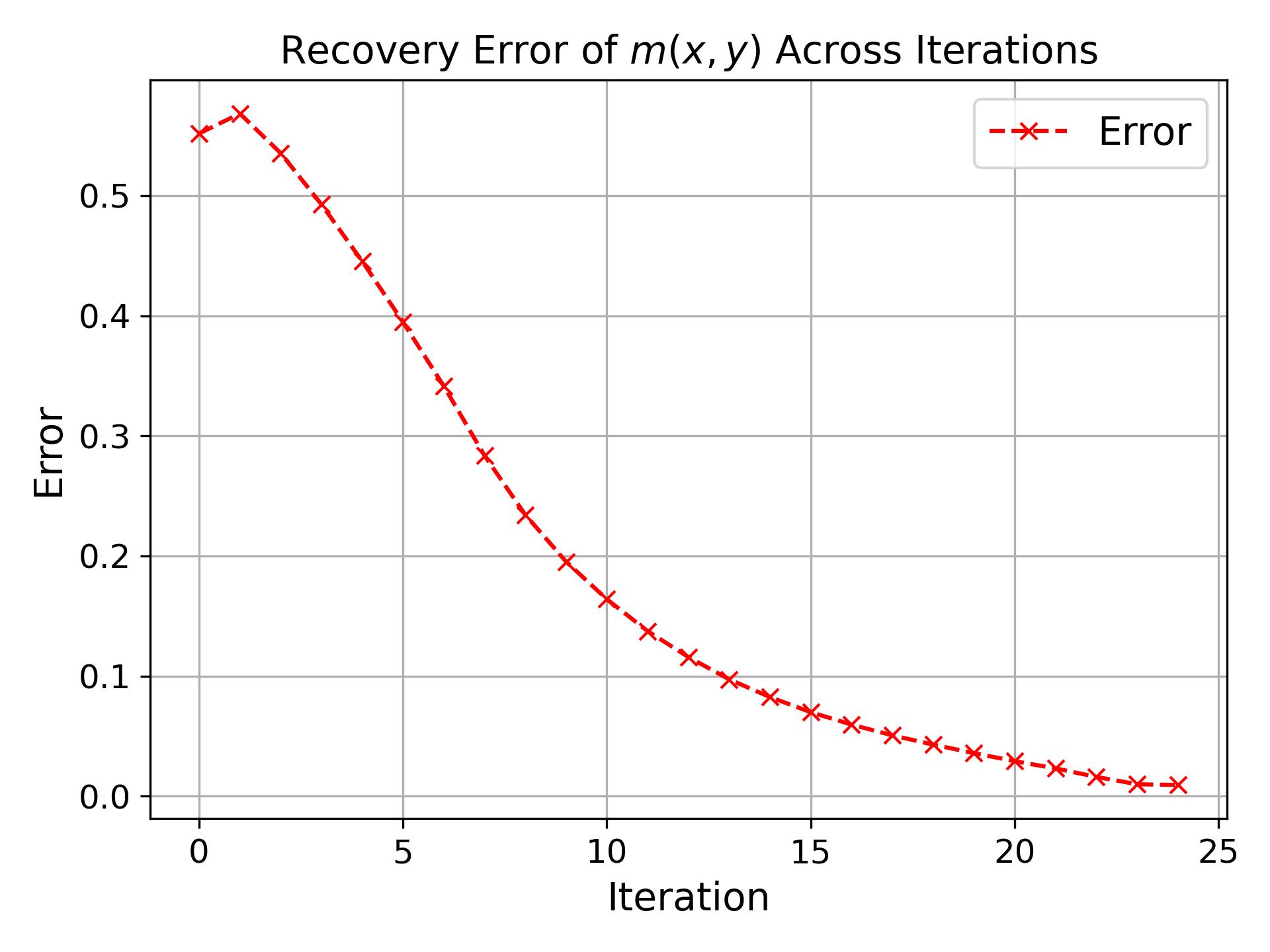}
        \caption{Error \( \mathcal{E}(m^k, m^*) \) vs.  Iteration  \( k \)}
        \label{fig:error2}
    \end{subfigure}
\begin{subfigure}[b]{0.33\textwidth}
        \includegraphics[width=\linewidth]{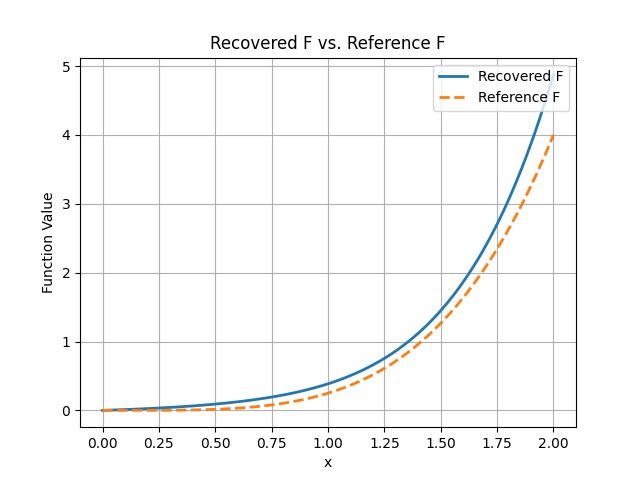}
    \caption{Recovered F vs. Reference F}
    \label{VfpolyQF}
    \end{subfigure}
    \begin{subfigure}[b]{0.33\textwidth}
        \centering
        \includegraphics[width=\linewidth]{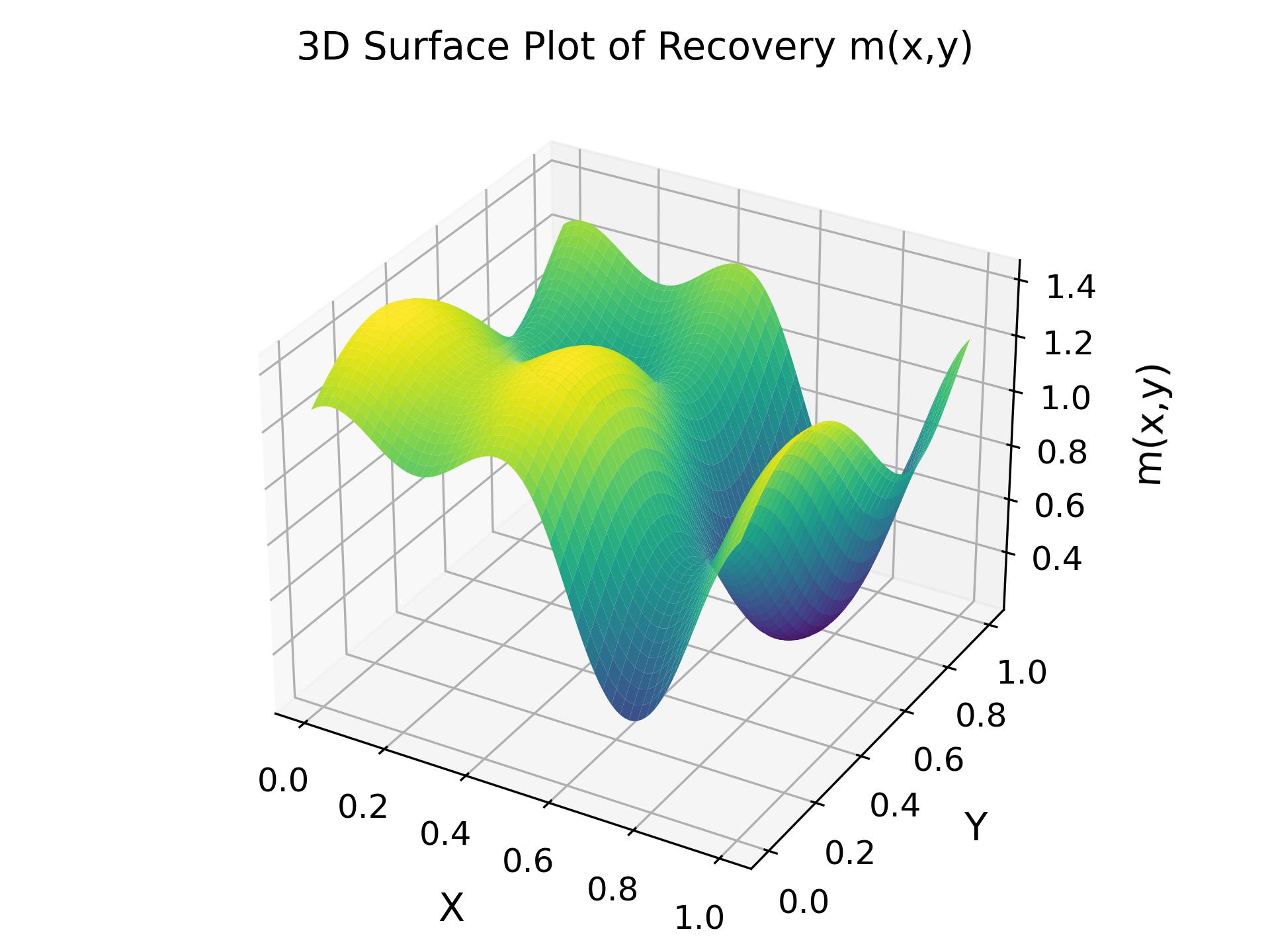}
        \caption{Recovered $m$}
        \label{fig:recover1}
    \end{subfigure}%
    \begin{subfigure}[b]{0.33\textwidth}
        \centering
        \includegraphics[width=\linewidth]{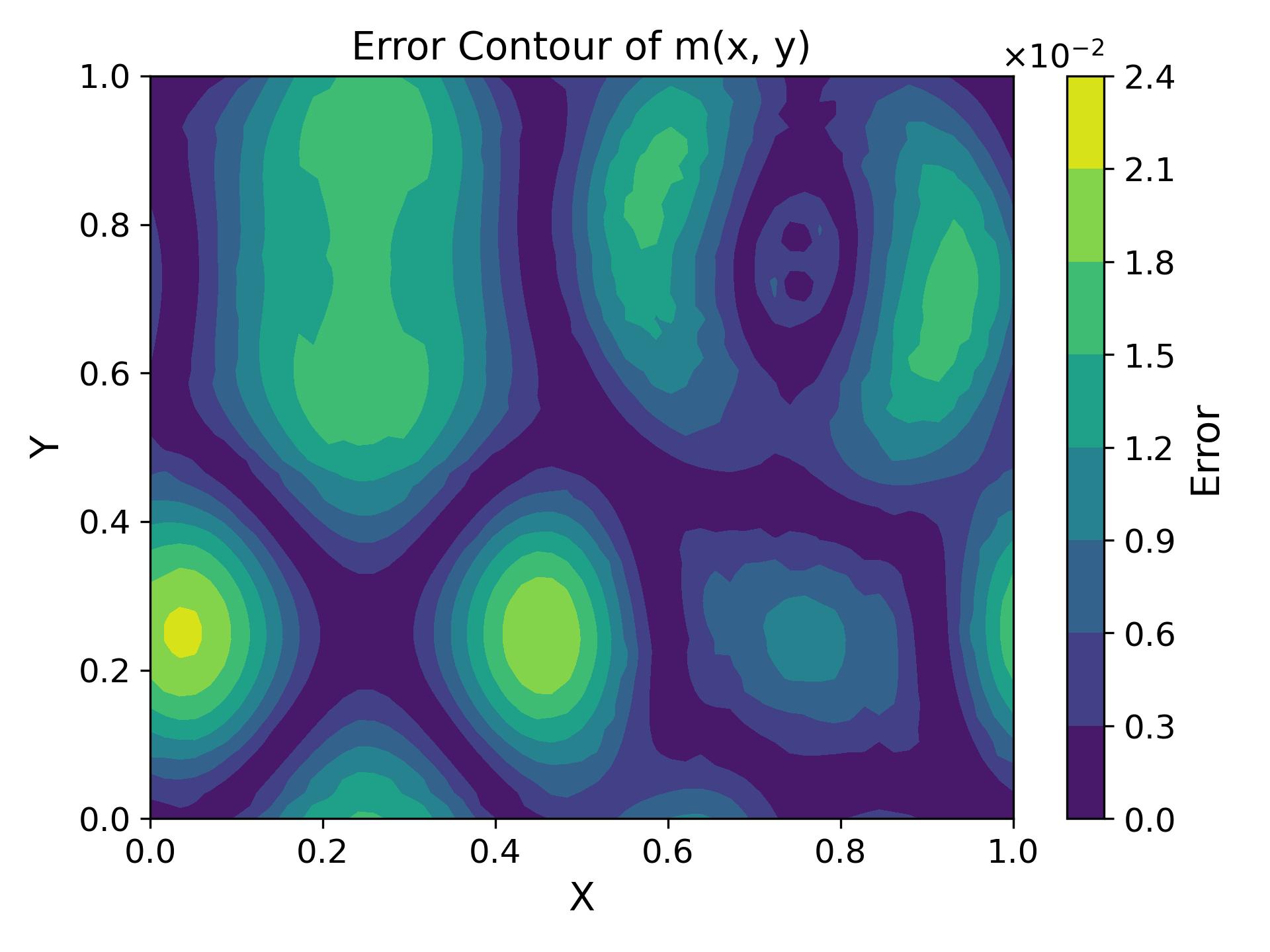}
        \caption{Error Contour of $m$}
        \label{fig:errorcontour1}
    \end{subfigure}
    % \begin{subfigure}[b]{0.33\textwidth}
    %     \centering
    %     \includegraphics[width=\linewidth]{FigureVfQpoly/afig4VfQpoly0.jpg}
    %     \caption{Recovered $V$}
    %     \label{fig:v01}
    % \end{subfigure}%
    % \begin{subfigure}[b]{0.33\textwidth}
    %     \centering
    %     \includegraphics[width=\linewidth]{FigureVfQpoly/afig5VfQpoly0.jpg}
    %     \caption{Error contour of $V$}
    %     \label{fig:error2}
    % \end{subfigure}
    
    \caption{
   Numerical results for solving the inverse problem of the MFG system in \eqref{Viscosity} using polynomial approximation for the coupling function \( F \):  
(a) sample (grid) points and observation points of \( m \);  
(b) discretized \( L^2 \) error \( \mathcal{E}(m^k, m^*) \) versus iteration number \( k \); 
(c) recovered \( F \) vs. reference \( F \); 
(d) recovered \( m \);  
(e) pointwise error between the recovered \( m \) and the exact \( m^* \).  Although the recovered \(F\) differs from the ground truth, it remains convex, and the resulting MFG solutions closely match the observed data.
% (g) recovered \( V \);  
% (h) pointwise error between the recovered \( V \) and its exact solution.
}
    \label{figVfQpoly}
\end{figure}

\textbf{Experiment Results.} 
Figure \ref{figVfQpoly} shows the collocation grid points for \( m \) and the observation points for  \( m \). It also depicts the discretized \( L^2 \) errors \(\mathcal{E}(m^k, m^*)\) from \eqref{eq:l2disc}, which measure the discrepancy between the approximate solution \( m^k \) at iteration \( k \) and the exact solution \( m^* \) (defined in \eqref{eq:general:explcit}), during the CP iterations. The figure includes the pointwise error contour for the approximated \( m \). The recovered matrix \( \Lambda \) is:
\begin{align}
\label{eq:Lambda:polynomials}
\Lambda=\begin{bmatrix}
0.9379 & 0.0070 & 0.0088 & 0.0132 \\
0.0100 & 0.9391 & 0.0164 & 0.0138 \\
0.0175 & 0.0210 & 0.9422 & 0.0173 \\
0.0212 & 0.0113 & 0.0163 & 0.9415 \\
\end{bmatrix}.
\end{align}
As illustrated in Figure \ref{VfpolyQF}, the recovered \( F \) differs from the ground truth functions that generated the data, and the recovered \( \Lambda \) in \eqref{eq:Lambda:polynomials} deviates from the identity matrix. This indicates that recovering both \( \Lambda \) and the coupling function \( F \) is inherently challenging. In the absence of prior information about the metric and the coupling, multiple parameterizations can yield solutions consistent with the observed data. Nevertheless, these results highlight the flexibility of our approach in finding meaningful surrogate parameters that produce solutions aligned with the given data.

\subsubsection{Gaussian Process Approximation and Adjoint Method Optimization}
\label{MaternQ}
\begin{comment}
In this experiment, we follow the framework proposed in \ref{subsec:Finite_Min_Prob} to solve the inverse problems related to \eqref{Viscosity}. The optimization process dynamically adjusts the learning parameters \( V, \boldsymbol{z}_F, \) and \( \Lambda \) to efficiently minimize the loss function. Given the large number of parameters, we employ a step-by-step optimization strategy which is useful for all the following experiments. Initially, we reduce the learning rate for \( \boldsymbol{z}_F \) and focus on optimizing \( V \) and \( \Lambda \). Once the loss decreases to a certain level and the estimates for \( V \) and \( \Lambda \) become more accurate, we increase the learning rate for \( \boldsymbol{z}_F \) and continue optimizing \( \boldsymbol{z}_F, V, \) and \( \Lambda \) simultaneously. This approach improves the stability of the learning process for the proximal operator \( \operatorname{prox}_{\tau \varphi} \). 

To calculate the gradient of the loss function with respect to the parameters \( V, \boldsymbol{z}_F, \) and \( \Lambda \), we apply the adjoint method described in \ref{sec:adjoint_method}. Using this adjoint method enhances the accuracy of gradient calculations, independently of the lower-level optimization method, and allows us to achieve the same recovery accuracy with fewer sample points. However, this method requires more memory compared to the automatic differentiation method when computing the gradient in equation \eqref{eq:final_derivative}. 
\end{comment}

In this experiment, we follow the framework proposed in \ref{subsec:Finite_Min_Prob} to solve the inverse problems related to \eqref{Viscosity}.

\textbf{Experimental Setup.} The grid size is \( h = \tfrac{1}{40} \). Of the 1,600 total grid points for \(m\), 400 are selected as observation points, while 400 observation points for \(V\) are randomly distributed in space without grid constraints. The regularization parameters are \(\alpha_{m^o} = 6 \times 10^4\) and \(\alpha_{v^o} = \infty\). Additional coefficients include \(\alpha_v = 2\), \(\alpha_{fp} = 1\), \(\alpha_f = 4 \times 10^2\), and \(\alpha_{\lambda} = 3 \times 10^4\). Gaussian noise \(\mathcal{N}(0, \gamma^2 I)\) with \(\gamma = 10^{-3}\) is added to the observations.

The initial value of \(\Lambda\) is set to \(\Lambda = I + \boldsymbol{\xi}\), where \(\boldsymbol{\xi}\) is a matrix sampled from independent standard Gaussian variables, and the initial values for \(V\) are zero vectors. In the bilevel formulation proposed in Section \ref{sec:time_dependent}, we approximate the coupling function \(f\) by modeling its primal function \(F\) (as defined in \eqref{eq:coupling_terminal}) through a GP  conditioned on observations at 10 fixed points \(\{r_i\}_{i=1}^{10}\) evenly spaced between 0.4 and 2.2. To approximate the expectation in the penalization term $-E_{\xi\sim\mu}E_{\zeta\sim\mu}\bigl[\langle F'(\xi) - F'(\zeta), \xi - \zeta\rangle\bigr],$ which enforces convexity, we sample 200 points \(\{\widetilde{r}_i\}_{i=1}^{200}\) evenly spaced between 0.4 and 1.5 and use the double sum $-\sum_{i=1}^{200} \sum_{j=1}^{200} \langle F'(\widetilde{r}_i) - F'(\widetilde{r}_j), \widetilde{r}_i - \widetilde{r}_j\rangle$ to approximate the expectation.

In our numerical experiments, we observed that the true coupling function \(F(m) = \frac{1}{4}m^4\) is non-negative for non-negative \(m\). To maintain this property, we parameterize the coefficients of \(F\), denoted \(\boldsymbol{z}_F\) in \eqref{eq:repre:VF}, as \(\boldsymbol{z}_F = \widetilde{\boldsymbol{z}}_F^2\) and optimize over \(\widetilde{\boldsymbol{z}}_F\). The initial values of \(\widetilde{\boldsymbol{z}}_F\) are chosen as equally spaced points in a sufficiently large interval to ensure that their squares cover the range of \(m\). We optimize the objective using the adjoint method detailed in Section \ref{sec:adjoint_method}.

\begin{figure}[h]
    \centering
    \begin{subfigure}[b]{0.33\textwidth}
        \centering
        \includegraphics[width=\linewidth]{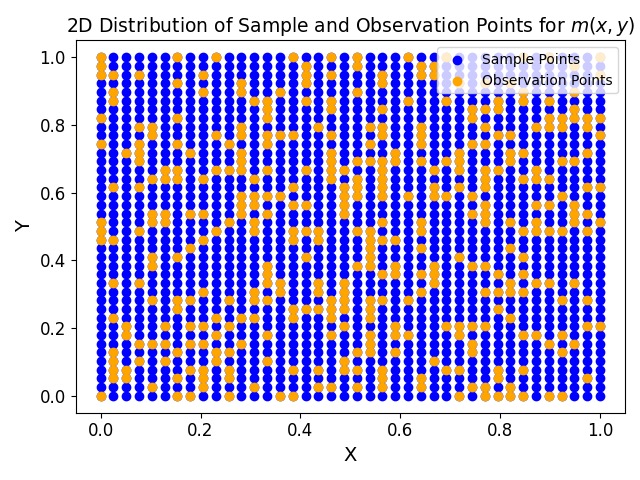}
        \caption{Samples \& Observations of $m$}
        \label{fig:msample1}
    \end{subfigure}%
    \begin{subfigure}[b]{0.33\textwidth}
        \centering
        \includegraphics[width=\linewidth]{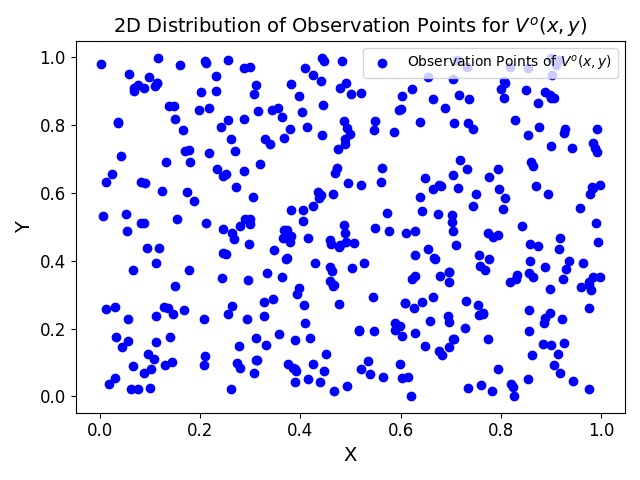}
        \caption{Observations of $V$}
        \label{fig:v01}
    \end{subfigure}%
    \begin{subfigure}[b]{0.33\textwidth}
        \centering
        \includegraphics[width=\linewidth]{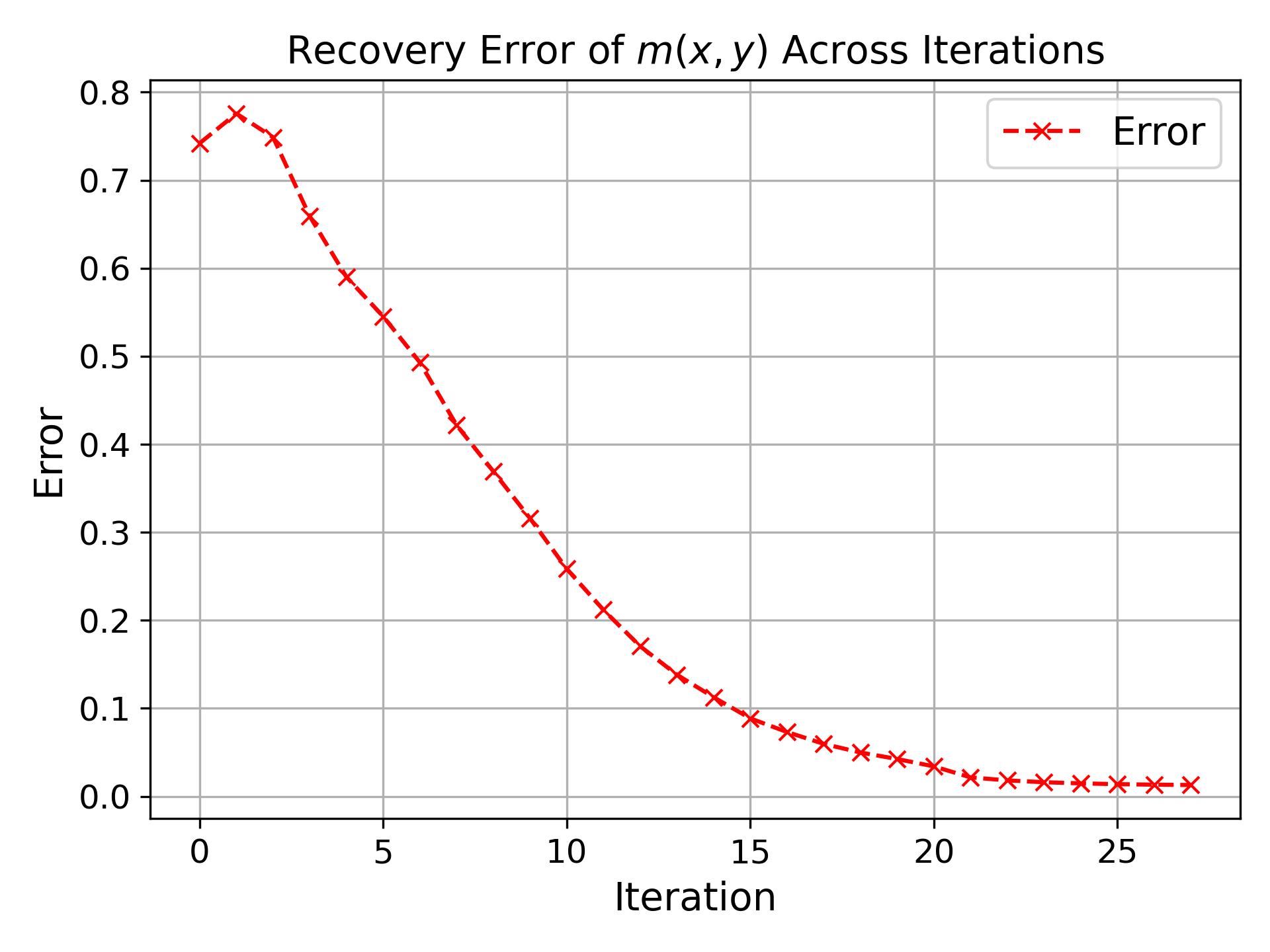}
        \caption{Error \( \mathcal{E}(m^k, m^*) \) vs.  Iteration  \( k \)}
        \label{fig:error2}
    \end{subfigure}
\begin{subfigure}[b]{0.33\textwidth}
        \includegraphics[width=\linewidth]{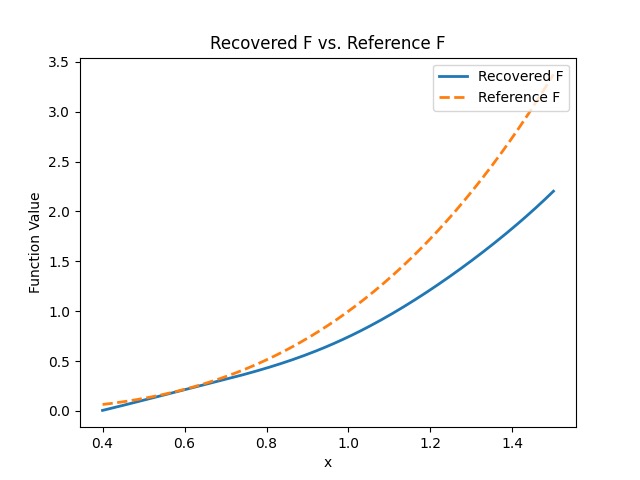}
    \caption{Recovered F vs. Reference F}
    \label{VfmaternQF}
    \end{subfigure}
   \begin{subfigure}[b]{0.33\textwidth}
        \centering
        \includegraphics[width=\linewidth]{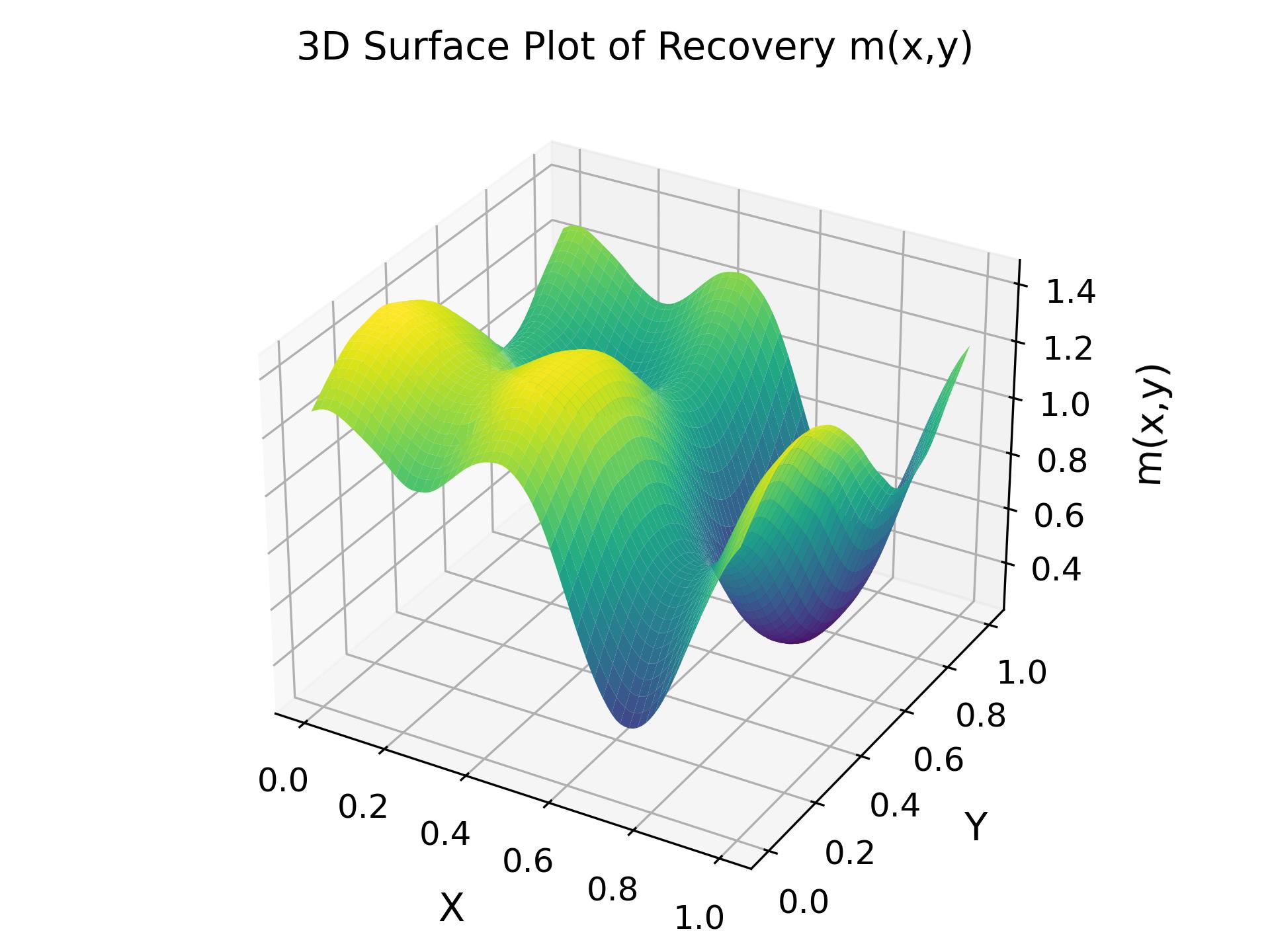}
        \caption{Recovered $m$}
        \label{fig:recover1}
    \end{subfigure}%
    \begin{subfigure}[b]{0.33\textwidth}
        \centering
        \includegraphics[width=\linewidth]{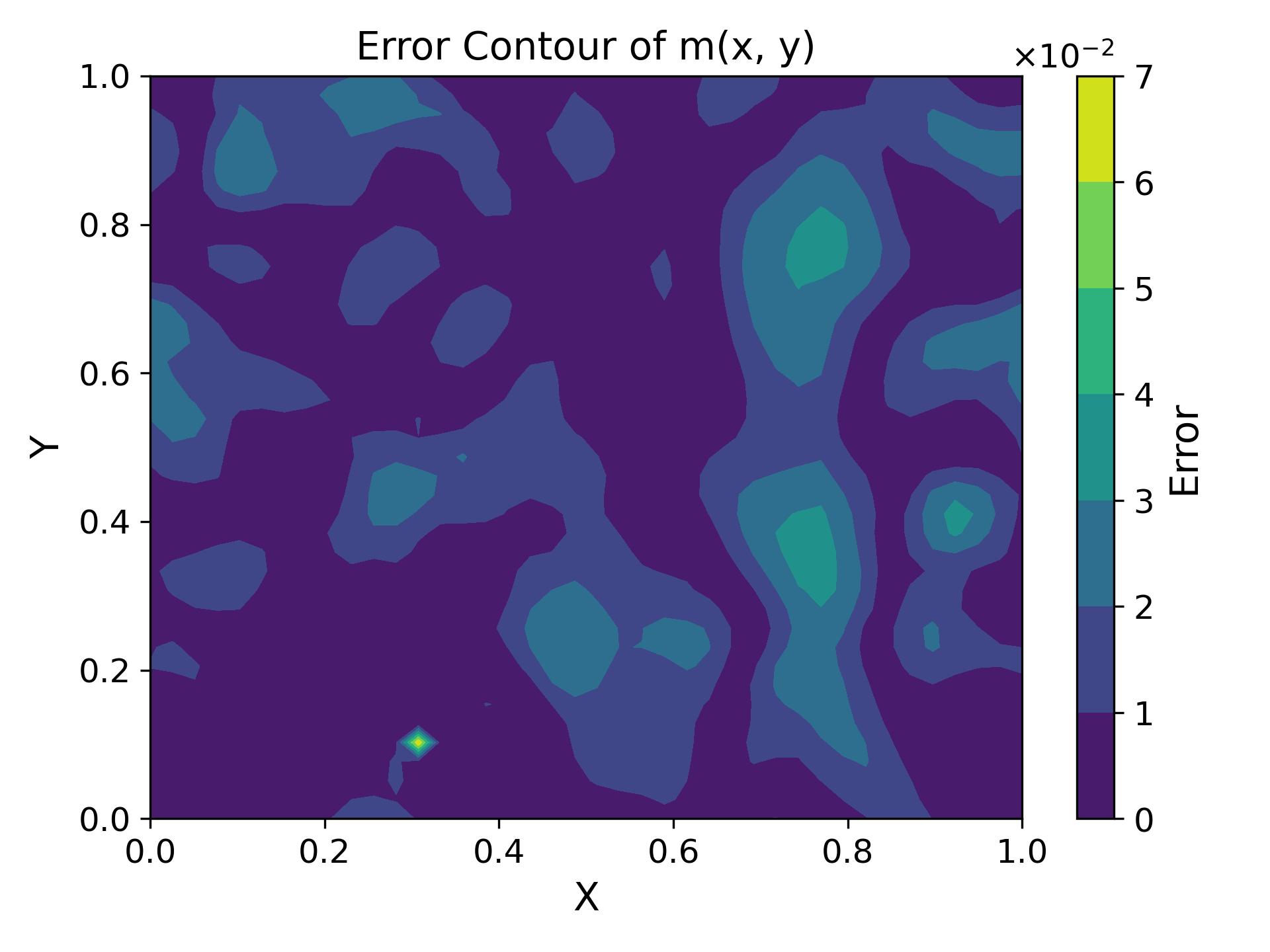}
        \caption{Error Contour of $m$}
        \label{fig:errorcontour1}
    \end{subfigure}
    \begin{subfigure}[b]{0.33\textwidth}
        \centering
        \includegraphics[width=\linewidth]{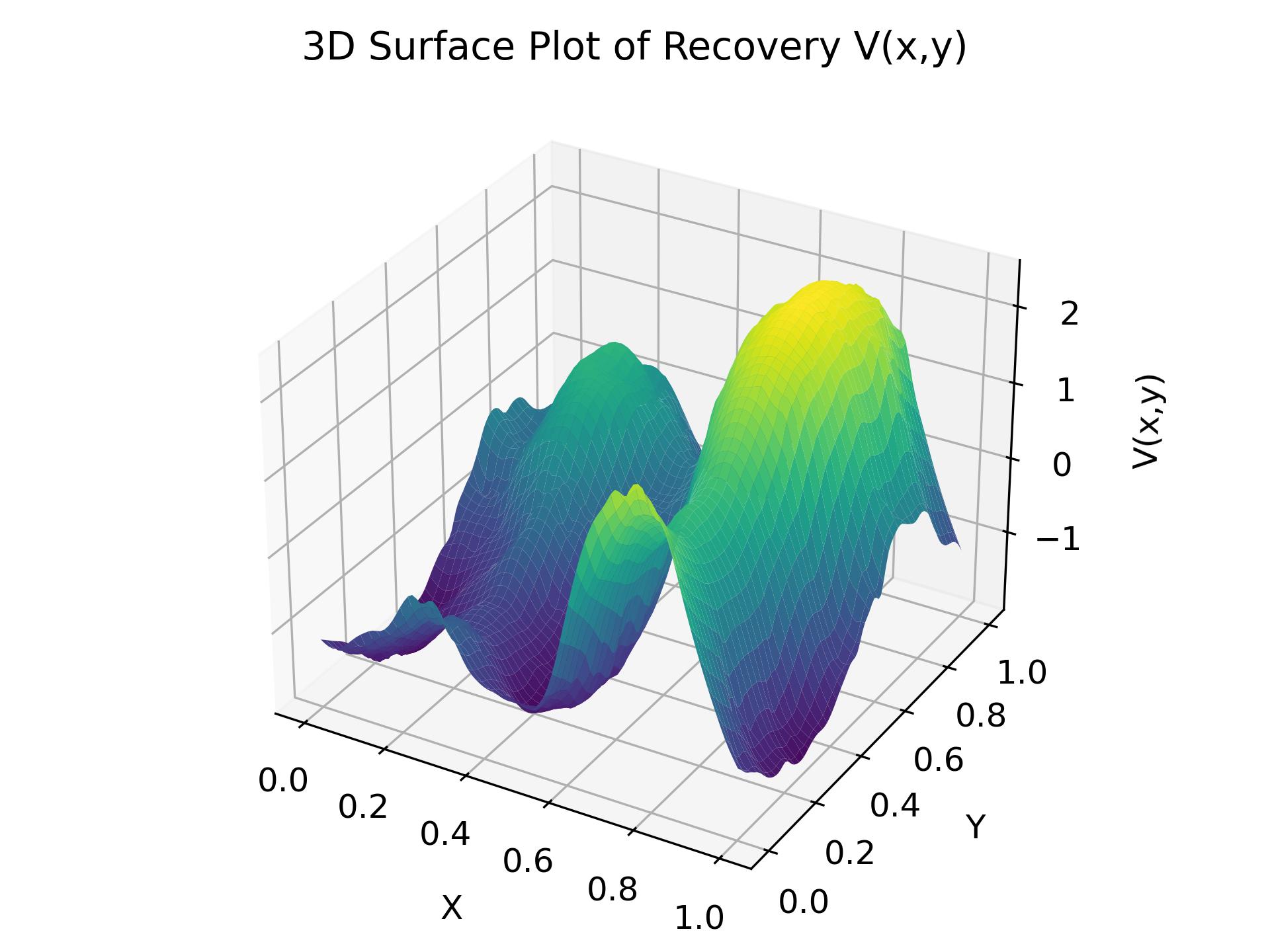}
        \caption{Recovered $V$}
        \label{fig:v01}
    \end{subfigure}%
    \begin{subfigure}[b]{0.33\textwidth}
        \centering
        \includegraphics[width=\linewidth]{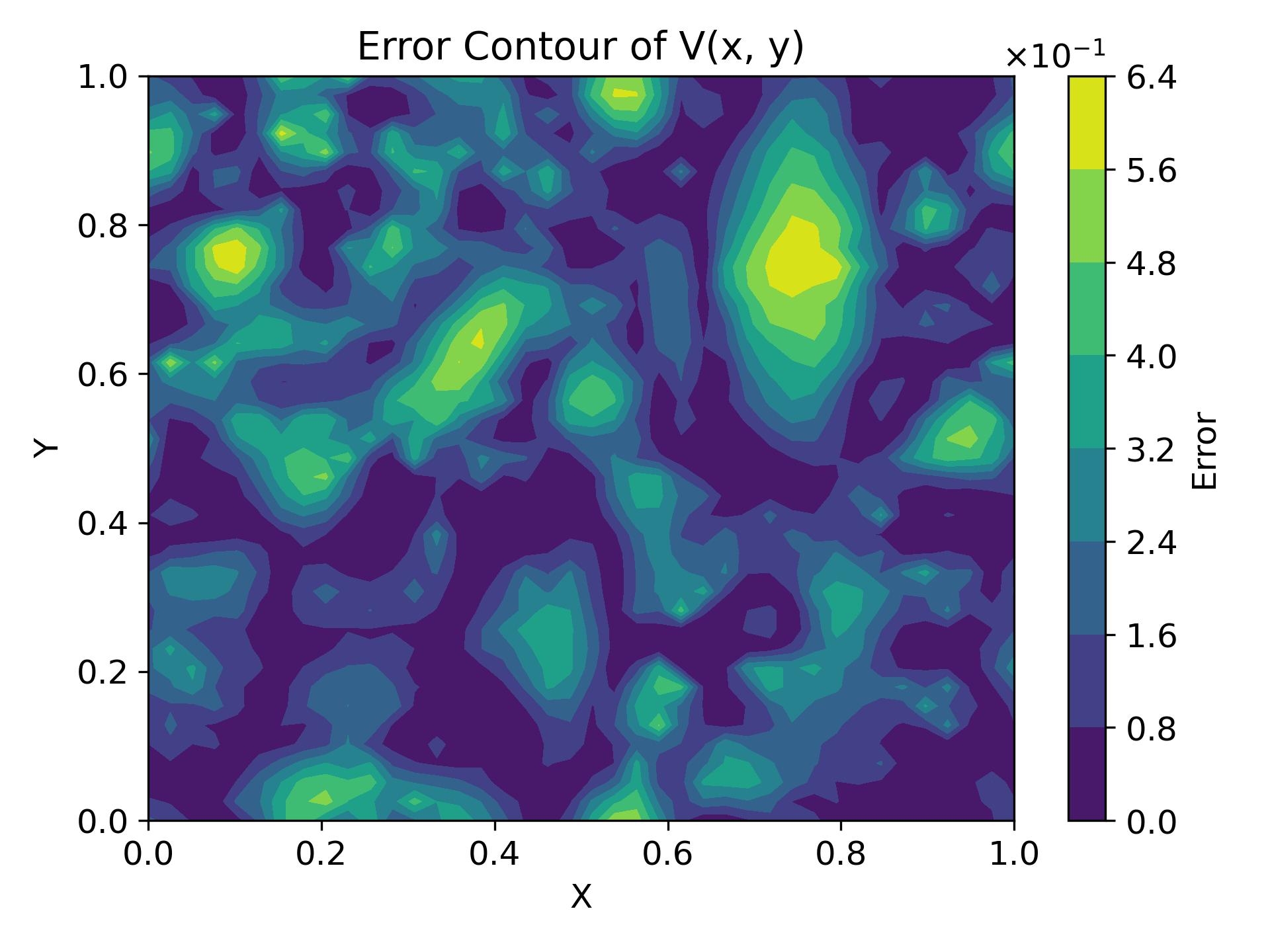}
        \caption{Error Contour of $V$}
        \label{fig:error2}
    \end{subfigure}
    
    \caption{ Numerical results for approximating the coupling function \( F \) using a GP and solving the inverse problem of the MFG system in \eqref{Viscosity} via adjoint method optimization:  
(a) sample (grid) points and observation points of \( m \);  
(b) observation points of \( V \);  
(c) discretized \( L^2 \) error \( \mathcal{E}(m^k, m^*) \) versus iteration number \( k \); 
(d) recovered \( F \) vs. reference \( F \); 
(e) recovered \( m \);  
(f) pointwise error between the recovered \( m \) and the exact \( m^* \);   
(g) recovered \( V \);  
(h) pointwise error between the recovered \( V \) and its exact solution. While the recovered \(F\) differs from the ground truth, it retains convexity, and the resulting MFG solutions remain consistent with the observed data.
}
    \label{figVfQmatern}
\end{figure}

\textbf{Experiment Results.} 
Figure \ref{figVfQmatern} shows the collocation grid points for \( m \) and the observation points for both \( m \) and \( V \). It also displays the discretized \( L^2 \) error \(\mathcal{E}(m^k, m^*)\) from \eqref{eq:l2disc}, quantifying the discrepancy between the approximate solution \( m^k \) and the exact solution \( m^* \) during the CP iterations. Additionally, the figure provides pointwise error contours for the approximated \( m \) and \( V \). The recovered matrix \(\Lambda\) is
\[
\Lambda = \begin{bmatrix}
    0.9713 & 0.0158 & 0.2712 & 0.0098 \\
    0.0503 & 0.9361 & 0.0084 & 0.1011 \\
    0.2980 & -0.0094 & 0.9437 & 0.0190 \\
    0.0449 & 0.1225 & 0.0421 & 0.9035 \\
\end{bmatrix}
.
\]
As illustrated in Figure \ref{VfmaternQF}, the method recovers an alternative coupling \( F \) that fits the data well. This example also demonstrates that the convexity penalization term in \eqref{biorp} effectively enforces the convexity of \( F \) and ensures the recovered coupling meets the Lasry--Lions monotonicity condition. 

\subsection{Recovering Non-Euclidean Metrics, Hamiltonian Exponents, Viscosity Constants, and Spatial Costs Together} 
\label{Maternq}
\label{subsec:num:NLambdaVF}
In this subsection, we address the inverse problem for the MFG in \eqref{eq:gsmfg} with \(q = 2.2\), and \(\Lambda = I + 0.2A\), where \(A\) is the all-ones matrix. The potential function is
$
V(x, y) = -\bigl(\sin(2\pi x) + \cos(4\pi x) + \sin(2\pi y)\bigr), (x, y)\in \mathbb{T}^2$, 
and the coupling function is \(F(m) = \frac{1}{4}m^4\). The viscosity coefficient is \(\nu = 0.1\). We identify \(\mathbb{T}^2\) with the domain \([0,1) \times [0,1)\).

\begin{figure}[h]
    \centering
    \begin{subfigure}[b]{0.33\textwidth}
        \centering
        \includegraphics[width=\linewidth]{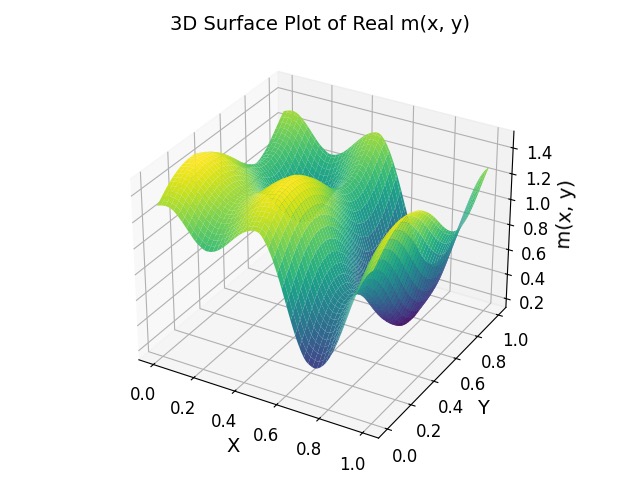}
        \caption{Reference Solution $m^*$}
        \label{figrealvfQq}
    \end{subfigure}%
    \begin{subfigure}[b]{0.33\textwidth}
        \centering
        \includegraphics[width=\linewidth]{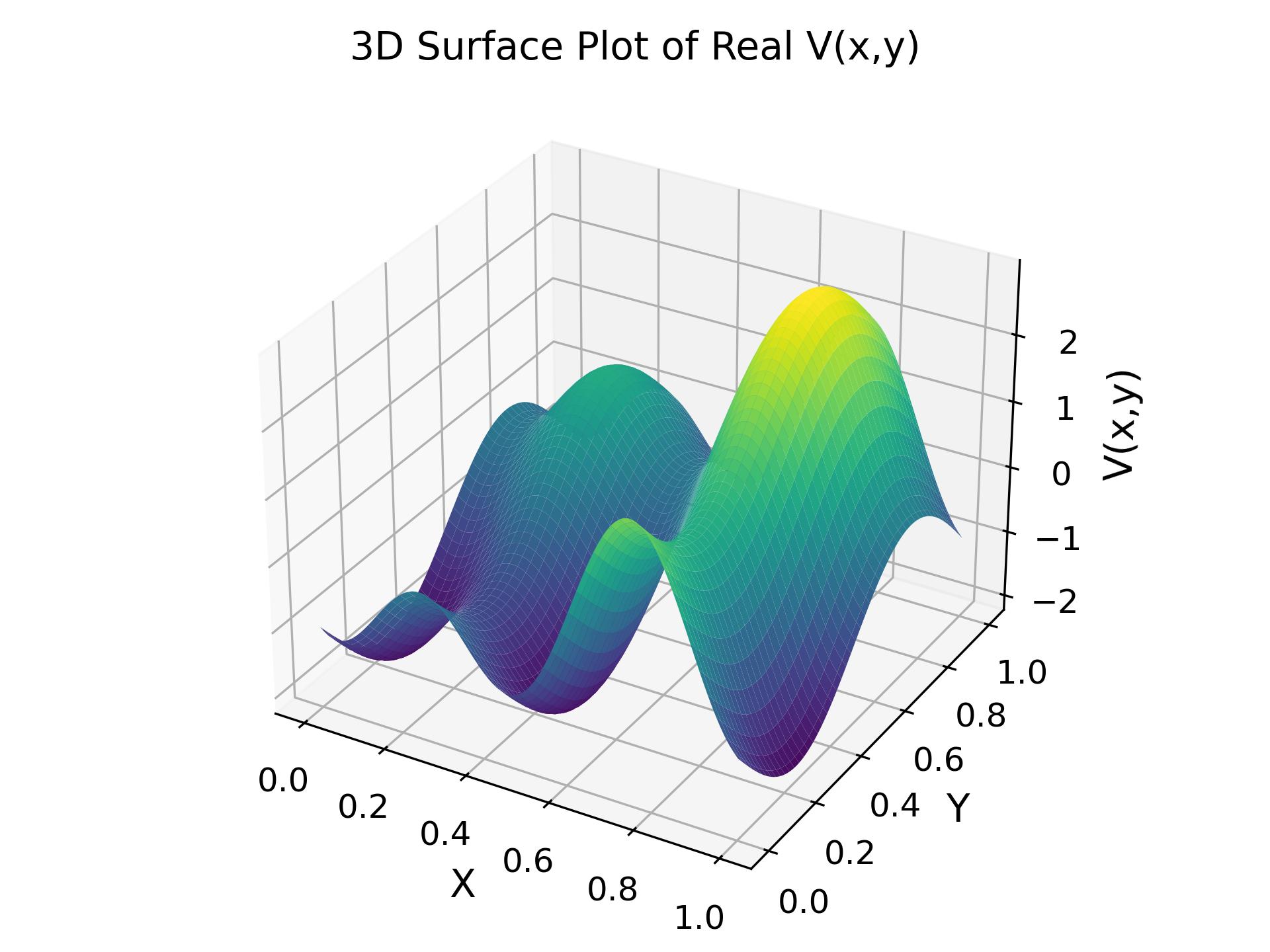}
        \caption{Ground Truth $V$}
        \label{fig:msample1}
    \end{subfigure}%   \end{subfigure}
\caption{Reference results of $m$ and $V$ for the MFG in Subsection~\ref{Maternq}.
}
\label{ReferenceViscosity}
\end{figure}

Our objective is to recover the MFG distribution \(m\), along with the unknown functions \(F\) and \(V\), the non-Euclidean metric \(\Lambda\), the Hamiltonian exponent \(q\), and the viscosity coefficient \(\nu\), using partial noisy observations of \(m\) and \(V\). We adopt the framework in Section~\ref{subsec:Finite_Min_Prob}.
The reference solution is obtained via the CP algorithm from \cite{briceno2018proximal} as shown in Figure \ref{figrealvfQq}. 

\begin{comment}
 The optimization process adjusts the learning parameters \( V, \boldsymbol{z}_F, \Lambda, q, \) and \( \nu \) dynamically to efficiently minimize the loss function. Initially, we reduce the learning rate for \( \boldsymbol{z}_F, \Lambda, q, \) and \( \nu \) while focusing on learning the parameter \( V \). Once the loss decreases to a certain level and the estimate of \( V \) becomes more accurate, we increase the learning rates of the other parameters and continue optimizing all parameters simultaneously.
\end{comment}

%\textbf{Learning Process}
%We follow the inverse problem framework outlined in \ref{Hamiltonian q}. The optimization process adjusts the learning parameters \( \mathbf{v}, f_z, \Lambda, q, \) and \( \nu \) dynamically to efficiently minimize the loss function. Initially, we reduce the learning rate for \( f_z, \Lambda, q, \) and \( \nu \) while focusing on learning the parameter \( V \). Once the loss decreases to a certain level and the estimate of \( V \) becomes more accurate, we increase the learning rates of the other parameters and continue optimizing all parameters simultaneously.

\textbf{Experimental Setup.} The grid size is \( h = \tfrac{1}{50} \). Among 2{,}500 grid points for \(m\), 625 are selected as observation points for \(m\), while another 625 points for \(V\) are randomly distributed without grid restrictions. We set the regularization parameters \(\alpha_{m^o} = 6 \times 10^4\), \(\alpha_{v^o} = \infty\), \(\alpha_v = 2\), \(\alpha_{fp} = 1\), \(\alpha_f = 4 \times 10^2\), \(\alpha_{\lambda} = 7 \times 10^4\), \(\alpha_q = 2 \times 10^4\), and \(\alpha_{\nu} = 3 \times 10^3\). Gaussian noise \(\mathcal{N}(0,\gamma^2 I)\) with \(\gamma = 10^{-3}\) is added to the observations.
The initial value of \(\Lambda\) is \(\Lambda = I + 0.4A\), where \(A\) is the all-ones matrix, and the initial values of \(V\) are zero vectors. The initial values for \(q\) and \(\nu\) are 2.4 and 0.2, respectively.

As in the previous examples, we employ the bilevel formulation from Section~\ref{sec:time_dependent} to recover the coupling function \(f\). Specifically, we approximate its primal form \(F\) (see \eqref{eq:coupling_terminal}) with a GP and penalize non-convexity using Monte Carlo samples. We also enforce the non-negativity of the true coupling function \(F(m) = \frac{1}{4}m^4\) by parameterizing its coefficients as \(\boldsymbol{z}_F = \widetilde{\boldsymbol{z}}_F^2\). The initial values for these parameters, as well as details on  Monte Carlo approximation and adjoint-based optimization, follow the same procedure described in Section~\ref{MaternQ}.

% When solving the proximal operator in CP Algorithm when $q\neq 2$, we need to calculate the root of some equations, where we may use the variable substitution method, Newton's method with dynamic learning rate and line search method for root finding.

\begin{figure}[h]
    \centering
    \begin{subfigure}[b]{0.33\textwidth}
        \centering
        \includegraphics[width=\linewidth]{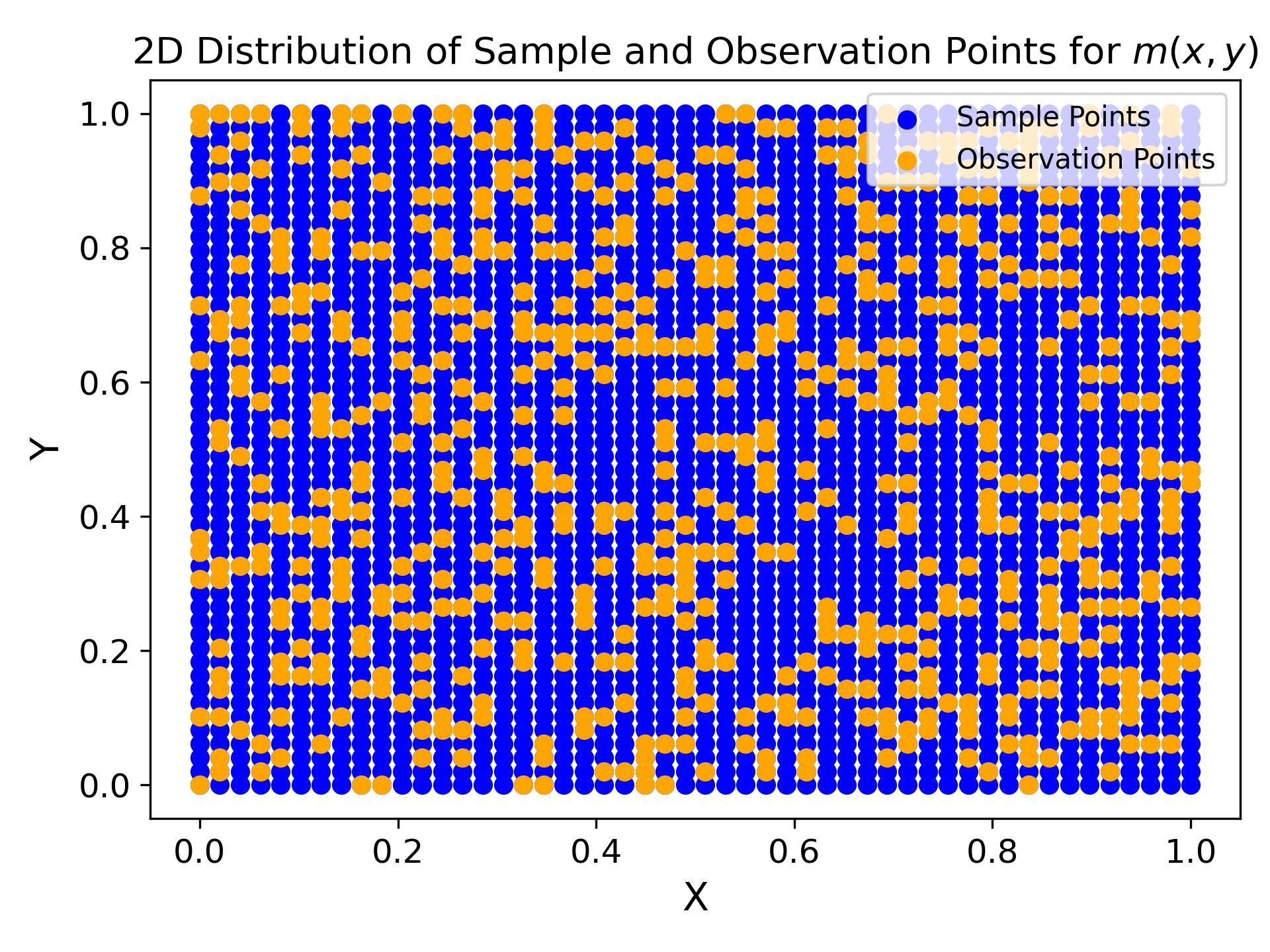}
        \caption{Samples \& Observations of $m$}
        \label{fig:msample1}
    \end{subfigure}%
    \begin{subfigure}[b]{0.33\textwidth}
        \centering
        \includegraphics[width=\linewidth]{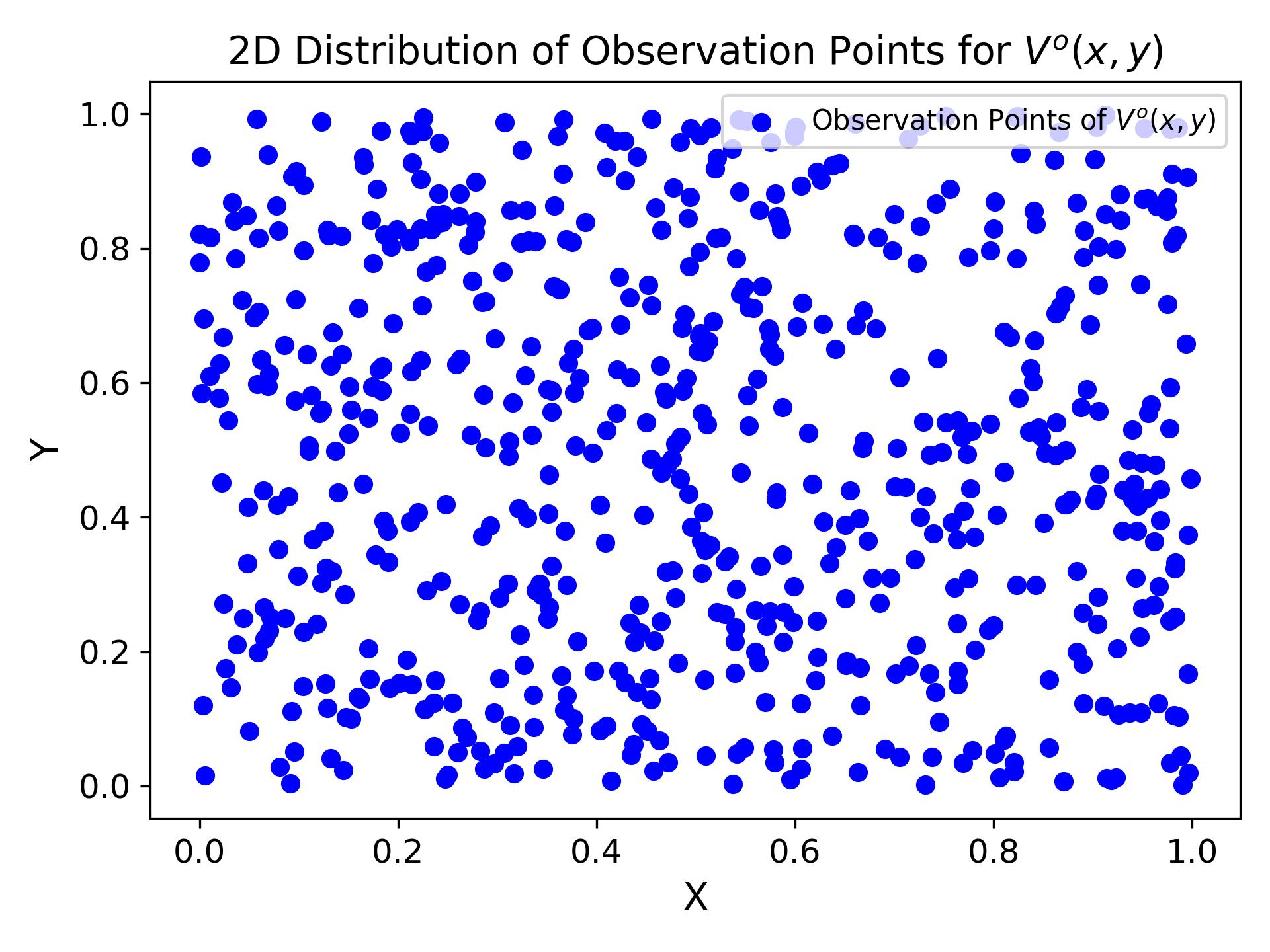}
        \caption{Observations of $V$}
        \label{fig:v01}
    \end{subfigure}%
    \begin{subfigure}[b]{0.33\textwidth}
        \centering
        \includegraphics[width=\linewidth]{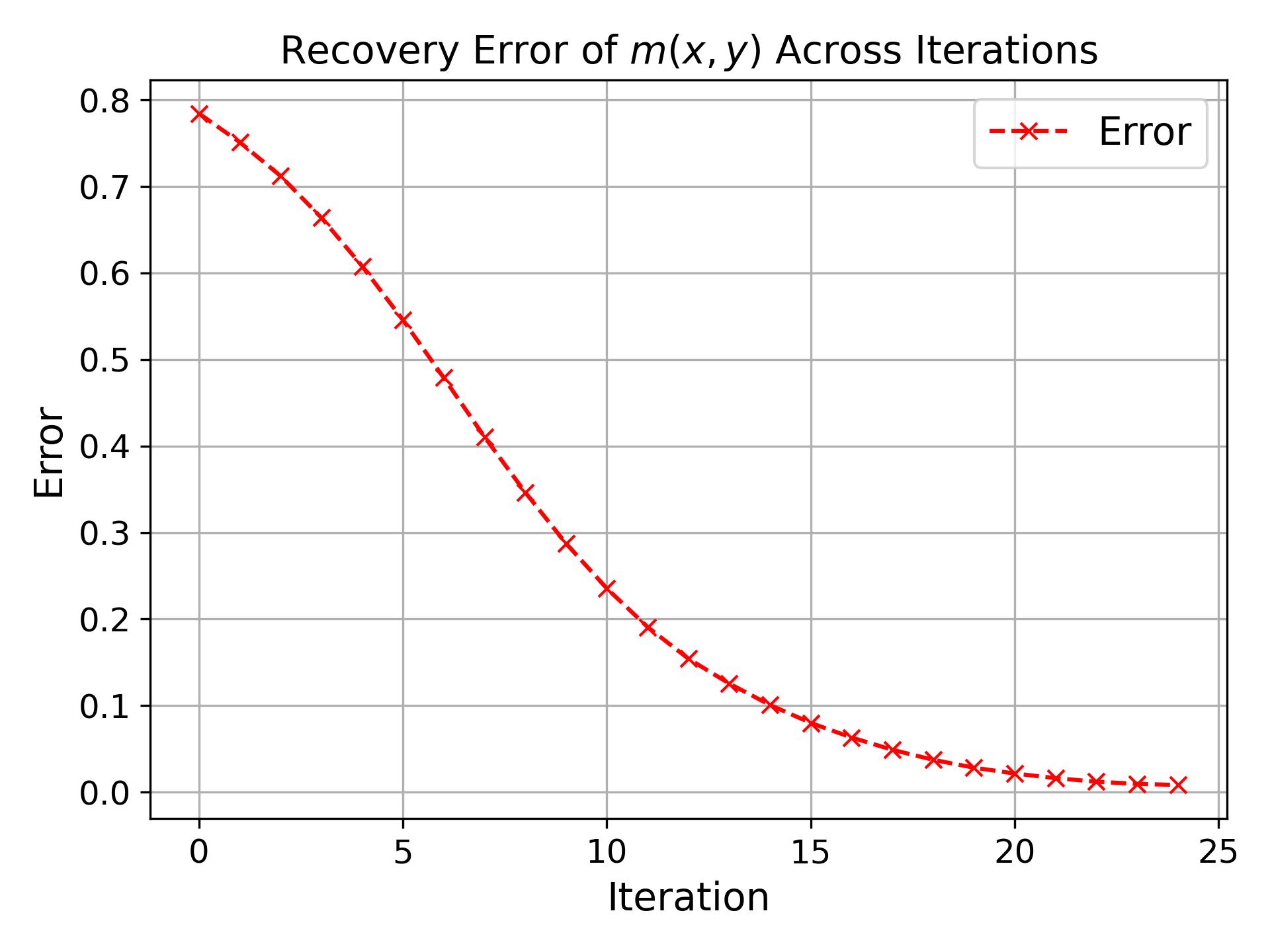}
        \caption{Error \( \mathcal{E}(m^k, m^*) \) vs.  Iteration  \( k \)}
        \label{fig:error2}
    \end{subfigure}
    \begin{subfigure}[b]{0.33\textwidth}
        \centering
        \includegraphics[width=\linewidth]{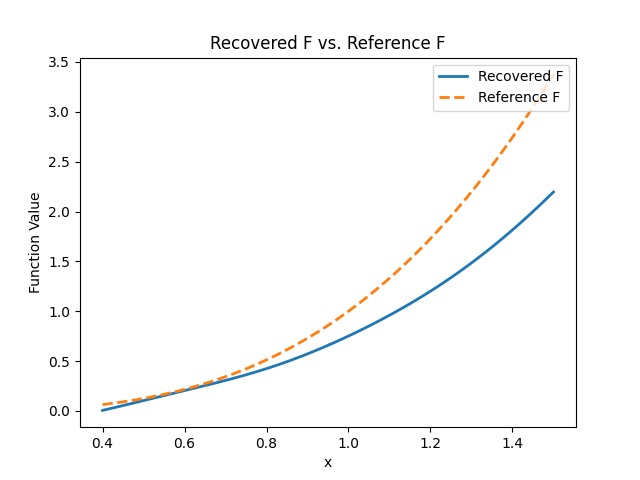}
    \caption{Recovered F vs. Reference F.}
    \label{VfmaternQFq}
    \end{subfigure}
    \begin{subfigure}[b]{0.33\textwidth}
        \centering
        \includegraphics[width=\linewidth]{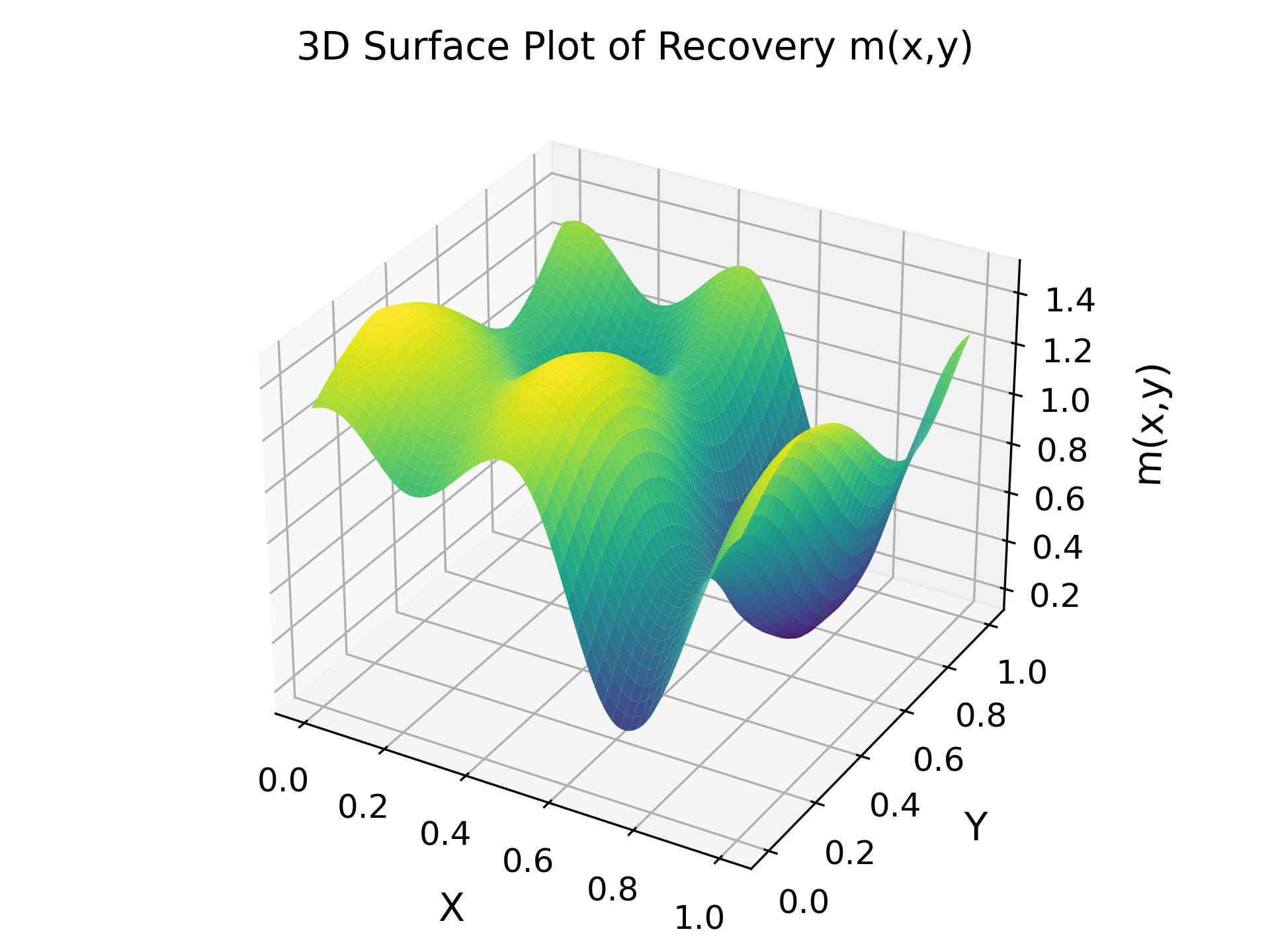}
        \caption{Recovered $m$}
        \label{fig:recover1}
    \end{subfigure}%
  \begin{subfigure}[b]{0.33\textwidth}
        \centering
        \includegraphics[width=\linewidth]{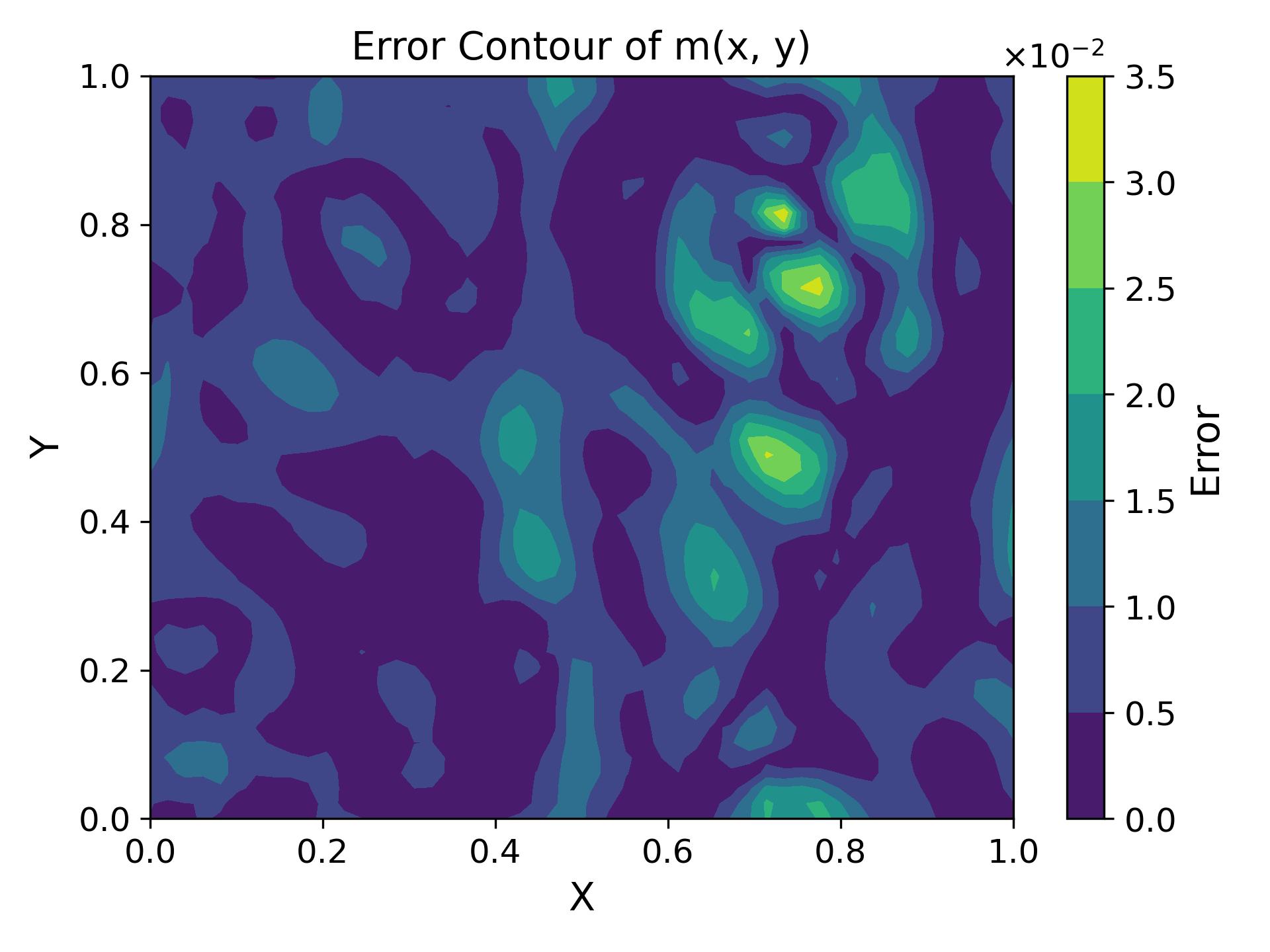}
        \caption{Error Contour of $m$}
        \label{fig:errorcontour1}
    \end{subfigure}
    \begin{subfigure}[b]{0.33\textwidth}
        \centering
        \includegraphics[width=\linewidth]{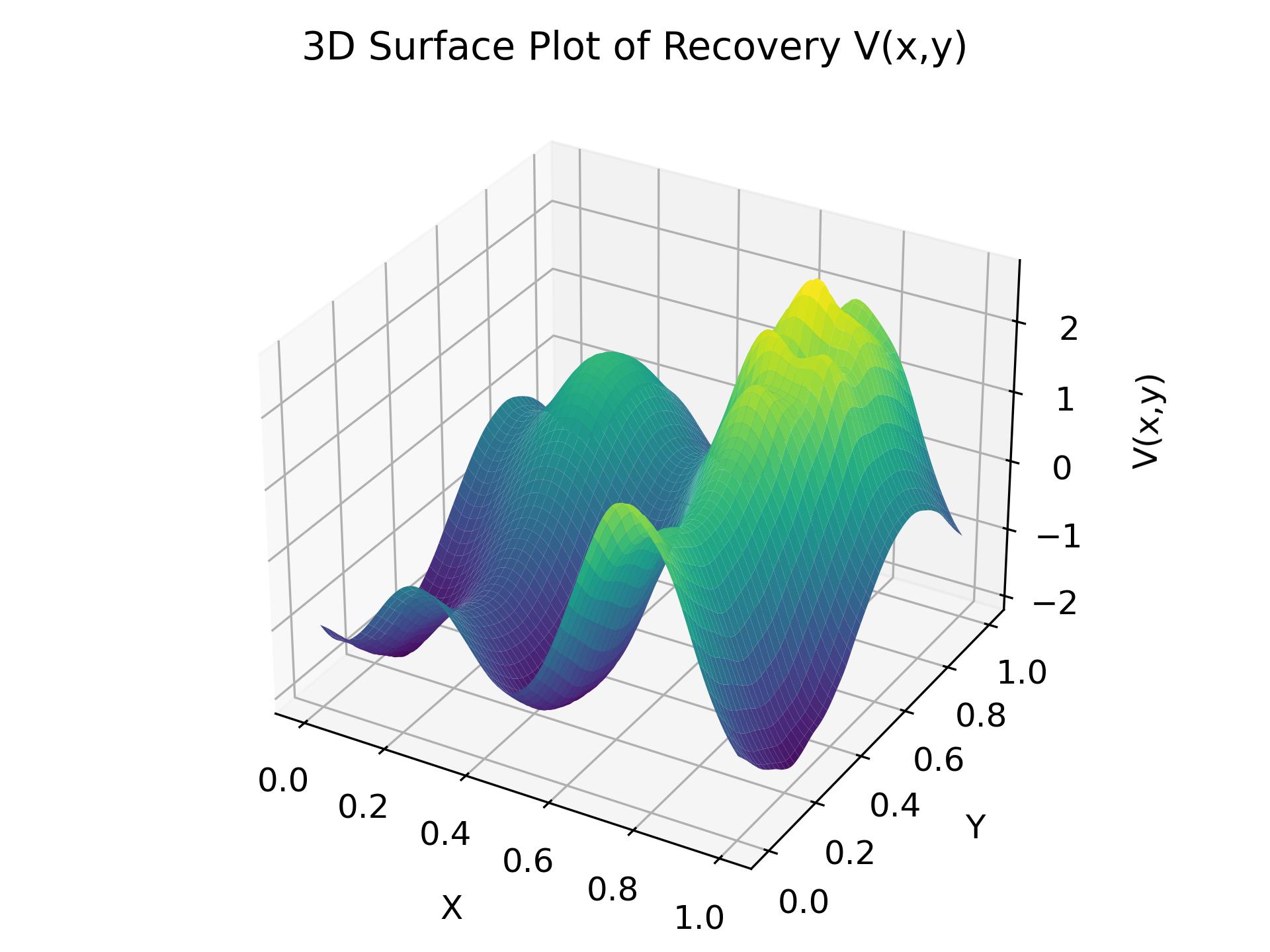}
        \caption{Recovered $V$}
        \label{fig:v01}
    \end{subfigure}%
    \begin{subfigure}[b]{0.33\textwidth}
        \centering
        \includegraphics[width=\linewidth]{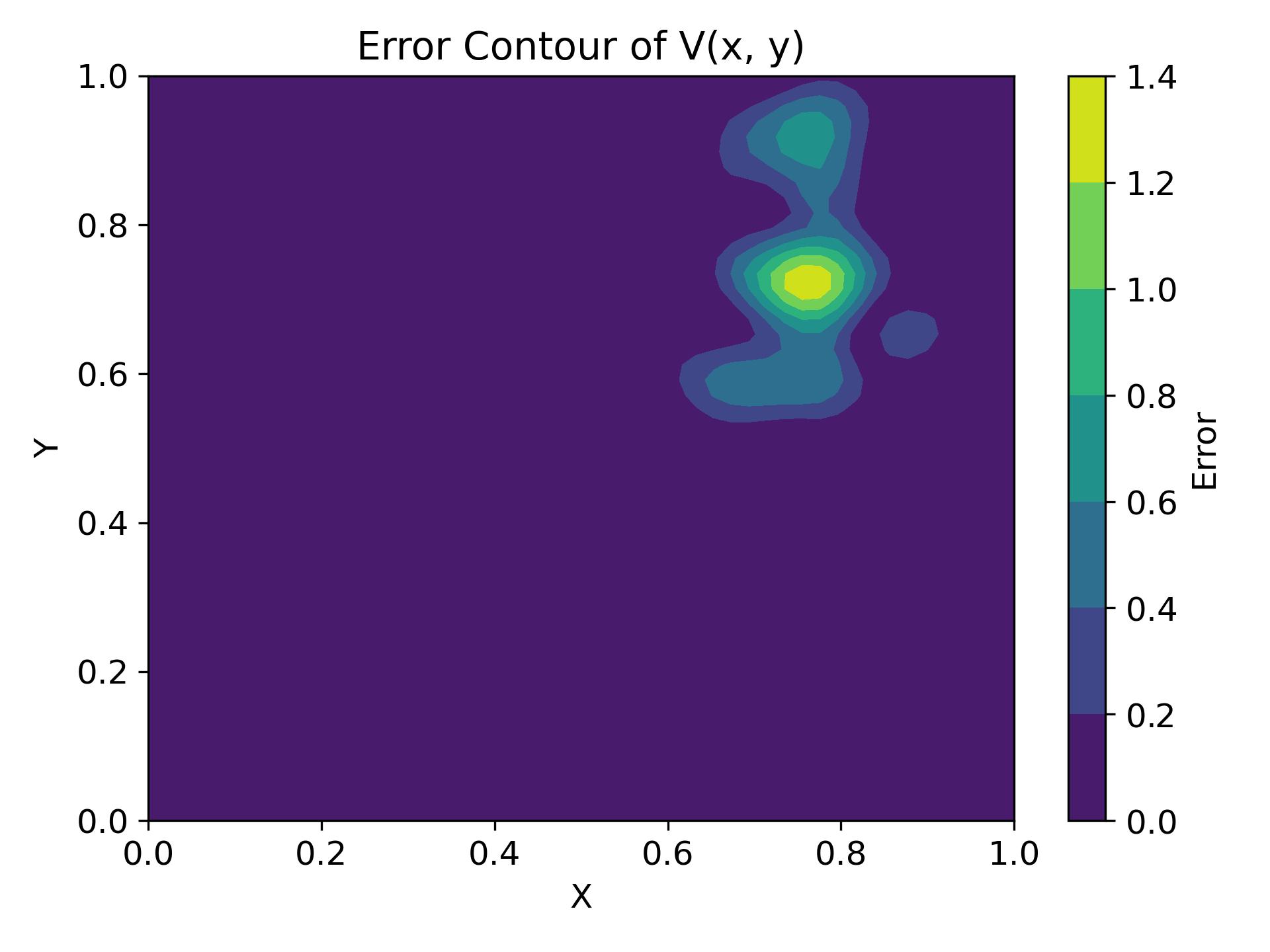}
        \caption{Error Contour of $V$}
        \label{fig:error2}
    \end{subfigure}
    
    \caption{Numerical results for the inverse problem of the MFG system in Subsection \ref{Maternq}: (a) sample (grid) points and observation points of \( m \);  
(b) observation points of \( V \);  
(c) discretized \( L^2 \) error \( \mathcal{E}(m^k, m^*) \) versus iteration number \( k \); 
(d) recovered \( F \) vs. reference \( F \); 
(e) recovered \( m \);  
(f) pointwise error between the recovered \( m \) and the exact \( m^* \);   
(g) recovered \( V \);  
(h) pointwise error between the recovered \( V \) and its exact solution. Although the recovered \(F\) does not match the ground truth, it maintains convexity, and the corresponding MFG solutions closely align with the observed data.}
    \label{figfVQq}
\end{figure}

\textbf{Experiment Results.}
Figure \ref{figfVQq} presents the numerical results. The recovered values of \(q\) and \(\nu\) are \(q = 2.1063\) and \(\nu = 0.0737\), respectively. The recovered \(\Lambda\) is
\[
\Lambda = \begin{bmatrix}
1.0435 & 0.1147 & 0.4045 & 0.0951 \\
0.1226 & 1.0316 & 0.0694 & 0.4058 \\
0.3743 & 0.0321 & 0.7086 & 0.0943 \\
0.0161 & 0.3606 & 0.0978 & 0.7118
\end{bmatrix}.
\]

Notably, the recovered \(F\) remains convex, and the recovered density data closely matches the observations, indicating that although the recovered MFG differs from the one that generated the data, it still reproduces the observed data effectively.

\begin{comment}
z Parameter containing:

         $$
z=\left[1.86 \times 10^{-4}, 0.466,0.640,0.872,1.09,1.35,1.62,1.88,2.06,2.15\right]
$$
\end{comment}

\begin{comment}
\begin{figure}[H]   
    \centering
    \includegraphics[scale=0.3]{FigureVfQq/afig10VfQq.jpg}
    \caption{Recovered F vs. Reference F.}
    \label{VfmaternQFq}
\end{figure}
\begin{figure}[H]   
    \centering
    \includegraphics[scale=0.3]{FigureVfQq/afig11VfQq.jpg}
    \caption{Recovered f vs. Reference f.}
    \label{VfmaternQfq}
\end{figure}
\end{comment}

\subsection{An Inverse Problem for A Time-Dependent MFG}
\label{subsec:num:td}
In this example, we study the inverse problem for the time-dependent MFG \eqref{eq:time-dependent} on the one-dimensional torus \(\mathbb{T}\). We set \(T = 1\) and define 
\[
V(x, t) = -\tfrac{1}{2}\bigl(\sin(2\pi x) + 2\cos(2\pi x)\bigr), 
\quad
F(t, m) = \tfrac{1}{4}m^4,
\quad
m_0=1, 
\quad
\mathcal{G} = 0,
\quad
q = 2.4,
\]
and \(\Lambda = I + 0.48\,A\), where \(A\) is the all-ones matrix. The domain \(\mathbb{T}\) is identified with \(\bigl[-0.5, 0.5\bigr)\), and the viscosity coefficient is \(\nu = 0.1\).

Given \(\nu\), \(\Lambda\), \(f\), \(V\), and \(q\), we solve  \eqref{eq:time-dependent} with these settings using the primal-dual method from \cite{briceno2019implementation} to obtain reference solutions \((w^*, m^*)\). Our inverse problem then seeks to recover \(\nu\), \(\Lambda\), \(f\), \(V\), \(q\), and \(m\) from noisy, partial observations of \(m\) and \(V\), applying the bilevel framework described in Section~\ref{sec:time_dependent}. We compare the errors arising from this procedure with the reference solutions.

%\textbf{Learning Process}
%The optimization process dynamically adjusts the learning parameters \( \mathbf{v}, f_z, \Lambda, q, \) and \( \nu \) to minimize the loss function efficiently. Initially, the learning rates for \( f_z \) and \( \Lambda \) are reduced to focus on learning the parameters \( V, q, \) and \( \nu \). Once the loss decreases and these parameters become more accurate, the learning rates for \( f_z \) and \( \Lambda \) are increased, and all parameters are optimized together. This process enhances the stability of the learning for the proximal operator \( \operatorname{prox}_{\tau \varphi} \). 

\begin{figure}[h]
    \centering
    \begin{subfigure}[b]{0.33\textwidth}
        \centering
        \includegraphics[width=\linewidth]{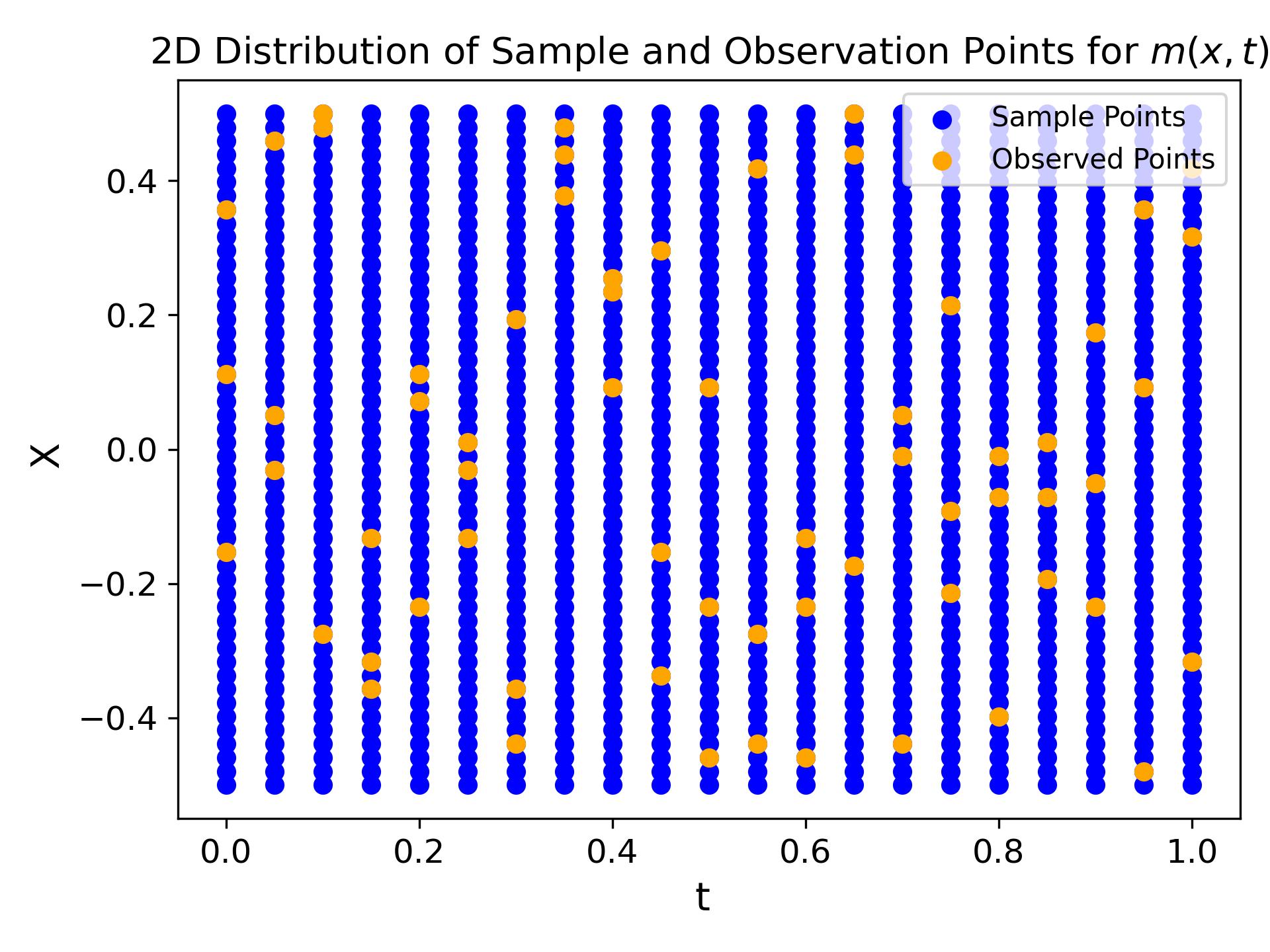}
        \subcaption{Samples \& Observations of $m$}
        \label{fig:msample1}
    \end{subfigure}%
    \begin{subfigure}[b]{0.33\textwidth}
        \centering
        \includegraphics[width=\linewidth]{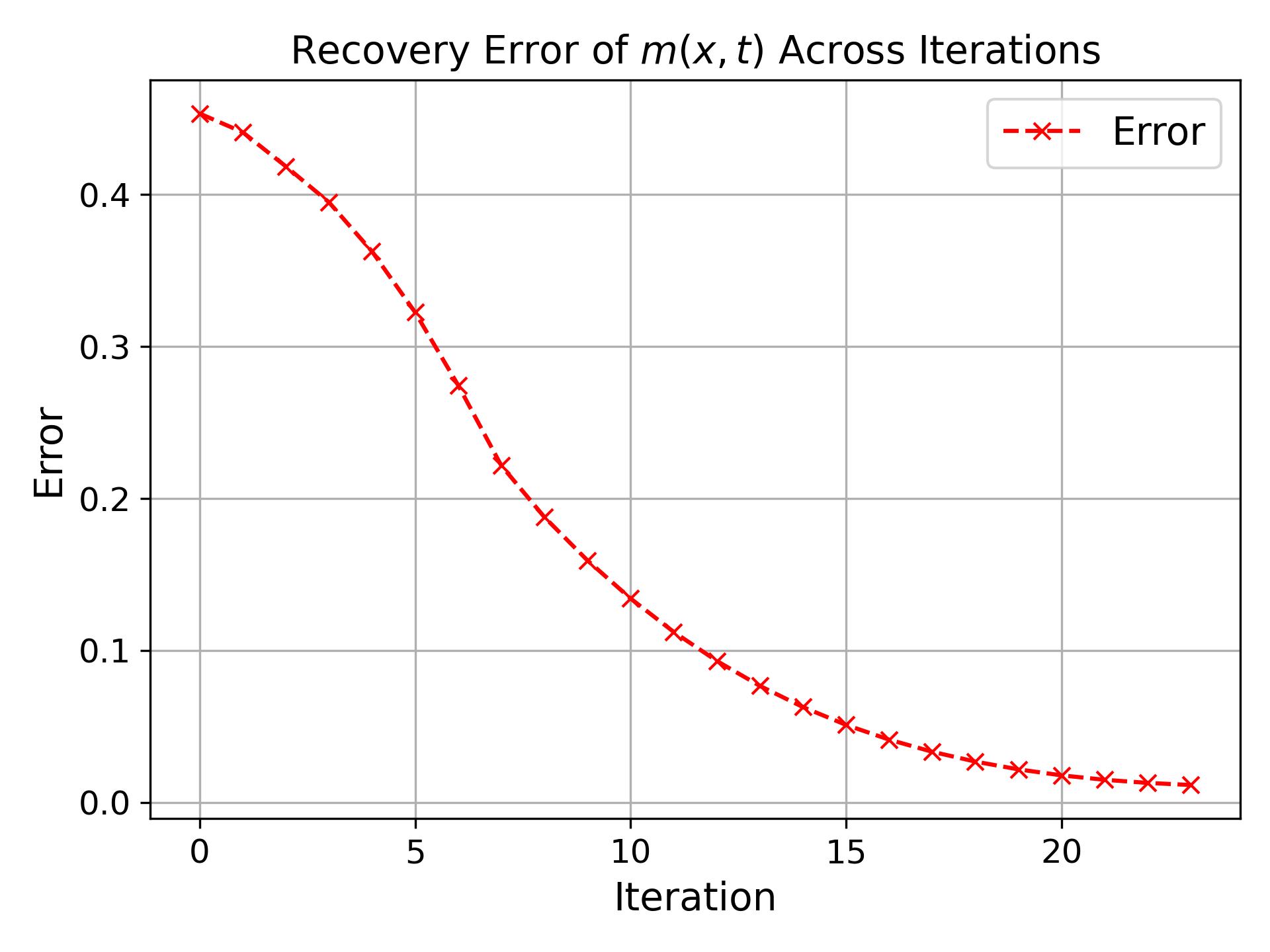}
        \subcaption{Error $\mathcal{E}(m^k, m^*)$ vs. Iteration $k$}
        \label{fig:error2}
    \end{subfigure}
    \begin{subfigure}[b]{0.33\textwidth} 
    \centering
    \includegraphics[width=\linewidth]{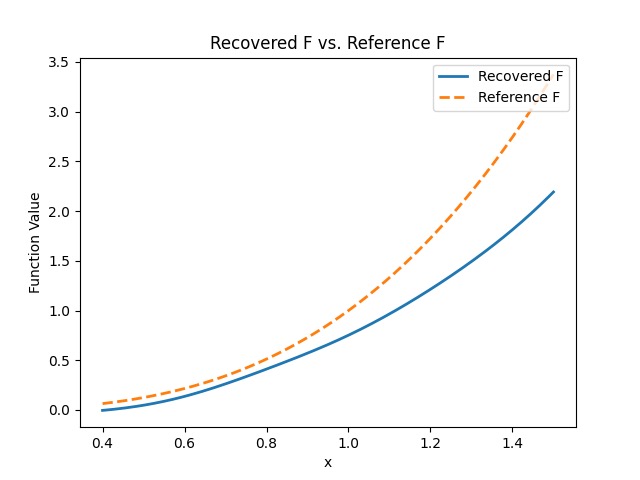}
    \caption{Recovered F vs. Reference F}
    \label{VfmaternQFqt}
\end{subfigure}
\begin{subfigure}[b]{0.33\textwidth}
        \centering
        \includegraphics[width=\linewidth]{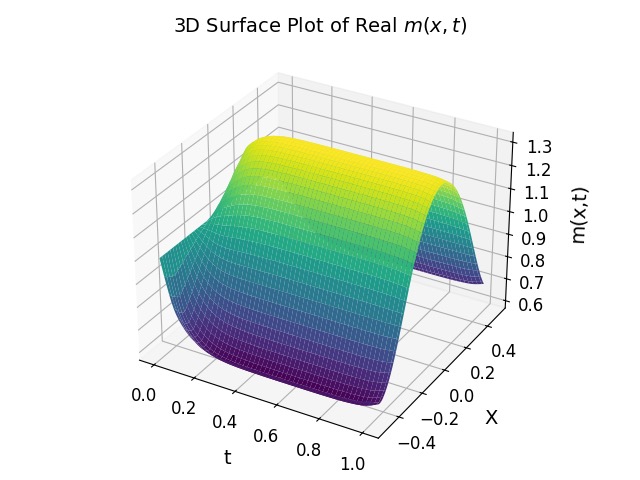}
        \subcaption{Reference Solution $m^*$}
        \label{fig:real1}
    \end{subfigure}%
    \begin{subfigure}[b]{0.33\textwidth}
        \centering
        \includegraphics[width=\linewidth]{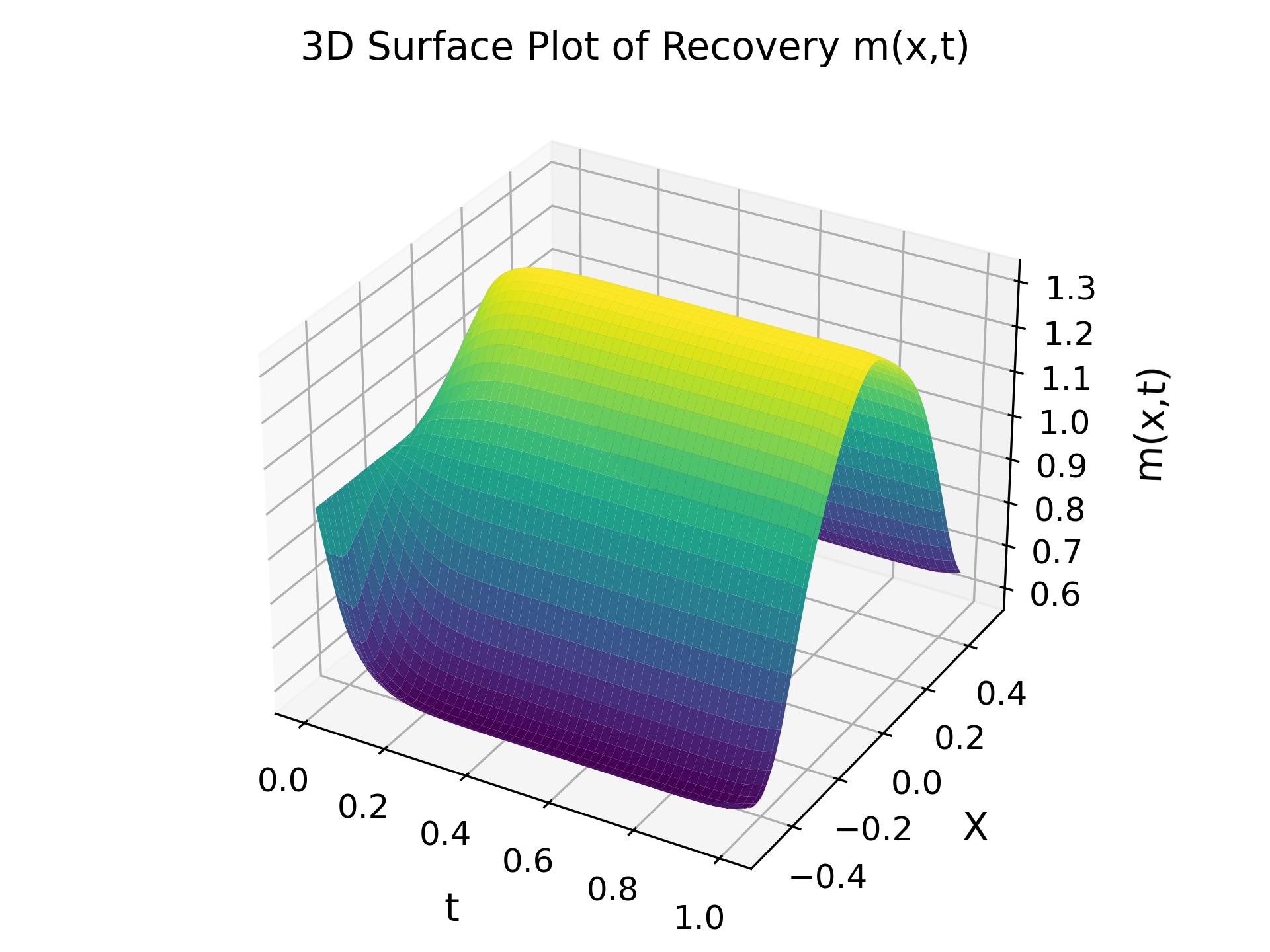}
        \subcaption{Recovered $m$}
        \label{fig:recover1}
    \end{subfigure}%
    \begin{subfigure}[b]{0.33\textwidth}
        \centering
        \includegraphics[width=\linewidth]{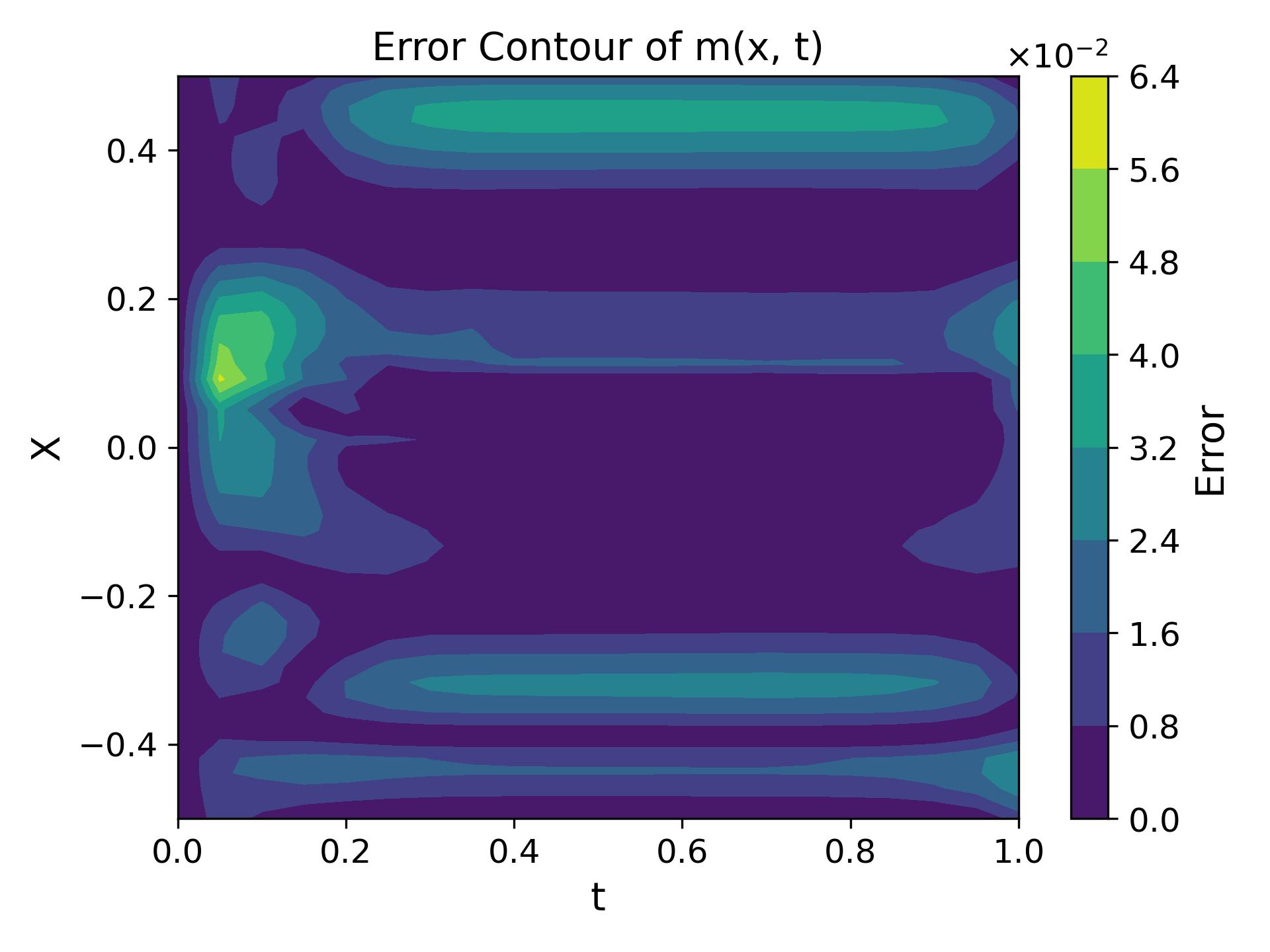}
        \subcaption{Error Contour of $m$}
        \label{fig:errorcontour1}
    \end{subfigure}

    \begin{subfigure}[b]{0.33\textwidth}
        \centering
        \includegraphics[width=\linewidth]{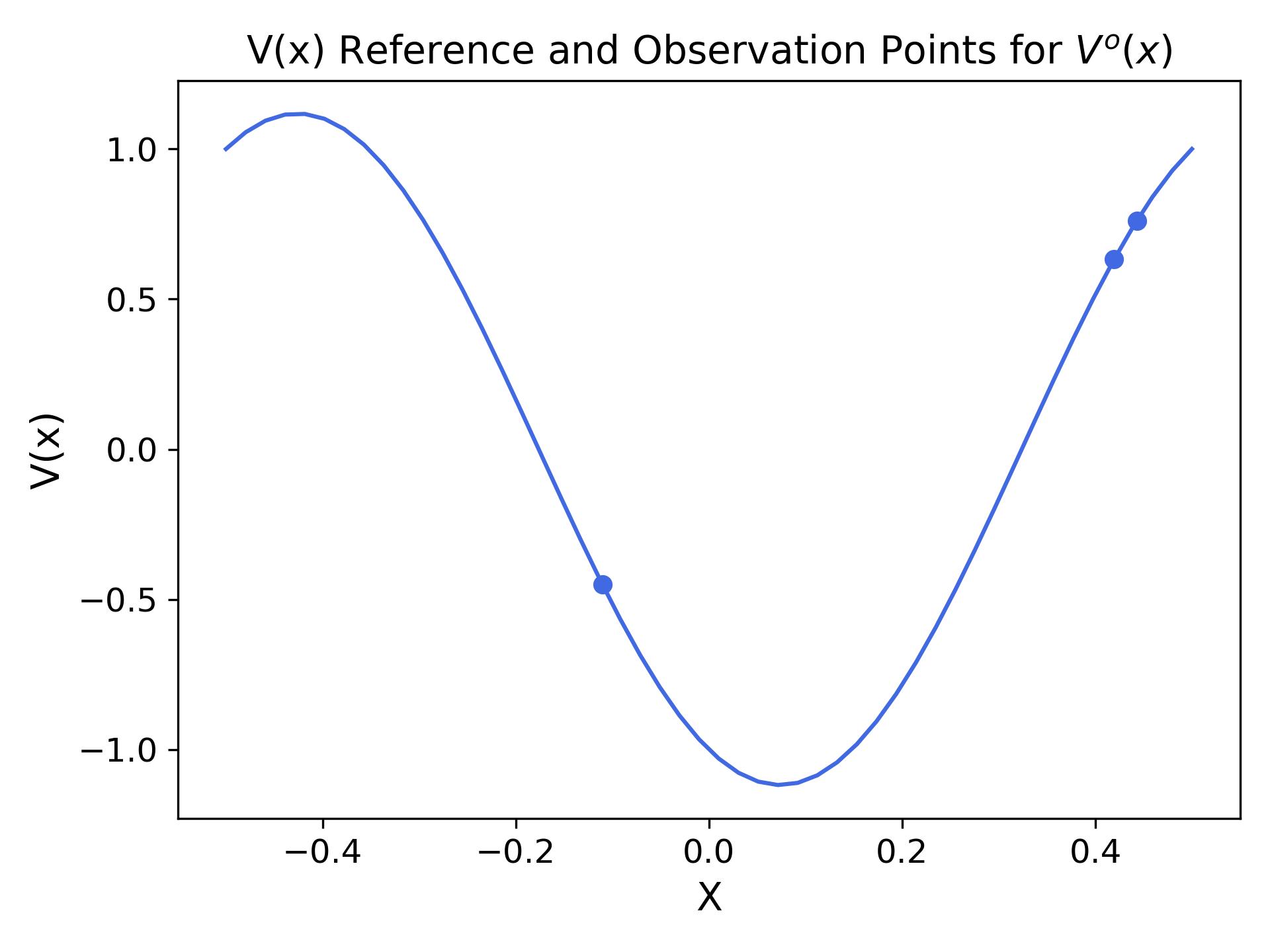}
        \subcaption{Ground Truth $V$ with  3 Observations}
        \label{fig:msample1}
    \end{subfigure}%
    \begin{subfigure}[b]{0.33\textwidth}
        \centering
        \includegraphics[width=\linewidth]{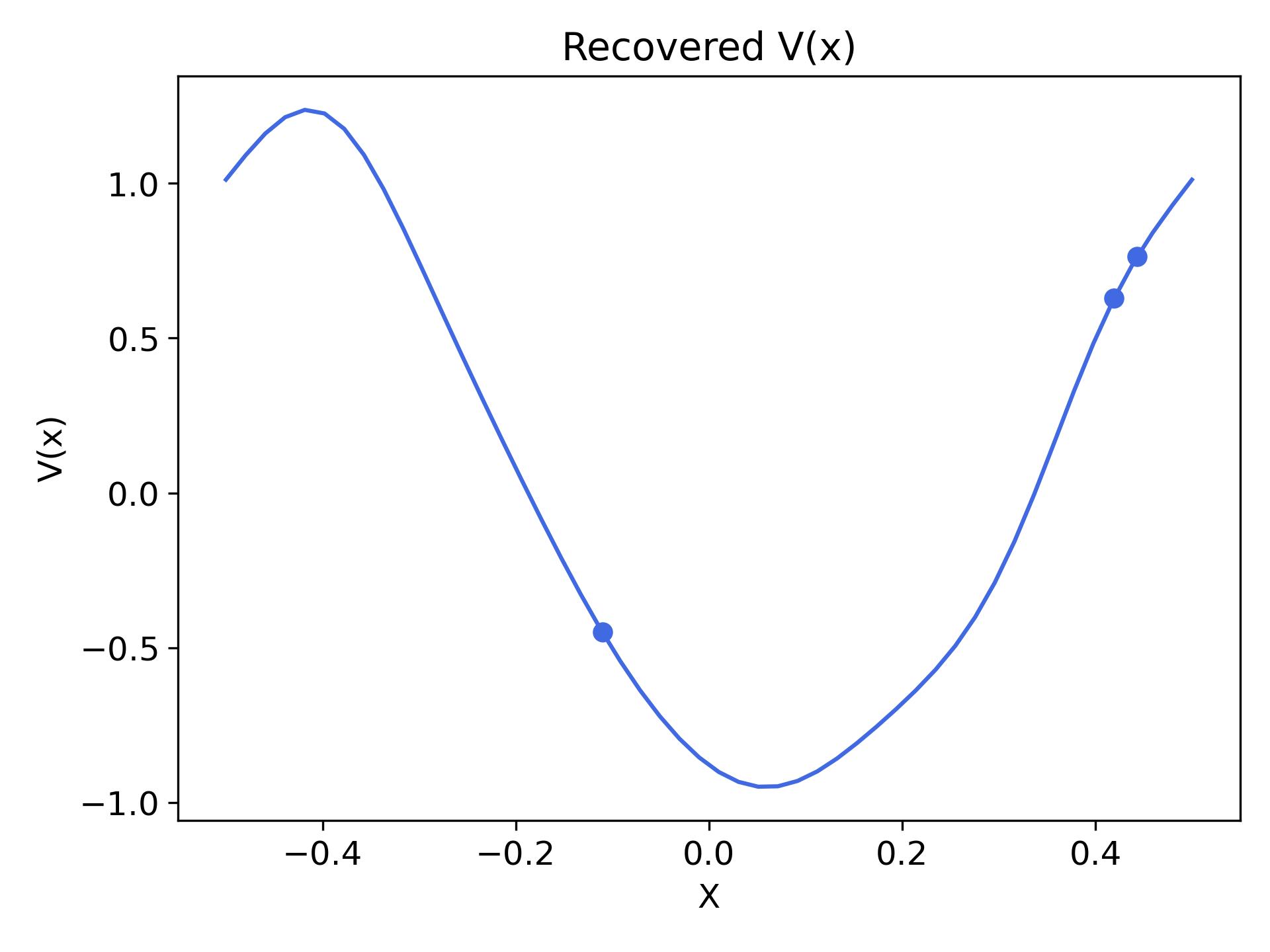}
        \subcaption{Recovered $V$ with  3 Observations}
        \label{fig:v01}
    \end{subfigure}%
    \begin{subfigure}[b]{0.33\textwidth}
        \centering
        \includegraphics[width=\linewidth]{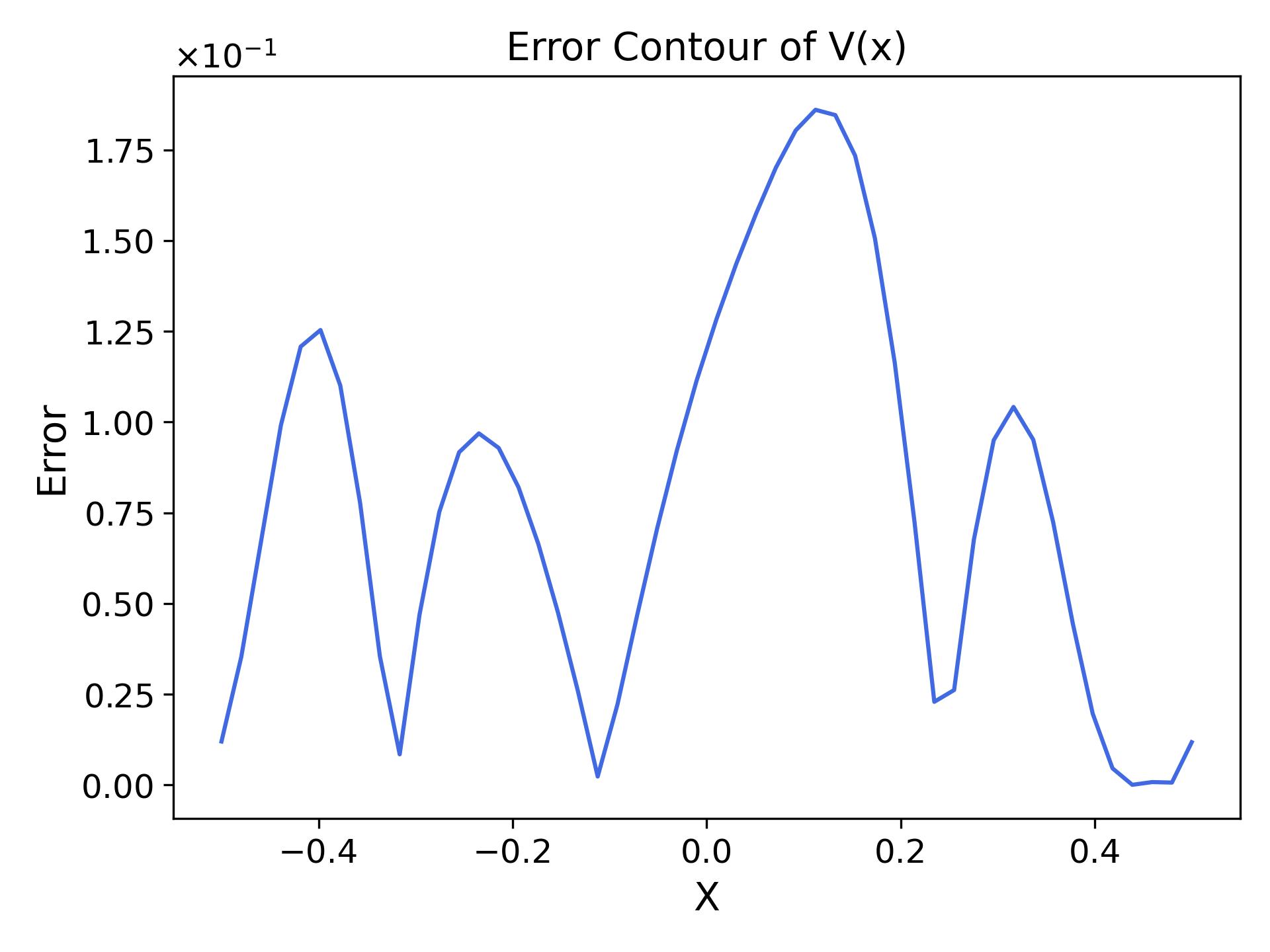}
        \subcaption{Pointwise Errors of $V$}
        \label{fig:error2}
    \end{subfigure}
    \caption{Numerical results for the inverse problem of the time-dependent MFG system in Subsection \eqref{subsec:num:td}: (a) grid and observation points for \(m\); (b) discretized \(L^2\) error \(\mathcal{E}(m^k, m^*)\) vs. iteration \(k\); (c)recovered \( F \) vs. reference \( F \); (d) reference solution \(m^*\); (e) recovered \(m\); (f) pointwise error \(|m - m^*|\); (g) ground truth \(V\) with three observation dots; (h) recovered \(V\) with three observation dots; (i) pointwise errors of \(V\) (recovered vs. ground truth). Despite differing from the ground truth, the recovered \(F\) remains convex, and the resulting MFG solutions fit the observed data well.
    }
    \label{figVQqtnotI}
\end{figure}

\textbf{Experimental Setup.} 
We discretize the spatial domain \(\mathbb{T}\) with a grid size \(h_x = \tfrac{1}{50}\) and the time interval \([0,1]\) with a grid size \(h_t = \tfrac{1}{20}\), resulting in 1,000 total grid points. Within these, 60 points are selected as observations for \(m\), while three observation points for \(V\) are randomly generated in the spatial domain, independent of the grid. The regularization parameters are set to \(\alpha_{m^o,k} = 1.5 \times 10^6\), \(\alpha_{v^o,k} = \infty\), \(\alpha_{v,k} = 2 \times 10^3\), \(\alpha_{fp,k} = 0.1\), \(\alpha_{f,k} = 30\), \(\alpha_{\lambda} = 1500\), \(\alpha_q = 5 \times 10^4\), and \(\alpha_{\nu} = 1.5 \times 10^6\). Gaussian noise \(\mathcal{N}(0,\gamma^2 I)\) with \(\gamma = 10^{-3}\) is added to the observations.

We initialize \(\Lambda = I + 0.7A\) (where \(A\) is an all-ones matrix) and set \(V = 0\), \(q = 2.7\), and \(\nu = 0.2\).  Since  \(F(m) = \frac{1}{4}m^4\) is non-negative for non-negative \(m\), we parameterize the coefficients of \(F\) as \(\boldsymbol{z}_F = \widetilde{\boldsymbol{z}}_F^2\). The initialization of \(\widetilde{\boldsymbol{z}}_F\), the choice of fixed points, and the adjoint-based optimization procedure follow the setup described in Section~\ref{MaternQ}, where we also employ the adjoint method to minimize the loss function.

\textbf{Experiment Results.} Figure \ref{figVQqtnotI} shows the collocation grid points and observation points for \( m \) along with the discretized \( L^2 \) errors \(\mathcal{E}(m^k, m^*)\) from \eqref{eq:l2disc}, measuring the discrepancy between the approximated \( m^k \) at iteration \( k \) and the reference solution \( m^* \). The figure also includes the reference solution, the recovered results, and pointwise error contours for both \( m \) and \( V \). The recovered values of \( q \) and \( \nu \) are \( q = 2.4046 \) and \( \nu = 0.10448 \), respectively, and the recovered matrix \(\Lambda\) is
\[
\Lambda = 
\begin{bmatrix}
1.3821 & 0.4033 \\
0.4033 & 1.3821
\end{bmatrix}.
\]

\noindent
This example illustrates the non-uniqueness of the model corresponding to the observed data. Although \( m \) and \( V \) are accurately recovered with very few observation points, the recovered metric matrix \(\Lambda\) and the coupling function \(F\) do not match the ground truth that generated the data. Nonetheless, the MFG with the recovered coefficients still explains the observations effectively.

\section{Conclusion and Future Works}
\label{sec:Conclusion}
In this work, we introduce GP methods to solve ill-posed inverse problems in potential MFGs. These problems involve recovering population, momentum, and environmental parameters from limited, noisy observations. We propose two solution frameworks: an inf-sup minimization approach, which applies when the unknowns appear as concave terms in the objective, and a bilevel formulation suitable for more general settings.  Leveraging the linearity of GPs preserves convexity and concavity, enabling standard convex solvers in the inf-sup case. For the bilevel problem, we adopt a gradient descent-based algorithm with two gradient-computation strategies: automatic differentiation, which integrates seamlessly with an existing solver for the inner MFG, and an adjoint-based method that is solver-agnostic. 

Although inverse problems are ill-posed without sufficient priors, our frameworks reliably yield surrogate MFG models that closely match observed data-offering insights into potential underlying dynamics and supporting tasks such as forecasting and scenario analysis. Potential extensions include integrating scalable techniques (e.g., Random Fourier Features, sparse GPs, and mini-batch methods) to handle large datasets more efficiently and applying our methods to real-world problems in economics, biology, and finance, where established MFG models are not yet available. Further directions include stronger priors, adaptive loss weighting, and advanced optimization methods to mitigate ill-posedness in both data-rich and data-scarce scenarios. Since potential MFGs can be reformulated as linear PDE-constrained convex minimization problems, our methods also apply to inverse problems in these broader settings.

%\bibliography{../bib/bibfile}
\section*{acknowledgement}
%JZ acknowledges the IoTeX Foundation Industry Grant A-8001180-00-00.
XY acknowledges support from the Air Force Office of Scientific Research through MURI award FA9550-20-1-0358 (Machine Learning and Physics-Based Modeling and Simulation). 
CM acknowledges the financial support of the Hong Kong Research Grants Council (RGC) under the Grants
No. GRF 11311422 and GRF 11303223. 
% CZ is supported by Singapore MOE AcRF Grant A-8000453-00-00, IoTeX Foundation Industry Grant A-8001180-00-00, and NSFC Grant 11871364.

%\bibliographystyle{plain}
%\bibliography{reference}

\end{document}